\documentclass[10pt,twocolumn,english]{IEEEtran}
\usepackage[T1]{fontenc}
\usepackage[latin9]{inputenc}
\usepackage{color}
\usepackage{babel}
\usepackage{array}
\usepackage{float}
\usepackage{multirow}
\usepackage{amsmath}
\usepackage{amssymb}
\usepackage{graphicx}
\usepackage[numbers]{natbib}
\usepackage[unicode=true,
 bookmarks=true,bookmarksnumbered=false,bookmarksopen=false,
 breaklinks=true,pdfborder={0 0 0},backref=false,colorlinks=true]
 {hyperref}

\makeatletter

\providecommand{\tabularnewline}{\\}
\floatstyle{ruled}
\newfloat{algorithm}{tbp}{loa}
\providecommand{\algorithmname}{Algorithm}
\floatname{algorithm}{\protect\algorithmname}

\@ifundefined{date}{}{\date{}}

\usepackage{microtype}

\usepackage{algpseudocode}

\@ifundefined{showcaptionsetup}{}{%
 \PassOptionsToPackage{caption=false}{subfig}}
\usepackage{subfig}
\makeatother

\begin{document}

\title{Robust Face Recognition by Constrained Part-based Alignment}

\author{Yuting Zhang, Kui Jia, Yueming Wang, Gang Pan, Tsung-Han Chan, Yi
Ma %
\thanks{Y. Zhang and G. Pan are with the Department of Computer Science, Zhejiang
University, Hangzhou 310027, P.\ R. China (e-mail: zyt@zju.edu.cn;
gpan@zju.edu.cn)%
}%
\thanks{K. Jia is with the Department of Electrical and Computer Engineering,
Faculty of Science and Technology, University of Macau, Macau, China.
(email: kuijia@gmail.com)%
}%
\thanks{Y. Wang is with the Qiushi Academy for Advanced Studies, Zhejiang
University, Hangzhou 310027, P.\ R. China (e-mail: ymingwang@gmail.com)%
}%
\thanks{T. Chan is with the Advanced Digital Sciensces Center (ADSC), Singapore 138632. (email: thchan@ieee.org) %
}%
\thanks{Y. Ma is with the School of Informatin Science and Technology, ShanghaiTech
Universitiy, Shanghai 200031, P.\ R.China. (email: mayi@shanghaitech.edu.cn)
He is also with the Department of Electrical and Computer Engineering,
University of Illinois at Urbana-Champaign, IL 61801, USA. (email: yima@uiuc.edu)%
}}

\maketitle



\begin{abstract}
Developing a reliable and practical face recognition system is a long-standing
goal in computer vision research. Existing literature suggests that
pixel-wise face alignment is the key to achieve high-accuracy face
recognition. By assuming a human face as piece-wise planar surfaces,
where each surface corresponds to a facial part, we develop in this
paper a Constrained Part-based Alignment (CPA) algorithm for face
recognition across pose and/or expression. Our proposed algorithm
is based on a trainable CPA model, which learns appearance evidence
of individual parts and a tree-structured shape configuration among
different parts. Given a probe face, CPA simultaneously aligns all
its parts by fitting them to the appearance evidence with consideration
of the constraint from the tree-structured shape configuration. This
objective is formulated as a norm minimization problem regularized
by graph likelihoods. CPA can be easily integrated with many existing
classifiers to perform part-based face recognition. Extensive experiments
on benchmark face datasets show that CPA outperforms or is on par
with existing methods for robust face recognition across pose, expression,
and/or illumination changes. 
\end{abstract}

\section{Introduction \label{sec:Intro}}

Developing a reliable and practical face recognition system is a long-standing
goal in computer vision research. A tremendous amount of works have
been done in the past three decades, however, most of them can only
work well in controlled scenarios. In more practical scenarios, performance
of these existing methods degrades drastically due to face variations
caused by illumination, pose, and/or expression changes \citep{ZhaoFaceRecogSurvey}.


To handle the variation caused by illumination change, the methods
of \citet{NS,andrew2012practical}, motivated by the illumination
cone model \citep{IllumCone,FromFewToMany}, used multiple carefully
chosen face images of varying illuminations per subject as gallery.
Given a probe face of unknown subject under arbitrary illumination,
face images under this specific illumination can be generated in the
gallery to match with the probe face. When multiple gallery images
of such kind are not available, alternative methods \citep{ChenIllumInv,ZhouIllumInv}
considered extracting illumination invariant features for face recognition.

To address pose or expression variations, earlier approaches extend
classic subspace or template based face recognition methods \citep{PentlandMultiView,BeymerVaryingPose}.
The use of these methods is rather restricted due to their dependence
on the availability of gallery images of multiple facial poses or
expressions. Recent approaches consider more practical scenarios where
only face images under the normal condition (frontal view and neutral
expression) are assumed to be available in the gallery. To recognize
a probe face with pose or expression changes, they either identified
an implicit identity feature/representation of the probe face \citep{pami2008tied-factor,cvpr2009expresion-invariant,fg2002light-field},
which is pose- and expression-invariant, or explicitly estimate global
or local mappings of facial appearance so that a virtual face under
the normal condition can be synthesized for recognition \citep{tip2007local-regression,cvpr2008patch,cvpr2011flowface,pami2011-mrf-face,cvpr2011widebaseline}.
However, most of these methods are still far from practice since they
assume both gallery and probe face images have been manually aligned
into some canonical form, and they cannot generally cope with illumination
variation either.

In literature, the most popular approaches for automatic face alignment
across pose or expression are based on facial landmark localization,
e.g., the Active Appearance Models (AAMs) \citep{cootes2001aam,AAMRevisited}
and elastic graph matching (EGM) \citep{ElasticBunch}. AAMs and EGM
used densely-connected elastic graphs, which however, are difficult
to optimize in that the solutions are likely to be trapped into undesirable
local minima. Consequently, localized landmarks using these methods
are often not accurate enough, especially when applied to unseen face
images. To improve the localization accuracy, an explicit shape constraint
for graph nodes was considered in the constrained local models (CLMs)
\citep{CLM}. CLMs \citep{CLM,BelhumeurCLM,SaragihCLM} are still
based on densely-connected graph models, and their shape constraints
are over-simplified so that the dependency among different graph nodes
is ignored. Recently, deformable part-based models (DPMs) show their
promise in many applications such as object detection \citep{pff2010dpm}
and facial landmark localization \citep{zhu2012dpmface}. In particular,
\citet{zhu2012dpmface} adopted a tree-structured part model, which
encodes node dependency while admitting efficient solutions. Such
a tree model was used by \citet{zhu2012dpmface} for facial landmark
localization. Nevertheless, it cannot be readily extended for pixel-wise
face alignment, and consequently for face recognition across pose
or expression. 

With the aim of developing a system that can simultaneously handle
illumination variation and minor changes of pose and expression, \citet{andrew2012practical}
recently leveraged sparsity optimization and a carefully prepared
gallery set (multiple images of varying illuminations per subject),
to align probe face images into a canonical form for better recognition.
In particular, they assumed human face as a planar surface and chose
a \textit{global} similarity transformation for face alignment. This
may be valid when probe faces are close to the normal condition. It
is however a rather simplified assumption when there exist pose and/or
expression changes. Indeed, a human face has non-planar geometry and
non-rigid deformation. Some portions of the face (e.g., the nose region)
undergo significant appearance changes as the pose varies, while other
portions (e.g. the mouth region) deform significantly as expression
changes. It is thus more appropriate to approximate a human face as
a piece-wise planar surface, and correspondingly use piece-wise geometric
transformation for face alignment.

Improving performance using enriched models is not an easy task in
computer vision research. In fact, opposite effects often happen.
As to the problems considered in this paper, we will show that by
partitioning a human face as a collection of parts, and carefully
characterizing the deformation relations among different parts, performance
of face recognition across pose or expression can be significantly
improved. In particular, we propose a Constrained Part-based Alignment
(CPA) method for this task. Our method is partially motivated by the
promise of piece-wise planar formulation and the explicit shape constraints
used in CLMs \citep{CLM,BelhumeurCLM,SaragihCLM} and DPMs \citep{pff2010dpm,zhu2012dpmface}. 

CPA partitions the object of interest, e.g., a human face, as a constellation
of parts, and uses a similarity transformation to model the deformation
of each part. A tree-structured shape model is used in CPA to constrain
the relations among the deformations of different parts. A CPA model
also has a batch of registered face images serving as the appearance
evidence of each part. For a probe face image, all its parts are simultaneously
aligned towards the registered model, by fitting them to the appearance
evidence and penalizing the cost of violating the constraint from
the tree-structured shape model. We formulate this objective as a
norm minimization problem regularized by graph likelihoods, which
is solved by an alternating method composed of two steps: one for
aligning the parts, and the other for adjusting the configuration
of the tree-structured model. The former can be reduced into a sequence
of convex problems, while the latter admits efficient solutions by
gradient decent methods. In this paper, we use the proposed CPA model
for face recognition, where registered gallery images are taken as
the appearance evidence, and a probe image is aligned using the CPA
model. After alignment, most of the off-the-shelf face recognition
methods can be readily used for part recognition.The overall decision
is made by aggregating predictions from different parts by a plurality
voting scheme. Robustness against illumination variation can also
be achieved by choosing specific face recognition methods, e.g., \citep{andrew2012practical,john2009src}.

Richer models are often more difficult to train. For the proposed
CPA model, both the appearance model and the tree-structured shape
model need to be automatically learned. On the one hand, the appearance
model is obtained by aligning the gallery images in batch with the
constraint from a given shape model. In particular, we use low-rank
and sparse matrix decomposition as the criteria to optimize batch
alignment of each part, and globally regularize these individual problems
of part alignment by graph likelihoods. On the other hand, the shape
model is learned by the probabilistic inference based on given part
constellations, where the maximum \emph{a posteriori} (MAP) estimation
is performed with the graph likelihoods and the given conjugate priors
of the likelihoods. As the two problems are coupled, we solve them
in a joint way. Moreover, we generalize our learning method to train
a mixture of CPA models (mCPA) for better handling pose and expression
variations.


Part-based face recognition methods \citep{EGFC,part-face-nir,kumar2011kernel-plurality,zhang2011joint-dynamic,Ocegueda2011which-part}
have shown improved performance over more standard approaches using
holistic faces. However, existing part-based methods can only be applied
when gallery and probe face images have been manually registered to
each other. Our proposed CPA method enables this promising part-based
strategy to be applicable in more practical scenarios, by automatically
registering the gallery and probe face images in terms of deformable
parts. \textit{\emph{In this paper, we present experiments on the
Multi-PIE \citep{MultiPIE} and MUCT \citep{MUCT} datasets and show
that our proposed CPA method can simultaneously and effectively handle
illumination, pose, and/or expression variations. }}
\begin{itemize}
\item \textit{\emph{Comparing with the natural alternative methods that
holistically align face images, followed by either holistic face recognition
or part-based recognition, our method gives significantly improved
performance. }}
\item \textit{\emph{State-of-the-art pose-invariant face recognition method~\citep{eccv2012facepose}
relies on model learning using a large number of 3D face shapes, while
training of our proposed CPA only requires 2D face images of a few
subjects. Nevertheless, our method outperforms~that of \citet{eccv2012facepose}.
when the degrees of pose change are within $\pm\mbox{15}^{\circ}$.
This range of pose change is often encountered in practical access
control scenarios, where test subjects would be cooperative with face
recognition systems. }}
\item \textit{\emph{Notably, while CPA is motivated to address the challenges
of face recognition across pose/expression, it performs surprisingly
well when probe face images are at frontal view and with neural expression.
Our results in the frontal-view, neutral-expression, varying-illumination,
and across-session setting of Multi-PIE dataset are better than all
existing methods, and as high as 99.6\%. This confirms that considering
human face as a piece-wise planar surface and aligning it part-wisely
are very effective for high-accuracy face recognition. }}
\item \textit{\emph{We also empirically investigate the discriminative power
of individual facial parts, and the robustness of our method against
partial occlusion. }}
\end{itemize}
\textit{\emph{Details of these investigations  are presented in Section
\ref{sec:experiments}.}}

The rest of this paper are organized as follows. Section~\ref{sec:Related-work}
reviews more related work in addition to those discussed in Section~\ref{sec:Intro}.
Section~\ref{sec:CPA} presents our proposed CPA model and how to
use it for alignment of probe face images. Section~\ref{sec:Recognition}
combines CPA with existing methods for part-based face recognition.
Details of CPA model learning are presented in Section~\ref{sec:Model-learning},
where we also extend CPA to a mixture of CPA for better handling pose
or expression variations. Intensive experiments are finally reported
in Section~\ref{sec:experiments} to show the efficacy of our proposed
method. 

\begin{figure*}
\centering{}\includegraphics[width=0.85\textwidth]{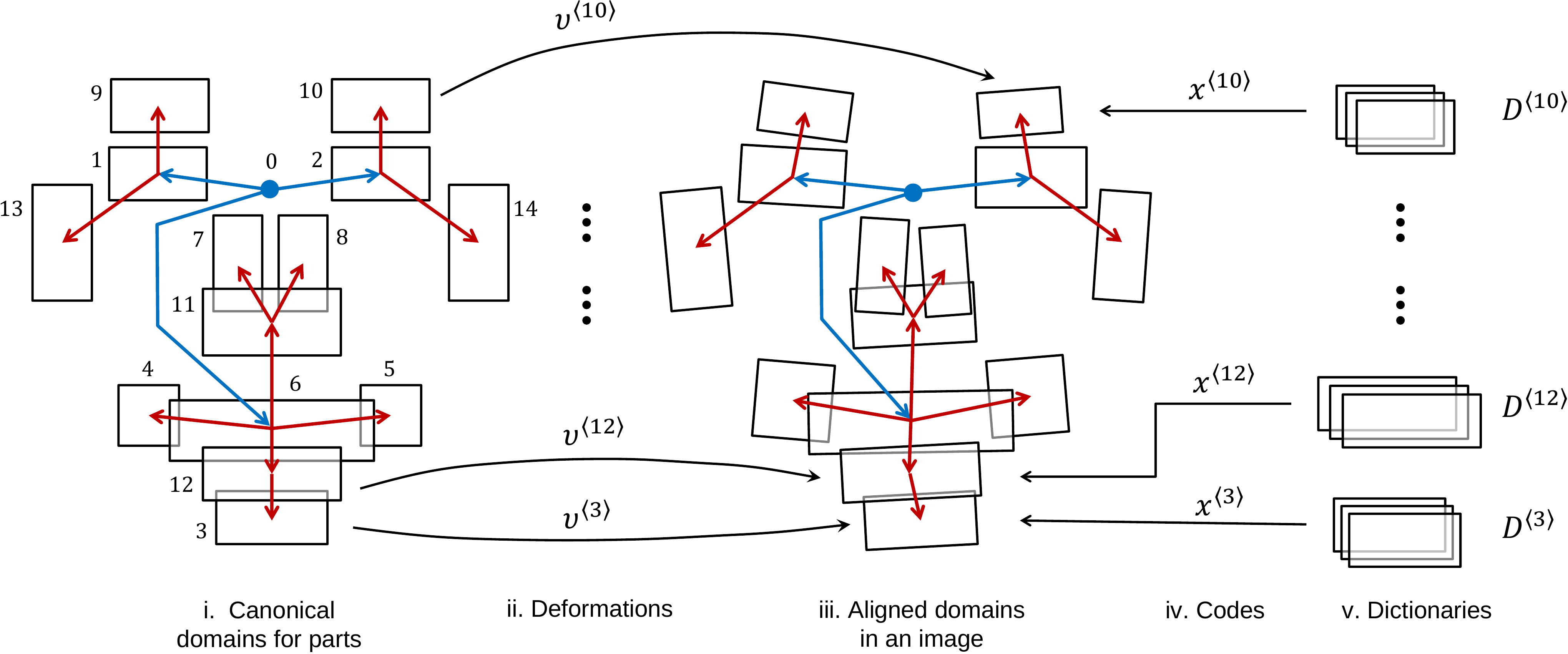}\protect\caption{\label{fig:CPA-model} Illustration of a CPA model instance with 14
facial parts: Face is considered as a constellation of parts constrained
by a tree structure, where each part corresponds to a pre-defined
domain (i.e., shape and location) (Sub-figure~i). For an image, deformations
are computed (Sub-figure~ii) for transforming the parts to the domains
(Sub-figure~iii) where they are aligned to the appearance evidence
given by the linear combination (Sub-figure~iv) of existing registered
images (Sub-figure~v). }
\end{figure*}

\section{Related Work \label{sec:Related-work}}

As discussed in Section \ref{sec:Intro}, most of existing methods
for face recognition across pose or expressions assume gallery and
probe face images have been manually registered, and they can only
be applied to controlled scenarios. In particular, \citet{tip2007local-regression}
learned local patch based mapping relations from non-frontal pose
to frontal pose by locally linear regression. \citet{cvpr2011flowface}
estimated pixel-wise registration from non-neutral expression to neutral
expression by optical flow. \citet{pami2011-mrf-face} considered
a Markov random field (MRF) model to regularize 2D displacements of
local patches across different poses. For automatic face recognition
across pose, given a probe face image, AAMs \citep{cootes2001aam,AAMRevisited}
optimized localization of a set of facial landmarks to realize face
alignment of pixel-level accuracy. However, the landmarks localized
by AAMs are usually not accurate enough, and also the pixel-wise correspondence
induced by matched landmarks are not consistent enough across different
poses and expressions \citep{cvpr2011pose3d}. Among existing methods,
maybe the most successful ones across pose and/or expression are based
on 3D models \citep{pami2003face3d,cvpr2011pose3d,eccv2012facepose}.
In spite of their promise, their relying on 3D data makes them less
relevant to the 2D techniques considered in this paper. However, we
will show that our proposed CPA method compares favorably with them
when the pose variations are in a reasonably confined range.

Deformable part models (DPM) have shown their success in object detection
\citep{pff2010dpm}, facial landmark localization \citep{zhu2012dpmface},
and human pose estimation\textbf{ }\citep{yang2012dpmhuman}. In particular,
these methods optimize an objective function that scores both the
appearance evidence and spatial constellation of the parts, where
the former is scored by part detectors, and the latter is scored by
a star- or tree-structured shape model, which constrain the pair-wise
offsets of different parts.\textbf{ }Our use of part-based models
is different from the DPM based methods \citep{pff2010dpm,zhu2012dpmface,yang2012dpmhuman}.
Our CPA method aims at pixel-wise image alignment rather than finding
the bounding box of an object/part or locating a small amount of landmarks.
We measure the appearance evidence of parts by their similarity to
aligned galleries, and constrain the part deformation relations by
a tree-structured shape model. Compared with shape constraints in
DPMs\textbf{ }\citep{zhu2012dpmface,yang2012dpmhuman},\textbf{ }our
proposed one models more complex relations with the part constellation
and holds strict probabilistic properties that are beneficial to the
model learning.

Our method is also closely related to the work of \citet{andrew2012practical}.
To cope with illumination variation, they used multiple gallery images
of varying illuminations per subject. Our method also follows this
strategy. However, we\textbf{ }take a human face as a collection of
parts rather than assuming it as a planar surface as in the method
of \citet{andrew2012practical}. The piece-wise planar assumption
used in our CPA model can handle much more complex facial appearance
variations such as large pose and/or facial expression changes. Furthermore,
we regularize the deformations of individual parts by a shape constraint
in order to prevent the alignment from degenerated solutions. Such
regularization term is absent in the method of \citet{andrew2012practical}.


\section{CPA: Constrained Part-based Alignment \label{sec:CPA}}

\begin{figure*}
\begin{centering}
\hfill{}\subfloat[Part dictionaries \label{fig:dictionary-single}]{\begin{centering}
\includegraphics[width=0.315\columnwidth]{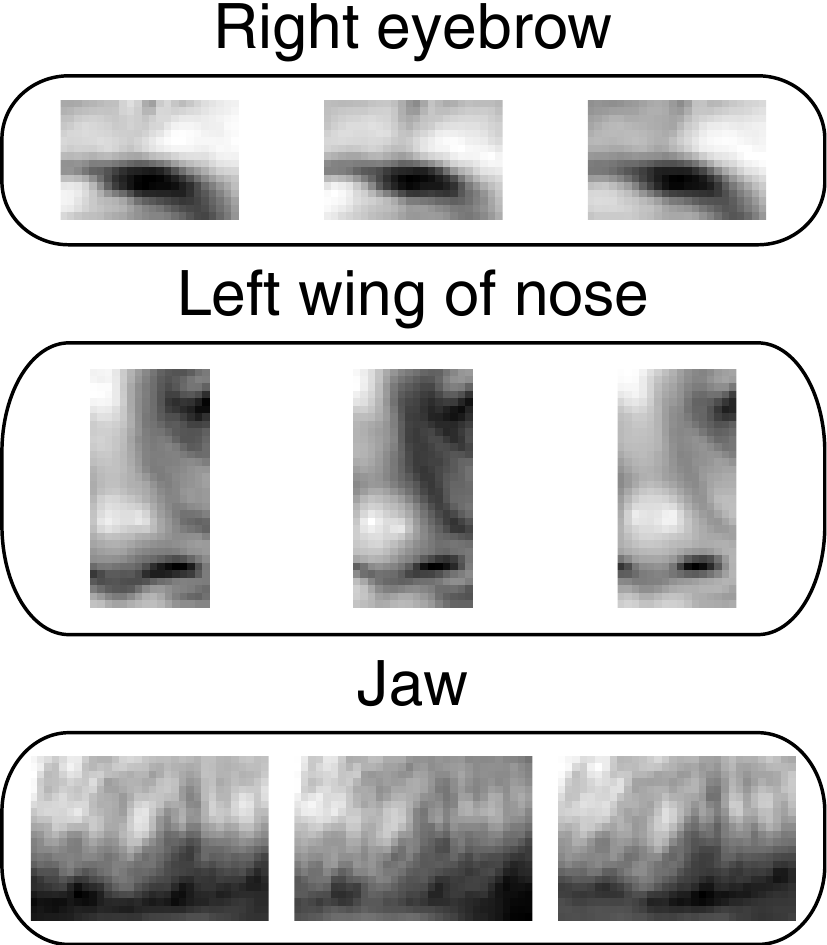}
\par\end{centering}

}\hfill{}\hfill{}\subfloat[Independent part alignment by \citet{andrew2012practical}'s method
at Iteration~17 \label{fig:independent-single}]{\begin{centering}
\includegraphics[width=0.27\columnwidth]{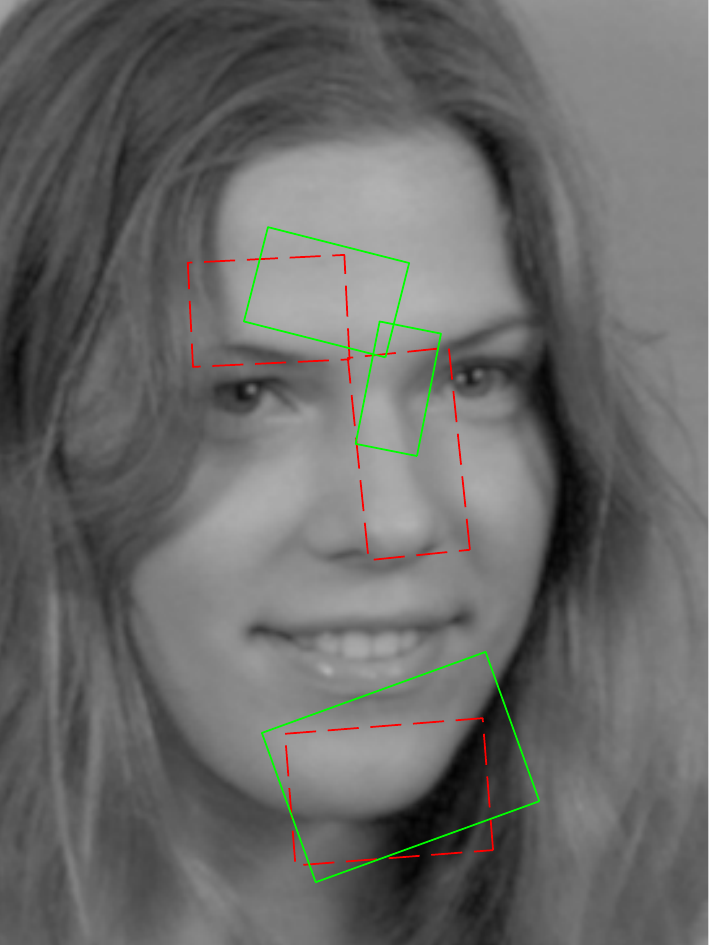}\hspace{0.05\columnwidth}\includegraphics[width=0.315\columnwidth]{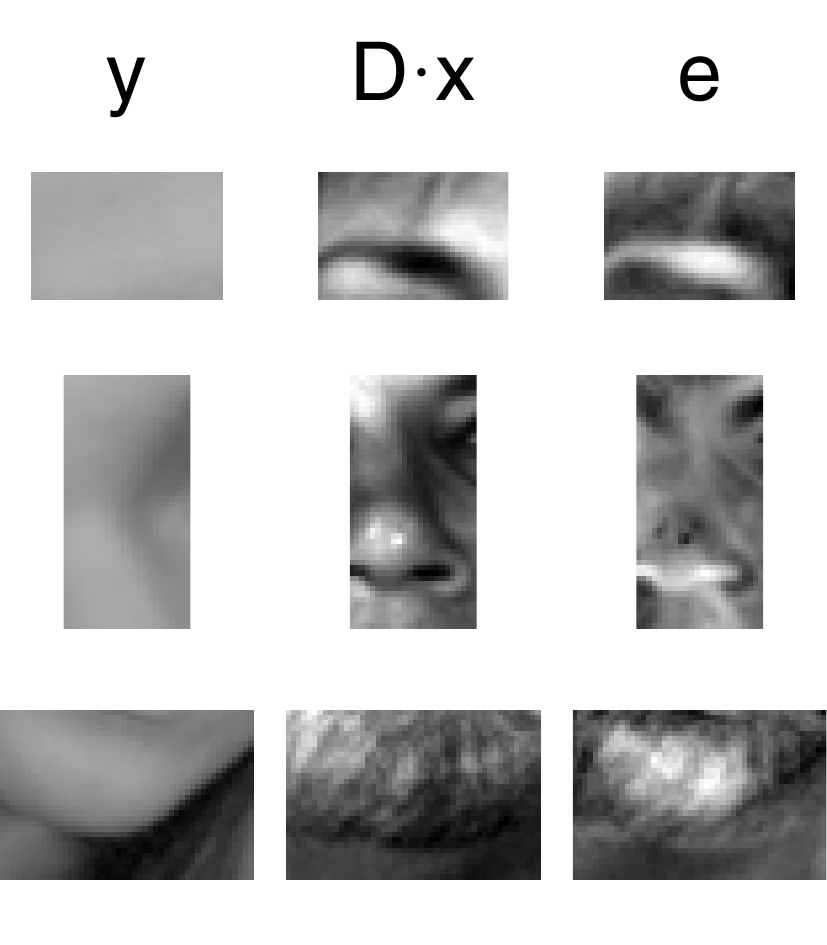}
\par\end{centering}

}\hfill{}\hfill{}\subfloat[Our CPA method at the converged stage \label{fig:CPA-single}]{\begin{centering}
\includegraphics[width=0.27\columnwidth]{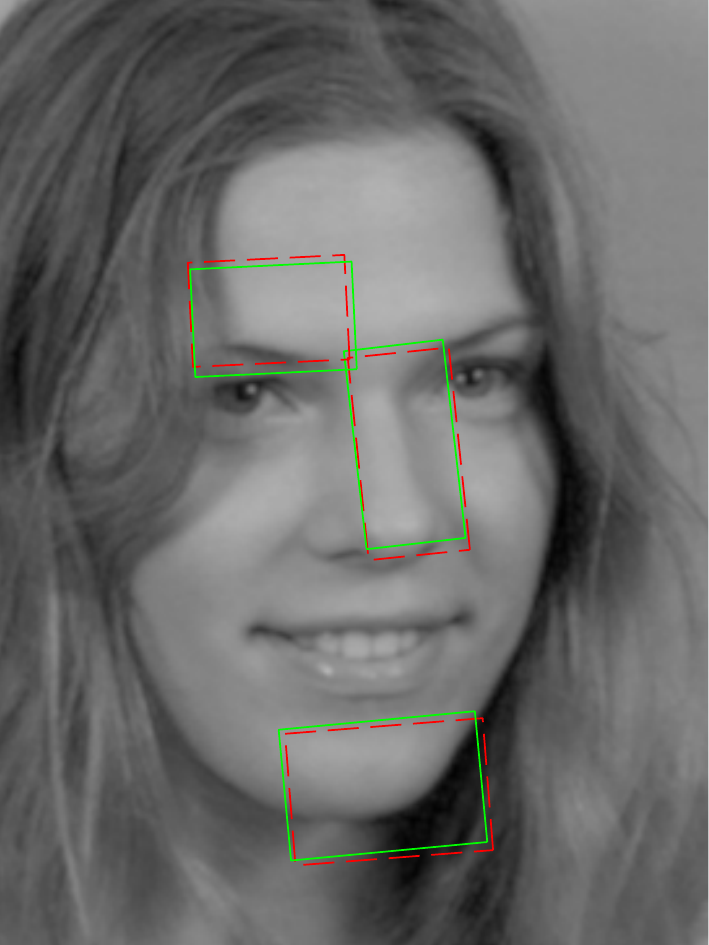}\hspace{0.05\columnwidth}\includegraphics[width=0.315\columnwidth]{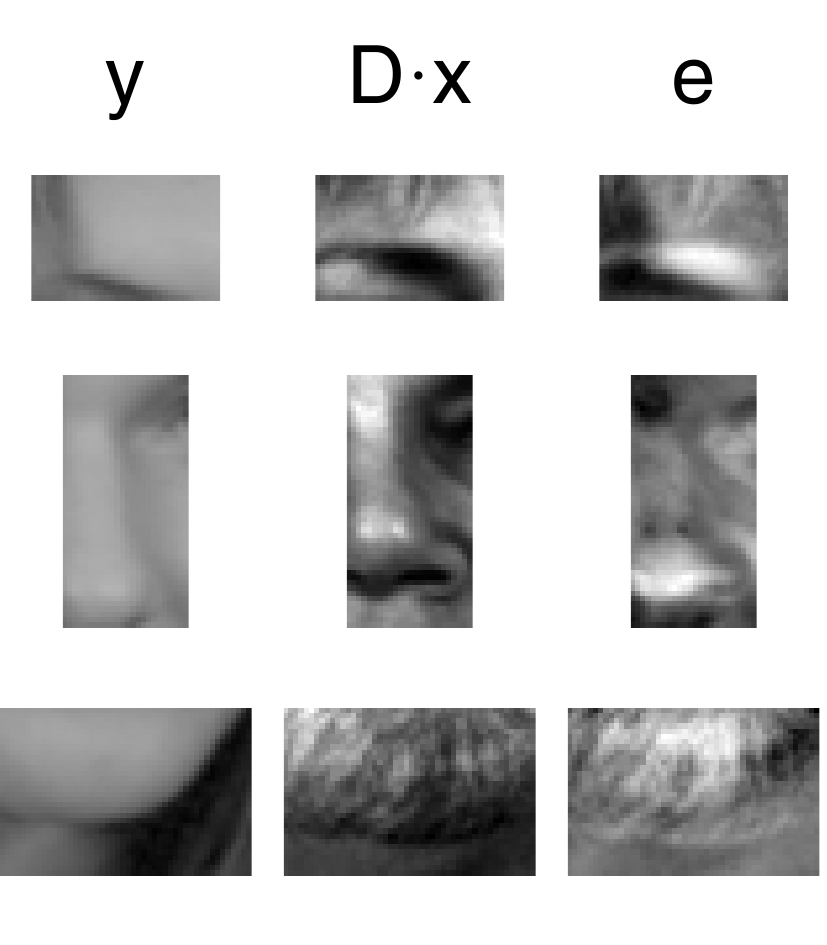}
\par\end{centering}

}\hfill{}
\par\end{centering}

\protect\caption{Part alignment obtained without/with a shape constraint. Parts are
initialized at dashed red boxes, and aligned to the appearance evidence
suggested by the part dictionaries. The solid green boxes denote the
obtained domains of the parts. $y$ is the image cropped from an obtained
domain, $D\cdot x$ is its best linear reconstruction with the given
part dictinoary, and $e$ is the reconstruction error. The cropped
images are normalized in terms of intensity for display convenience. }
\end{figure*}

Assume we have face images of multiple subjects in a database, and
these images have been registered into some canonical form, i.e.,
aligned to some template. The template we consider in this paper is
composed of a collection of parts. These parts of varying sizes are
spatially arranged in a proper manner so that overall they define
our face template (Fig.~\ref{fig:CPA-model}-i). Suppose there are
$m$ parts in the template. Each of the $m$ parts has its associated
face region in each image in the database. We call these face regions
associated with every $i^{\textrm{th}}$ part a \emph{part dictionary},
which is denoted as $D^{\langle i\rangle}$ (Fig.~\ref{fig:CPA-model}-v).
Let $y$ be a test face image that is not generally in the canonical
form (e.g, due to pose change or misalignment). As the name of CPA
suggests, constrained part-based alignment aims to align $y$ to the
face template so that different face regions of $y$ are respectively
registered with the corresponding parts of the face template. This
part-based alignment can be realized by pursuing a set of transformations
$\boldsymbol{\nu}=(\nu^{\langle1\rangle},\nu^{\langle2\rangle},\ldots,\nu^{\langle m\rangle})$
that act on the image domain of $y$ by $y\circ\nu^{\langle i\rangle}$,
$i=1,\dots,m$. %
\footnote{Let $\{(u,v)\}$ denote image coordinates of $y$, and $\nu^{\langle i\rangle}$
be a 2-D coordinate transformation function. Then, image transformation
is realized by $y\circ\nu^{\langle i\rangle}(u,v)=y(\nu^{\langle i\rangle}(u,v))$,
and we also use $y\circ\nu^{\langle i\rangle}$ denotes the transformed
image (the $i^{th}$ part). $\{\nu^{\langle i\rangle}\}_{i=1}^{m}$
belong to a $d$-dimensional transformation group $\mathbb{G}$, e.g.
the similarity group or affine group. In this paper, we assume $\mathbb{G}$
to be the 2-D similarity group and parameterize the similarity transformation
mapping $(u,v)$ to $(u',v')$ as $(t_{u},t_{v},s,\theta)^{T}\in\mathbb{R}^{4}$
that satisfies 
\begin{equation}
\begin{pmatrix}u'\\
v'
\end{pmatrix}=\exp(s)\begin{pmatrix}\cos\theta & -\sin\theta\\
\sin\theta & \cos\theta
\end{pmatrix}\begin{pmatrix}u\\
v
\end{pmatrix}+\begin{pmatrix}t_{u}\\
t_{v}
\end{pmatrix}.\label{eq:similarity-para}
\end{equation}
With a little abuse of notation, we simultaneously take $\nu^{\langle i\rangle}$
as a transform function and a column vector composed of the $d$ parameters
of $\mathbb{G}$ ($d$ = 4). Accordingly, $\boldsymbol{\nu}\in\mathbb{R}^{d\times m}$. %
}

To pursue $\boldsymbol{\nu}$, we consider techniques used in \citep{john2009src,andrew2012practical}.
In particular, \citet{john2009src,andrew2012practical} assumed there
exist multiple registered face images of varying illuminations per
subject in the database. A well-aligned test image could be represented
by a linear combination of registered training images, plus a sparse
error term to compensate for data corruption or various intra-subject
variations. By leveraging the error sparsity assumption, \citet{andrew2012practical}
optimized a holistic similarity transformation by solving a $\ell^{1}$-norm
minimization problem. Extending techniques in \citep{andrew2012practical}
directly to part-based alignment gives the following objective 
\begin{gather}
\min_{\underset{i=1,2,\ldots,m}{x^{\langle i\rangle},e^{\langle i\rangle},\boldsymbol{\nu}}}\;\sum_{i}^{m}\lambda^{\langle i\rangle}\Vert e^{\langle i\rangle}\Vert_{1}\nonumber \\
\mathrm{s.t.}\quad y\circ\nu^{\langle i\rangle}=D^{\left\langle i\right\rangle }x^{\langle i\rangle}+e^{\langle i\rangle},\label{eq:CPA-constraint-0}
\end{gather}
where $\|\cdot\|_{1}$ is the $\ell^{1}$-norm that encourages the
sparsity of errors $\{e^{\langle i\rangle}\}_{i=1}^{m}$. $y\circ\nu^{\langle i\rangle}$
aligns the $i^{\textrm{th}}$ part of $y$, and $\{\lambda^{\langle i\rangle}\in\mathbb{R}_{+}\}_{i=1}^{m}$
balances the alignment errors of the $m$ parts. The matrix $D^{\langle i\rangle}$
is the $i^{\textrm{th}}$ part dictionary, whose columns represent
the aligned regions of face images in the database, and $x^{\langle i\rangle}$
denotes the reconstruction coefficient.

\begin{figure}
\begin{centering}
\includegraphics[width=0.27\columnwidth]{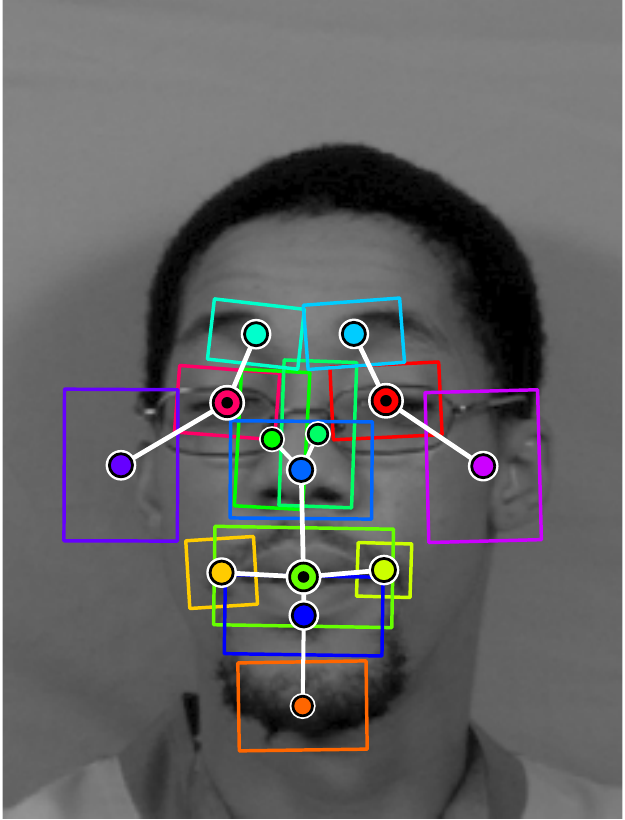}\hfill{}\includegraphics[width=0.27\columnwidth]{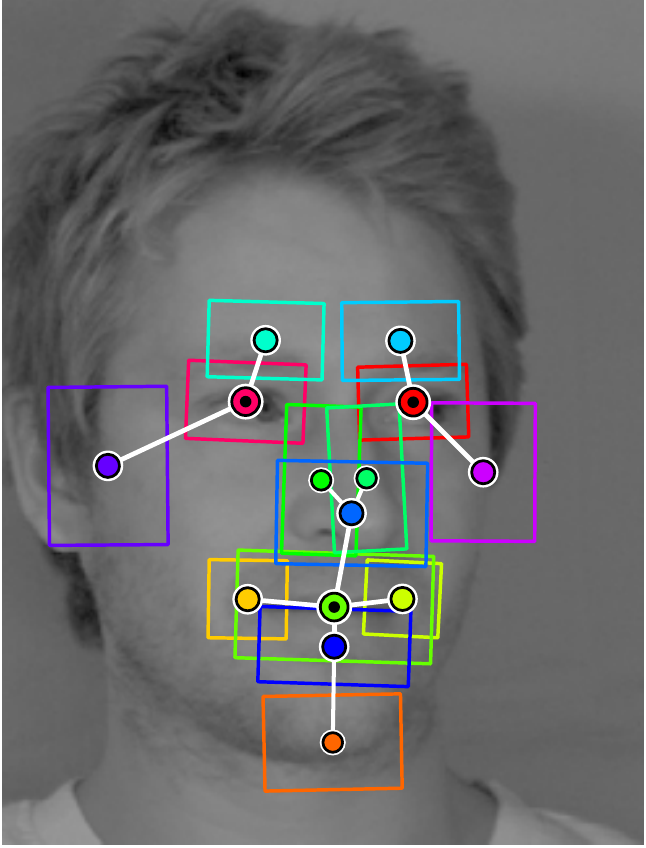}\hfill{}\includegraphics[width=0.27\columnwidth]{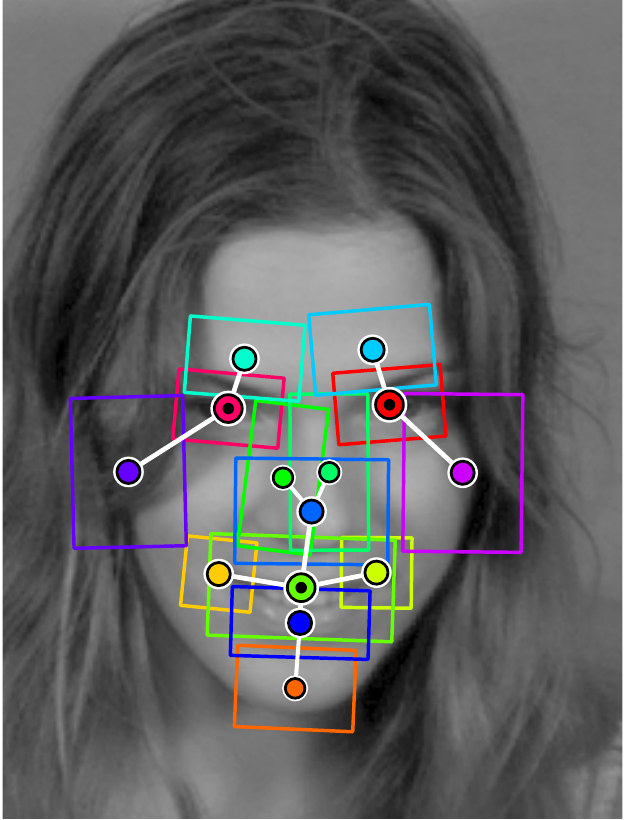}
\par\end{centering}

\protect\caption{\label{fig:CPA-instance}CPA model instances fitted to face images.
Each facial part is denoted by a box in a specific color. The circled
disk at its center denotes the associated tree node. The white solid
lines denote the tree edges. Bigger nodes correspond to higher levels
of the tree. The nodes with black dots are linked with the root node,
which is not displayed in the figure. }
\end{figure}

The above direct extension of \citep{andrew2012practical} essentially
aligns the $m$ parts independently. Unfortunately, it empirically
appears to be very unstable for facial part alignment. As shown in
Fig.~\ref{fig:independent-single}, the parts may often drift away
from their initialization to some more ``flat'' face regions. Indeed,
alignment by optimizing $\nu^{\langle i\rangle}$ in $y\circ\nu^{\langle i\rangle}$
is a non-convex procedure (as explained in more details in Section
\ref{sub:CPA-part}). Compared with an entire face, individual parts
contain less visual structure and thus alignment of them is prone
to meaningless local minima. To overcome this problem, we consider
in this paper incorporating some sort of global information of facial
structure to regularize the deformations of individual parts. In particular,
we are motivated by \citep{zhu2012dpmface} to use a tree-structured
shape model to constrain the difference of transformation parameters
of different parts. We write $g\left(\boldsymbol{\nu},\mathcal{Z}\right)$
for the regularization term determined by the tree-structured shape
model, where $\mathcal{Z}$ denotes model parameters in the form of
a tuple. The tree-structured shape model is illustrated in Fig.~\ref{fig:CPA-model}
i - iii and will be elaborated in Section \ref{sub:tree}. With consideration
of the tree-structured shape constraint, we revise (\ref{eq:CPA-constraint-0})
to formulate the alignment objective of our proposed CPA model as
\begin{gather}
\min_{\underset{i=1,2,\ldots,m}{x^{\langle i\rangle},e^{\langle i\rangle},\boldsymbol{\nu}}}\;\sum_{i}^{m}\lambda^{\langle i\rangle}\Vert e^{\langle i\rangle}\Vert_{1}+\eta g\left(\boldsymbol{\nu},\mathcal{Z}\right)\nonumber \\
\mathrm{s.t.}\quad y\circ\nu^{\langle i\rangle}=D^{\left\langle i\right\rangle }x^{\langle i\rangle}+e^{\langle i\rangle},\label{eq:CPA-constraint-1}
\end{gather}
where $\eta\in\mathbb{R}_{+}$ weights the regularization term. The
choice of $\{\lambda^{\langle i\rangle}\}_{i=1}^{m}$ will be addressed
when we learn the CPA model in Section \ref{sec:Model-learning}.
As shown in Fig.~\ref{fig:CPA-single}, our proposed CPA method can
produce very stable alignment results using the same part dictionaries
(Fig.~\ref{fig:dictionary-single}) as independent part alignment
does. In Fig.~\ref{fig:CPA-instance}, we give all the part instances
of the CPA model that are fitted to real face images. We note that
CPA is not designed to produce a seamless face constituted by deformed
individual parts, as AAMs \citep{cootes2001aam,AAMRevisited} can
do. Instead, it aims to establish correspondence of facial parts across
pose or expression so that face images with intra-subject variations
can be matched at the part level to improve face recognition.

\subsection{Tree-structured Shape Model \label{sub:tree}}

Let $(\mathcal{V},\mathcal{E})$ denote the tree, where $\mathcal{V}=\{0,1,2,\ldots,m\}$
is the vertex set, and $\mathcal{E}$ is the set of directed edges.
$\mathcal{V}$ is composed of the nodes of the $m$ facial parts and
a \emph{root node} (the node ``$0$'') corresponding to the holistic
face. Every $i^{\textrm{th}}$ node takes $\nu^{\langle i\rangle}$
as its variables corresponding to the transformation parameters of
the facial part. For the root node, we use $\nu^{\langle0\rangle}=(0,0,0,0)^{T}$
when assuming the face \emph{at the global level} has been registered
into the canonical form, which suggests $(u,v)=\nu^{\langle0\rangle}(u,v)$.
We consider this simplified case in the following derivation of the
tree-structured shape model.

In order to build a model on the tree, we first associate with each
edge the difference of transformation parameters of its two end nodes.
We then concretize $g\left(\boldsymbol{\nu},\mathcal{Z}\right)$ as
follows. In particular, for the edge $(j,i)\in\mathcal{E}$ with parent
node $j$ and child node $i$, we use a multivariate Gaussian distribution
with mean $\mu^{\langle i\rangle}\in\mathbb{R}^{d}$ and precision
$\Lambda^{\langle i\rangle}\in\mathbb{R}^{d\times d}$ to model the
transformation difference $\nu_{\delta}^{\langle i\rangle}=\nu^{\langle i\rangle}-\nu^{\langle j\rangle}\in\mathbb{R}^{d}$,
i.e., $p(\nu_{\delta}^{\langle i\rangle}|z^{\langle i\rangle})=\mathcal{N}(\nu_{\delta}^{\langle i\rangle};\mu^{\langle i\rangle},{\Lambda^{\langle i\rangle}}^{-1})$
with $z^{\langle i\rangle}=(\mu^{\langle i\rangle},\Lambda^{\langle i\rangle})$,
where the Gaussian probability distribution function (PDF) is 
\begin{equation}
\mathcal{N}(\nu_{\delta};\mu,\Lambda^{-1})=\frac{|\Lambda|^{1/2}}{(2\pi)^{d/2}}\exp\left\{ -\frac{1}{2}(\nu_{\delta}-\mu)^{T}\Lambda(\nu_{\delta}-\mu)\right\} .\label{eq:norm-PDF}
\end{equation}
Thus, let $\mathcal{Z}=(z^{\langle1\rangle},z^{\langle2\rangle},\ldots,z^{\langle m\rangle})$
be a tuple of Gaussian parameters.

Since the definition of $p(\nu_{\delta}^{\langle i\rangle}|z^{\langle i\rangle})$
is free of $\nu^{\langle j\rangle}$ for $(j,i)\in\mathcal{E}$, we
have essentially assumed the independence between $\nu_{\delta}^{\langle i\rangle}$
and $\nu^{\langle j\rangle}$, which suggests $p(\nu^{\langle i\rangle}|\nu^{\langle j\rangle},z^{\langle i\rangle})=p(\nu_{\delta}^{\langle i\rangle}|z^{\langle i\rangle})$.
Thus, we take our tree-structured shape model as a Bayesian network,
whose joint probability is 
\begin{align}
p(\{\nu^{\langle i\rangle}\}_{i=0}^{m}|\mathcal{Z}) & =p(\nu^{\langle0\rangle})\cdot\prod_{(j,i)\in\mathcal{E}}p(\nu^{\langle i\rangle}|\nu^{\langle j\rangle},z^{\langle i\rangle})\nonumber \\
 & =\prod_{i=1}^{m}p(\nu_{\delta}^{\langle i\rangle}|z^{\langle i\rangle}).\label{eq:joint-prob}
\end{align}
where $p(\nu^{\langle0\rangle})\equiv1$. Finally, taking the negative
logarithm of the joint probability gives the regularization term $g\left(\boldsymbol{\nu},\mathcal{Z}\right)$
in (\ref{eq:CPA-constraint-1}) as 
\begin{align}
g\left(\boldsymbol{\nu},\mathcal{Z}\right)= & -\ln p(\{\nu^{\langle i\rangle}\}_{i=0}^{m}|\mathcal{Z})\nonumber \\
= & \frac{1}{2}\sum_{i=1}^{m}(\nu_{\delta}^{\langle i\rangle}-\mu^{\langle i\rangle})^{T}\Lambda^{\langle i\rangle}(\nu_{\delta}^{\langle i\rangle}-\mu^{\langle i\rangle})+b,\label{eq:quadratic-cost}
\end{align}
where $b=\frac{dm}{2}\ln(2\pi)-\frac{1}{2}\sum_{i=1}^{m}\ln|\Lambda^{\langle i\rangle}|$
is the term independent of $\{\nu_{\delta}^{\langle i\rangle}\}_{i=1}^{m}$.

Similar tree-structured shape model and quadratic regularization term
were also used in \citep{zhu2012dpmface} for face detection and facial
landmark localization. However, they ignored the dependency among
variables of the difference of transformation parameters associated
with any pair of facial parts connected by an edge in the tree, which,
in our case, equals to constrain $\Lambda^{\langle i\rangle}$ to
be diagonal. Compared to \citep{zhu2012dpmface}, our shape constraint
is derived from a Bayesian network formulation, which not only interprets
the underlying probabilistic properties of tree-structured shape models,
but also enables maximum \emph{a posteriori} (MAP) estimation of the
model parameters $\mathcal{Z}$ in the training stage of CPA. We will
show later that the probabilistic prior introduced for MAP estimation
prevents the CPA training from degenerate solutions when image alignment
and tree-model learning are jointly formulated in an unsupervised
way. This scenario is different from \citep{zhu2012dpmface}, as their
tree-structured shape model is integrated in a classification problem,
which is strongly supervised.

In fact, parametrization of AAMs and CLMs \citep{CLM,SaragihCLM}
is also based on a joint Gaussian model, which seems to be similar
to (\ref{eq:quadratic-cost}) derived from a tree model. To see the
difference, if we concatenate the variables $\{\nu^{\langle i\rangle}\}_{i=1}^{m}$
as $\operatorname{vec}(\boldsymbol{\nu})$ and model it as a Gaussian
distribution, the corresponding precision matrix, denoted as $\boldsymbol{\Lambda}\in\mathbb{R}^{dm\times dm}$,
will have a block sparse structure with at most $d^{2}(m+|\{(j,i)\in\mathcal{E}:j\neq0\}|)$
non-zero entries. The assumption on the independence between $\nu_{\delta}^{\langle i\rangle}$
and $\nu^{\langle j\rangle}$ for $(i,j)\in\mathcal{E}$, which constrains
$\boldsymbol{\Lambda}$'s degree of freedom to be $d^{2}m$, distinguishes
the tree-structured shape model from a general joint Gaussian model
with block diagonal precision. 

Up to now we have assumed the face \emph{at the global level} has
been registered into the canonical form. In practice, however, observed
face images are not generally in this canonical form. Consequently,
the learned tree-structured shape model for a globally canonical face
will not apply directly. To mitigate this problem, we consider optimizing
an additional holistic transformation $\sigma$, so that the learned
tree-structured shape model can still be useful to constrain the optimization
of $\boldsymbol{\nu}$. The CPA objective (\ref{eq:CPA-constraint-1})
is then rewritten as 
\begin{gather}
\min_{\underset{i=1,2,\ldots,m}{x^{\langle i\rangle},e^{\langle i\rangle},\boldsymbol{\nu},\sigma}}\;\sum_{i}^{m}\lambda^{\langle i\rangle}\Vert e^{\langle i\rangle}\Vert_{1}+\eta g\left(\boldsymbol{\nu},\mathcal{Z}\right)\nonumber \\
\mathrm{s.t.}\quad y\circ\sigma\circ\nu^{\langle i\rangle}=D^{\left\langle i\right\rangle }x^{\langle i\rangle}+e^{\langle i\rangle}.\label{eq:CPA-constraint-2}
\end{gather}
We call $\sigma$ the \emph{holistic deformation}, and keep calling
$\nu^{\langle i\rangle}$ the $i^{\textrm{th}}$ \emph{part deformation}.

\subsection{Optimization \label{sub:CPA-Optimization}}

Given a learned tree-structured shape model, we present algorithms
to solve our CPA objective (\ref{eq:CPA-constraint-2}) in this subsection.
The main difficulty of solving (\ref{eq:CPA-constraint-2}) comes
from the non-convexity of its constraints $y\circ\sigma\circ\nu^{\langle i\rangle}=D^{\left\langle i\right\rangle }x^{\langle i\rangle}+e^{\langle i\rangle}$,
$i=1,\dots,m$, which couples the nonlinear operations of holistic
deformation $\sigma$ and part deformations $\{\nu^{\langle i\rangle}\}_{i=1}^{m}$
on the image domain. We choose the strategy of alternating optimization:
we first update $\{x^{\langle i\rangle}\}_{i=1}^{m}$, $\{e^{\langle i\rangle}\}_{i=1}^{m}$,
and $\boldsymbol{\nu}$ together while fixing $\sigma$, and we then
update $\sigma$ %
\footnote{Instead of updating $\sigma$, $\{x^{\langle i\rangle}\}_{i=1}^{m}$,
and $\{e^{\langle i\rangle}\}_{i=1}^{m}$ while fixing $\boldsymbol{\nu}$
in the second step, we propose a more efficient approach that jointly
updates $\sigma$ and $\boldsymbol{\nu}$ while keeping $\{x^{\langle i\rangle}\}_{i=1}^{m}$
and $\{e^{\langle i\rangle}\}_{i=1}^{m}$ fixed. Details of the approach
will be presented in Section \ref{sub:CPA-holistic}. %
}. These two steps are alternately applied until the algorithm converges.
Note that this alternating strategy requires relatively good initialization
of $\sigma$ and $\boldsymbol{\nu}$, so that the learned tree-structured
shape model can be effective to constrain the optimization of $\boldsymbol{\nu}$.
We defer the discussion of this issue in Section \ref{sub:CPA-InitializationIssue}
while assuming at this moment that the initial $\sigma$ and $\boldsymbol{\nu}$
are good enough to start the alternating process.

\subsubsection{Solving Part Deformations \label{sub:CPA-part}}

Given fixed $\sigma$, we update each $\nu^{\langle i\rangle}$ by
a generalization of the Gauss-Newton method. More specifically, for
a linear update from $\nu^{\langle i\rangle}$ to $\nu^{\langle i\rangle}+\Delta\nu^{\langle i\rangle}$,
the left-hand side of the equality constraint in (\ref{eq:CPA-constraint-2})
can be approximated by its first-order Taylor expansion at $\nu^{\langle i\rangle}$,
i.e., $y\circ\sigma\circ(\nu^{\langle i\rangle}+\Delta\nu^{\langle i\rangle})\approx y\circ\sigma\circ\nu^{\langle i\rangle}+J^{\langle i\rangle}\Delta\nu^{\langle i\rangle}$,
where $J^{\langle i\rangle}\doteq\partial(y\circ\sigma\circ\nu^{\langle i\rangle})/\partial\nu^{\langle i\rangle}$
is the Jacobian with respect to (w.r.t.) $\nu^{\langle i\rangle}$.
Let $\Delta\boldsymbol{\nu}=[\Delta\nu^{\langle1\rangle},\Delta\nu^{\langle2\rangle},\ldots,\Delta\nu^{\langle m\rangle}]$.
The above linearization leads to the following problem to optimize
$\{x^{\langle i\rangle}\}_{i=1}^{m},\{e^{\langle i\rangle}\}_{i=1}^{m},\Delta\boldsymbol{\nu}$
\begin{gather}
\min_{\underset{i=1,2,\ldots,m}{x^{\langle i\rangle},e^{\langle i\rangle},\Delta\boldsymbol{\nu}}}\;\sum_{i}^{m}\lambda^{\langle i\rangle}\Vert e^{\langle i\rangle}\Vert_{1}+\eta g\left(\boldsymbol{\nu}+\Delta\boldsymbol{\nu},\mathcal{Z}\right)\nonumber \\
\mathrm{s.t.}\quad y\circ\sigma\circ\nu^{\langle i\rangle}+J^{\langle i\rangle}\Delta\nu^{\langle i\rangle}=D^{\left\langle i\right\rangle }x^{\langle i\rangle}+e^{\langle i\rangle}.\label{eq:CPA-obj-linearized}
\end{gather}
We repeatedly solve the problem (\ref{eq:CPA-obj-linearized}) to
linearly update $\boldsymbol{\nu}$, until converging to a local minimum,
which gives the solution to the original problem (\ref{eq:CPA-constraint-2}).
Similar iterative techniques have also been used in related works
\citep{peng2011rasl,andrew2012practical}, and showed good behaviors
of convergence.


We solve the convex problem (\ref{eq:CPA-obj-linearized}) by adapting
the Augmented Lagrange Multiplier (ALM) method \citep{lin2009alm}.
Let 
\begin{align}
h( & x^{\langle i\rangle},e^{\langle i\rangle},\Delta\nu^{\langle i\rangle})=\nonumber \\
 & y\circ\sigma\circ\nu^{\langle i\rangle}+J^{\langle i\rangle}\Delta\nu^{\langle i\rangle}-D^{\left\langle i\right\rangle }x^{\langle i\rangle}-e^{\langle i\rangle}.
\end{align}
The augmented Lagrangian function for (\ref{eq:CPA-obj-linearized})
can be written as 
\begin{align}
 & L_{\beta}(\{x^{\langle i\rangle}\}_{i=1}^{m},\{e^{\langle i\rangle}\}_{i=1}^{m},\Delta\boldsymbol{\nu},\{\gamma^{\langle i\rangle}\}_{i=1}^{m})=\nonumber \\
 & \sum_{i}^{m}\biggl\{\lambda^{\langle i\rangle}\Vert e^{\langle i\rangle}\Vert_{1}+\left\langle \gamma^{\langle i\rangle},h(x^{\langle i\rangle},e^{\langle i\rangle},\Delta\nu^{\langle i\rangle})\right\rangle \begin{array}{c}
\\
\\
\end{array}\label{eq:CPA-augmented-fun}\\
 & +\frac{\beta}{2}\left\Vert h(x^{\langle i\rangle},e^{\langle i\rangle},\Delta\nu^{\langle i\rangle})\right\Vert _{F}^{2}\biggl\}+g\left(\boldsymbol{\nu}+\Delta\boldsymbol{\nu},\mathcal{Z}\right),\nonumber 
\end{align}
where $\{\gamma^{\langle i\rangle}\}_{i=1}^{m}$ are the Lagrange
multiplier vectors, $\left\Vert \cdot\right\Vert _{F}$ denotes matrix
Frobenius norm, and $\langle\cdot,\cdot\rangle$ denotes inner product
of vectors or matrices. Instead of directly solving the constrained
problem (\ref{eq:CPA-obj-linearized}), ALM searches for a saddle
point of (\ref{eq:CPA-augmented-fun}). Given initial $\{\gamma^{\langle i\rangle}\}_{i=1}^{m}$,
it iteratively and alternately updates $\{x^{\langle i\rangle}\}_{i=1}^{m},\{e^{\langle i\rangle}\}_{i=1}^{m},\Delta\boldsymbol{\nu}$
and $\{\gamma^{\langle i\rangle}\}_{i=1}^{m}$ by 
\begin{enumerate}
\item $\{x^{\langle i\rangle}\}_{i=1}^{m},\{e^{\langle i\rangle}\}_{i=1}^{m},\Delta\boldsymbol{\nu}\leftarrow$\vspace{-0.4em}

\begin{equation}
\mbox{\hspace{-2em}}\underset{\underset{i=1,2,\ldots,m}{x^{\langle i\rangle},e^{\langle i\rangle},\Delta\boldsymbol{\nu}}}{\arg\min}\, L_{\beta_{l}}(\{x^{\langle i\rangle}\}_{i=1}^{m},\{e^{\langle i\rangle}\}_{i=1}^{m},\Delta\boldsymbol{\nu},\{\gamma^{\langle i\rangle}\}_{i=1}^{m});\label{eq:CPA-alm-iter}
\end{equation}

\item $\gamma^{\langle i\rangle}\leftarrow\gamma^{\langle i\rangle}+\beta_{l}\cdot h(x^{\langle i\rangle},e^{\langle i\rangle},\Delta\nu^{\langle i\rangle})$,
\\
 for $i=1,2.\ldots,m$; 
\end{enumerate}
where $l$ is the iteration number, and $\{\beta_{l}\}_{l=1,2,\ldots}$
is an increasing sequence. ALM is proven to converge to the optimum
of the original problem as $\beta_{l}$ becomes sufficient large \citep{bertsekas1999nonlinear}.

It is still difficult to directly solve (\ref{eq:CPA-alm-iter}) w.r.t.\ all
the three groups of variables: $\{x^{\langle i\rangle}\}_{i=1}^{m}$,
$\{e^{\langle i\rangle}\}_{i=1}^{m}$, and $\Delta\boldsymbol{\nu}$.
Instead, we consider updating them in an alternating manner. And it
turns out that each subproblem associated with any one group of variables
has a closed form solution. More specifically, let
\begin{equation}
\mathcal{S}_{\alpha}(x)=\operatorname{sign}(x)\cdot\max\{|x|-\alpha,0\}\label{eq:soft-thresholding}
\end{equation}
be the soft thresholding operator. For $i=1,2,\ldots,m$, we update
$x^{\langle i\rangle}$ and $e^{\langle i\rangle}$ sequentially by
\begin{align*}
r^{\langle i\rangle}\leftarrow\; & y^{\langle i\rangle}\circ\sigma\circ\nu^{\langle i\rangle}+(1/\beta_{l})\gamma^{\langle i\rangle},\\
x^{\langle i\rangle}\leftarrow\; & \left({D^{\langle i\rangle}}^{T}D^{\langle i\rangle}\right)^{-1}{D^{\langle i\rangle}}^{T}(r^{\langle i\rangle}+J^{\langle i\rangle}\Delta\nu^{\langle i\rangle}-e^{\langle i\rangle}),\\
e^{\langle i\rangle}\leftarrow\; & \mathcal{S}_{\lambda^{\langle i\rangle}/\beta_{l}}(r^{\langle i\rangle}+J^{\langle i\rangle}\Delta\nu^{\langle i\rangle}-D^{\langle i\rangle}x^{\langle i\rangle}),
\end{align*}
where $r^{\langle i\rangle}$ is an auxiliary variable used for notation
convenience only. Then, the optimum $\Delta\boldsymbol{\nu}$ can
be found by solving 
\begin{align}
\mathbf{0}= & \frac{\partial L_{\mu}(\{x^{\langle i\rangle}\}_{i=1}^{m},\{e^{\langle i\rangle}\}_{i=1}^{m},\Delta\boldsymbol{\nu},\{\gamma^{\langle i\rangle}\}_{i=1}^{m})}{\partial\Delta\nu^{\langle i\rangle}}\\
= & \beta_{l}\left(r^{\langle i\rangle}+J^{\langle i\rangle}\Delta\nu^{\langle i\rangle}-D^{\langle i\rangle}x^{\langle i\rangle}-e^{\langle i\rangle}\right){J^{\langle i\rangle}}\nonumber \\
 & +\eta\frac{\partial g(\boldsymbol{\nu}+\Delta\boldsymbol{\nu},\mathcal{Z})}{\partial\Delta\nu^{\langle i\rangle}}.
\end{align}
for $i=1,2,\ldots,m$. Let $\{\varsigma_{i}\}_{i=1}^{m}$ denote the
standard basis of $\mathbb{R}^{m}$. The $m$ equations together form
a system about $\Delta\boldsymbol{\nu}$ with the following form:
\begin{equation}
\sum_{i=1}^{m}G^{\langle i\rangle}\Delta\boldsymbol{\nu}\varsigma_{i}\varsigma_{i}^{T}+\eta\frac{\partial g(\boldsymbol{\nu}+\Delta\boldsymbol{\nu},\mathcal{Z})}{\partial\Delta\boldsymbol{\nu}^{T}}=Q,\label{eq:dt-linear-system}
\end{equation}
where $G^{\langle i\rangle}\in\mathbb{R}^{d\times d}$, and $Q\in\mathbb{R}^{d\times m}$.
Here, $G^{\langle i\rangle}=\beta_{l}{J^{\langle i\rangle}}^{T}J^{\langle i\rangle}$,
and the $i^{\textrm{th}}$ column of $Q$ is $\beta_{l}{J^{\langle i\rangle}}^{T}(D^{\langle i\rangle}x^{\langle i\rangle}+e^{\langle i\rangle}-r^{\langle i\rangle})$.
As $g(\boldsymbol{\nu}+\Delta\boldsymbol{\nu},\mathcal{Z})$ has the
form of the summation of the quadratics of $\{\Delta\nu^{\langle i\rangle}\}_{i=1}^{m}$,
(\ref{eq:dt-linear-system}) is a linear system. It is usually sparse
due to the tree structure of the shape model. Hence, even a large
number of facial parts are present, we can still simultaneously and
efficiently solve $\{\Delta\nu^{\langle i\rangle}\}_{i=1}^{m}$ with
high precision. We describe the expanded form of (\ref{eq:dt-linear-system})
in Appendix~\ref{sec:DT-Linear-System}.

Instead of solving (\ref{eq:CPA-alm-iter}) exactly by converged iterations
of alternating optimization of $\{x^{\langle i\rangle}\}_{i=1}^{m},\{e^{\langle i\rangle}\}_{i=1}^{m}$
and $\Delta\boldsymbol{\nu}$, we use inexact ALM that updates the
three groups of variables alternately for only once in each iteration
of the ALM method. Compared to exact ALM, inexact ALM shows better
practical performance in terms of optimization efficiency \citep{lin2009alm}.
In summary, we solve (\ref{eq:CPA-constraint-2}) by repeatedly solving
the linearized problem (\ref{eq:CPA-obj-linearized}), which itself
can be efficiently solved by inexact ALM.

\subsubsection{Solving Holistic Deformation \label{sub:CPA-holistic}}

Given fixed $\boldsymbol{\nu}$ in (\ref{eq:CPA-constraint-2}), the
most straightforward way to optimize $\sigma$, together with $\{x^{\langle i\rangle}\}_{i=1}^{m}$
and $\{e^{\langle i\rangle}\}_{i=1}^{m}$, is to use similar techniques
as in Section \ref{sub:CPA-part}, i.e., linearizing the constraint
in (\ref{eq:CPA-constraint-2}) w.r.t.\ $\sigma$ and using ALM to
solve the linearized problem. In this paper, we propose an alternative
approach that involves the variables $\sigma$ and $\boldsymbol{\nu}$
only, without performing the actual image deformation, and is thus
much more efficient compared to the aforementioned standard approach.
More specifically, given fixed $\{x^{\langle i\rangle}\}_{i=1}^{m}$
and $\{e^{\langle i\rangle}\}_{i=1}^{m}$ obtained in the previous
iteration of part deformations, the right-hand side of the equality
constraint in (\ref{eq:CPA-constraint-2}) is also fixed. If we keep
it unchanged in the update of $\sigma$, it indicates that $\sigma\circ\nu^{\langle i\rangle}$
for a combined deformation will not change. Denote $\zeta^{\langle i\rangle}=\sigma\circ\nu^{\langle i\rangle}$
for this fixed combined deformation, we propose to update $\sigma$
and $\boldsymbol{\nu}$ jointly by optimizing 
\begin{equation}
\min_{\boldsymbol{\nu},\sigma}g\left(\boldsymbol{\nu},\mathcal{Z}\right)\quad\mathrm{s.t.}\;\sigma\circ\nu^{\langle i\rangle}=\zeta^{\langle i\rangle},\label{eq:CPA-obj-holistic-2}
\end{equation}
which does not involve any actual image deformation and can thus be
efficiently solved. For similarity transformation considered in this
paper, we present in Appendix \ref{sec:solving-transforms} the associated
parametrization of (\ref{eq:CPA-obj-holistic-2}) and the optimization
algorithm.

Algorithm \ref{alg:outer-2} gives the outer loop of the algorithms
presented in Sections \ref{sub:CPA-part} and \ref{sub:CPA-holistic}
for the proposed constrained part-based alignment.

\begin{algorithm}
\begin{algorithmic}[1]

\State Initialize $\boldsymbol{\nu}$ and $\sigma$

\While{NOT converged}

\State  \label{algln:outer-opt2-partial} $\boldsymbol{\nu}$$\leftarrow\underset{\boldsymbol{\nu}}{\arg\min}\underset{\underset{i=1,2,\ldots,m}{x^{\langle i\rangle},e^{\langle i\rangle}}}{\min}\sum_{i}^{m}\Vert e^{\langle i\rangle}\Vert_{1}+\eta g\left(\boldsymbol{\nu},\mathcal{Z}\right)$
s.t. (\ref{eq:CPA-constraint-2})

\State $\zeta^{\langle i\rangle}\leftarrow\sigma\circ\nu^{\langle i\rangle}$

\State \label{algln:outer-opt2-holistic} $\boldsymbol{\nu},\sigma$$\leftarrow\underset{\boldsymbol{\nu},\sigma}{\arg\min}\; g\left(\boldsymbol{\nu},\mathcal{Z}\right)$
s.t. $\sigma\circ\nu^{\langle i\rangle}=\zeta^{\langle i\rangle}$

\EndWhile

\end{algorithmic}

{\footnotesize 

}

\protect\caption{Outer loop for CPA -- Default Option \label{alg:outer-2}}
\end{algorithm}

\subsubsection{Initializing Holistic and Part Deformations\label{sub:CPA-InitializationIssue}}

Algorithms in Sections \ref{sub:CPA-part} and \ref{sub:CPA-holistic}
require relatively good initialization of $\sigma$ and $\boldsymbol{\nu}$.
In practice, we rely on off-the-shelf face detectors \citep{viola2001adaboost,zhu2012dpmface},
which provide a rough bounding box of the face or the holistic pose
and locations of facial parts \citep{zhu2012dpmface}. We initialize
$\sigma$ using the bounding box provided by face detectors. If locations
of facial parts are available, we also use them to initialize $\boldsymbol{\nu}$.
Otherwise, we choose the initial $\boldsymbol{\nu}$ as the part deformations
that maximize the likelihood of the tree-structured shape model. This
initialization acts as the average template of the shape constraint,
which is independent of any probe image.

The initialization based on face detectors is generally not accurate
enough for face recognition. Fortunately, in most cases they are good
enough for automatic alignment. Fig.~\ref{fig:CPA-instance} present
examples that illustrate the converged solutions of our CPA algorithm.

In some cases when face detectors give worse outputs, our CPA algorithm
might take a longer time to converge or might converge to an inconvenient
solution. To handle this situation, we use the holistic alignment
method of \citet{andrew2012practical} to initialize the holistic
transformation $\sigma$. Their method is not able to align non-frontal
faces well due to its holistic planar surface assumption, however,
it is good enough to be used as our initialization.

\section{CPA based Face Recognition \label{sec:Recognition}}

\begin{figure*}
\begin{centering}
\includegraphics[width=1\textwidth]{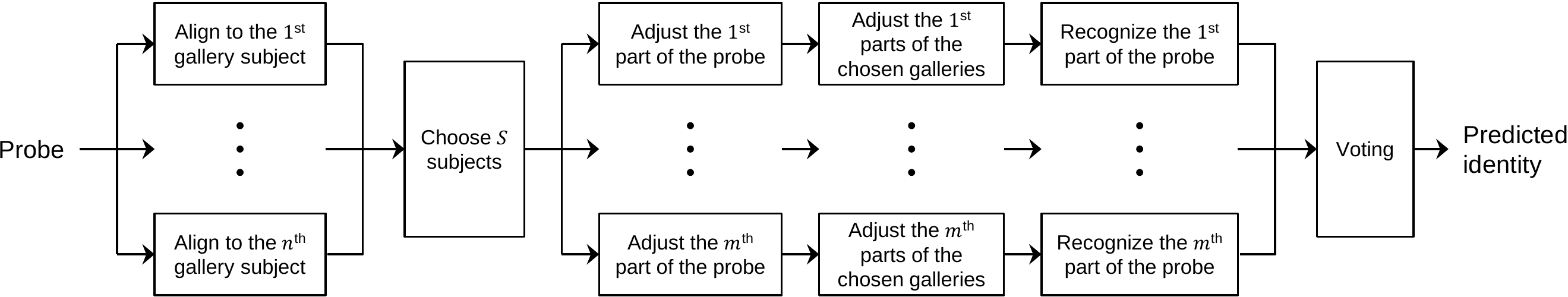}
\par\end{centering}

\protect\caption{The framework of the recognition algorithm in accordance with Algorithm~\ref{alg:recognition}
\label{fig:recognition-framework}}
\end{figure*}

In this section, we propose a CPA based face recognition method. Suppose
there are $N$ subjects in the gallery set, and each of them has multiple
face images. Suppose these face images have been aligned at the part
level to form the part dictionaries, which are denoted as $D^{\langle i\rangle}=[D_{1}^{\langle i\rangle},\dots,D_{N}^{\langle i\rangle}]$
for $i=1,\dots,m$. Given a probe face $y$, CPA suggests aligning
facial parts of $y$ to $\{D^{\langle i\rangle}\}_{i=1}^{m}$ by solving
(\ref{eq:CPA-constraint-2}) for a part-based face recognition. However,
similar to the situation of holistic face alignment in \citep{andrew2012practical},
the presence of facial parts of multiple subjects in $\{D^{\langle i\rangle}\}_{i=1}^{m}$
makes (\ref{eq:CPA-constraint-2}) have many local minima, corresponding
to aligning $y$ to the facial parts of different subjects. Instead,
one can perform CPA in a subject-wise manner by optimizing 
\begin{gather}
\min_{\underset{i=1,2,\ldots,m}{x_{s}^{\langle i\rangle},e_{s}^{\langle i\rangle},\boldsymbol{\nu}_{s},\sigma_{s}}}\;\sum_{i}^{m}\lambda^{\langle i\rangle}\Vert e_{s}^{\langle i\rangle}\Vert_{1}+\eta g\left(\boldsymbol{\nu}_{s},\mathcal{Z}\right)\nonumber \\
\mathrm{s.t.}\quad y\circ\sigma_{s}\circ\nu_{s}^{\langle i\rangle}=D_{s}^{\left\langle i\right\rangle }x_{s}^{\langle i\rangle}+e_{s}^{\langle i\rangle},\label{eq:CPA-SubjectWise}
\end{gather}
where $\boldsymbol{\nu}_{s}=[\nu_{s}^{\langle1\rangle},\dots,\nu_{s}^{\langle m\rangle}]$,
$\sigma_{s}$, and $\{e_{s}^{\langle i\rangle}\}_{i=1}^{m}$ are variables
of part deformations, holistic deformation, and the alignment residuals
w.r.t.\ the $s^{\textrm{th}}$ subject respectively. After solving
(\ref{eq:CPA-SubjectWise}) for all the $N$ subjects, for every $i^{\textrm{th}}$
part we sort the subject-wise alignment residuals $\{\Vert e_{s}^{\langle i\rangle}\Vert_{1}\}_{s=1}^{N}$
and select the top $C$ subjects with the smallest alignment residuals.
Part dictionaries of these $C$ selected subjects are then put together
to form a pruned gallery of facial parts. To make the facial parts
in the pruned gallery all aligned to the $i^{\textrm{th}}$ part of
$y$, we transform $D_{s}^{\langle i\rangle}$ by $(\sigma_{s}\circ\nu_{s}^{\langle i\rangle})^{-1}$
instead of transforming $y$ by $\sigma_{s}\circ\nu_{s}^{\langle i\rangle}$
for each selected subject $s$. For part-based face recognition, many
existing methods such as SRC \citep{john2009src}, LBP \citep{pami2006lbp}
and LDA \citep{pami1997fisherface} can be used at the part level,
based on the pruned gallery. Final recognition can be performed by
aggregating the part-level decisions using the basic plurality voting
scheme.

For obtaining better performance, we may adapt advanced aggregating
schemes, such as the kernelized plurality voting \citep{kumar2011kernel-plurality},
and joint recognition method for multiple observations, such as the
joint dynamic sparse representation \citep{zhang2011joint-dynamic}.
Nonetheless, we consider only the basic choice to keep our work from
obesity.

Algorithm \ref{alg:recognition} gives a summary of our proposed CPA
based face recognition method. An illustrative procedure with $6$
sequential modules is also shown in Fig.~\ref{fig:recognition-framework}.
Note that there are a few technical details in Algorithm~\ref{alg:recognition}
that can make a difference to recognition performance. In particular,
as the selected subjects are probably inconsistent for different parts,
we combine them together to substitute the selected subject set for
each part (Line~\ref{algln:overall-selected-subj}) so that the groundtruth
subject has a higher possibility to be included for part recognition.
Further, to control the number of overall selected subjects, we set
$C$ to the smallest integer making the pruned gallery size no less
than a given parameter $P$ (Line~\ref{algln:overall-selected-subj}).
We also adjust the $i^{\textrm{th}}$ part of $y$ by the averaged
transform of the $i^{\textrm{th}}$ part dictionaries of the selected
subjects before aligning them to it (Line~\ref{algln:recog-average-deform},\ref{algln:recog-spatial-adjust}).

\begin{algorithm}
\begin{algorithmic}[1]

\Require Subject-wise part dictionaries $\left\{ \{D_{s}^{\langle i\rangle}\}_{i=1}^{m}\right\} _{s=1}^{N}$
for the $m$ parts and $N$ subjects; the transformation group $\mathbb{G}$;
the parameter tuple $\mathcal{Z}$ of tree-structured shape model
; the probe image $y$; and, the chosen subject number $P$

\Ensure  $\operatorname{identity}(y)$

\Comment  Subject-wise alignment

\For{each subject $s$}

\State Align $y$ by CPA using $\{D_{s}^{\langle i\rangle}\}_{i=1}^{m}$, 

and get $\{e_{s}^{\left\langle i\right\rangle }\}_{i=1}^{m},\{\nu_{s}^{\langle i\rangle}\}_{i=1}^{m},\sigma_{s}$

\For{each part $i$ }

\State $\zeta_{s}^{\langle i\rangle}\leftarrow\nu_{s}^{\langle i\rangle}\circ\sigma_{s}$

\EndFor

\EndFor

\Comment  Pruning subjects

\For{each part $i$} 

\State  Sort subjects in ascending order according to $\{\Vert e_{s}^{\left\langle i\right\rangle }\Vert_{1}\}_{s=1}^{N}$, 

and get the orders $s_{1}^{\langle i\rangle},s_{2}^{\langle i\rangle},\ldots,s_{N}^{\langle i\rangle}$ 

\EndFor

\For{$j=1,2,\ldots,N$} 

\State  $\mathcal{B}_{j}\leftarrow\bigcup_{i=1}^{m}\{s_{l}^{\langle i\rangle}:l=1,2\ldots,j\}$

\EndFor

\State  $C\leftarrow j$ s.t. $|\mathcal{B}_{j-1}|<P\leq|\mathcal{B}_{j}|$
\label{algln:overall-selected-subj}

\Comment Part-wise recognition

\For{ $i=1,2,\ldots,m$ }

\State $\bar{\zeta}^{\langle i\rangle}\leftarrow\operatorname{mean}\{\zeta_{s}^{\langle i\rangle}:s=s_{1}^{\langle i\rangle},s_{2}^{\langle i\rangle},\ldots,s_{C}^{\langle i\rangle}\}$
\label{algln:recog-average-deform}

\State  $\hat{y}^{\langle i\rangle}\leftarrow y\circ\bar{\zeta}^{\langle i\rangle}$\label{algln:recog-spatial-adjust}

\For{each $s\in\mathcal{B}_{C}$ }

\State $\hat{\zeta}_{s}^{\langle i\rangle}\leftarrow\zeta_{s}^{\langle i\rangle}\circ{(\bar{\zeta}^{\langle i\rangle})}^{-1}$\label{algln:recog-spatial-adjust-ext}

\State $\hat{D}_{s}^{\langle i\rangle}\leftarrow D_{s}^{\langle i\rangle}\circ\zeta_{s}^{\langle i\rangle}$

\EndFor

\State  Use an existing recognition method to recognize $\hat{y}^{\langle i\rangle}$ 

with the pruned gallery $\{\hat{D}_{s}^{\langle i\rangle},s\}_{s\in\mathcal{B}_{C}}$, 

and get the predicted label $id^{\langle i\rangle}$

\EndFor

\Comment Decision by voting

\State  $\operatorname{identity}(y)$$\leftarrow$the label occurs
the most times in $\{id^{\langle i\rangle}\}_{i=1}^{m}$ 

\end{algorithmic}

\protect\caption{CPA based face recognition\label{alg:recognition}}
\end{algorithm}

\section{Model Learning for CPA \label{sec:Model-learning}}

Algorithms in the preceding sections assume that the CPA model, i.e.,
the tree-structured shape model and its associated part dictionaries,
have been given. In this section, we present how to learn them from
a training gallery of face images. Note that each part dictionary
is composed of a set of registered facial parts of gallery images.
For some facial parts, e.g., those around the chin position, it is
even difficult to manually annotate facial landmarks in order to align
them with high precision. To realize the automatic alignment of facial
parts to form the part dictionaries, we propose an approach that couples
the learning of a tree-structured shape model and that of part dictionaries.
Indeed, the relational constraints among deformations of different
parts provided by the tree-structured shape model make alignment of
part dictionaries become stable, and the deformation parameters of
facial parts of the training images also give statistical evidence
to learn the parameters $\mathcal{Z}$ in the CPA model. Learning
them in a coupled way is thus a natural choice. In the following,
before presenting algorithmic details of this coupled CPA model learning,
we first present how to align and form the part dictionaries given
the constraint from a known tree-structured shape model, which will
serve as one of the alternating steps in the coupled CPA model learning
algorithm.

\subsection{Learning Part Dictionaries with Constraint from Known Tree-structured
Shape Model}

\begin{figure}
\begin{centering}
\subfloat[\label{fig:batch-independent} Independent learning by \citet{peng2011rasl}'s
method at 20 iterations]{\begin{centering}
\begin{minipage}[t]{1\columnwidth}%
\begin{center}
\includegraphics[width=0.27\columnwidth]{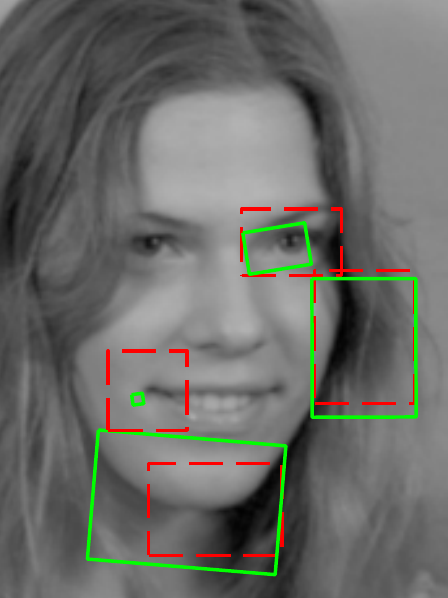}\hspace{0.08\columnwidth}\includegraphics[width=0.27\columnwidth]{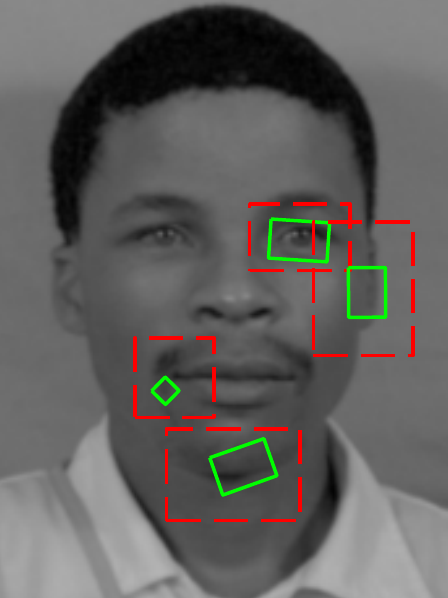}\hspace{0.08\columnwidth}\includegraphics[width=0.27\columnwidth]{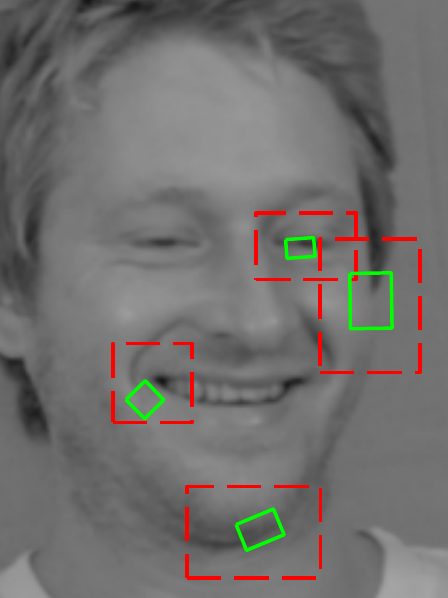}
\par\end{center}

\medskip{}

\begin{center}
\includegraphics[width=1\columnwidth]{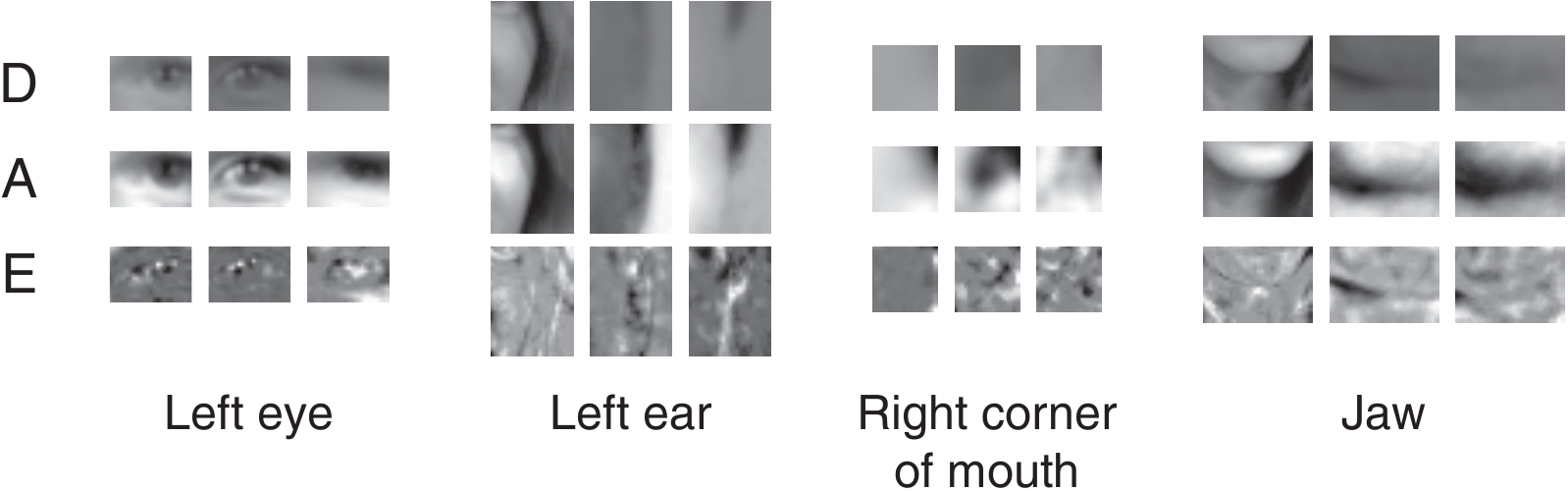}
\par\end{center}%
\end{minipage}
\par\end{centering}

}
\par\end{centering}

\begin{centering}
\subfloat[\label{fig:batch-CPA} Learning with a known shape constraint at the
converged stage. ]{\begin{centering}
\begin{minipage}[t]{1\columnwidth}%
\begin{center}
\includegraphics[width=0.27\columnwidth]{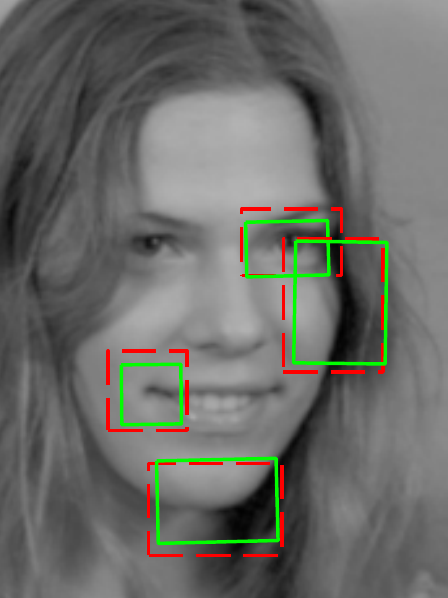}\hspace{0.08\columnwidth}\includegraphics[width=0.27\columnwidth]{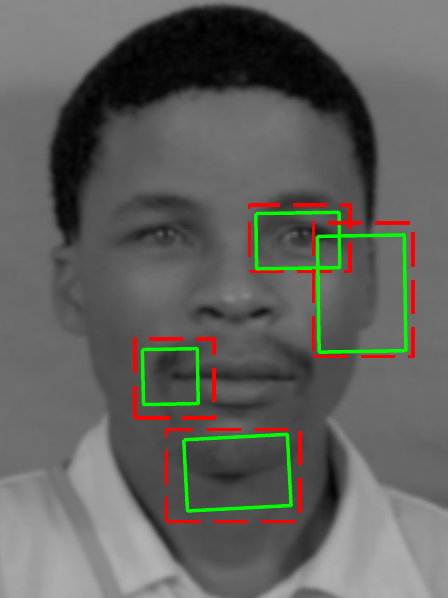}\hspace{0.08\columnwidth}\includegraphics[width=0.27\columnwidth]{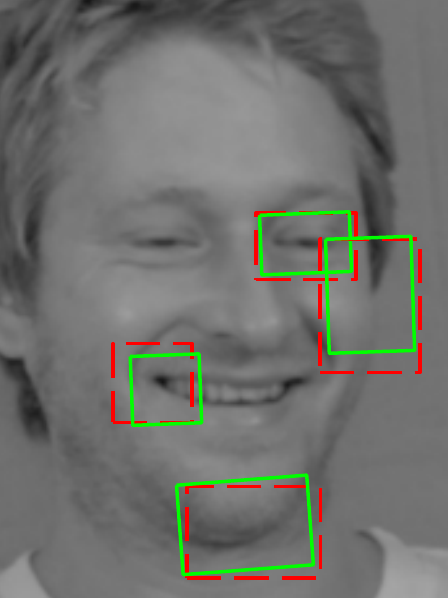}
\par\end{center}

\medskip{}

\begin{center}
\includegraphics[width=1\columnwidth]{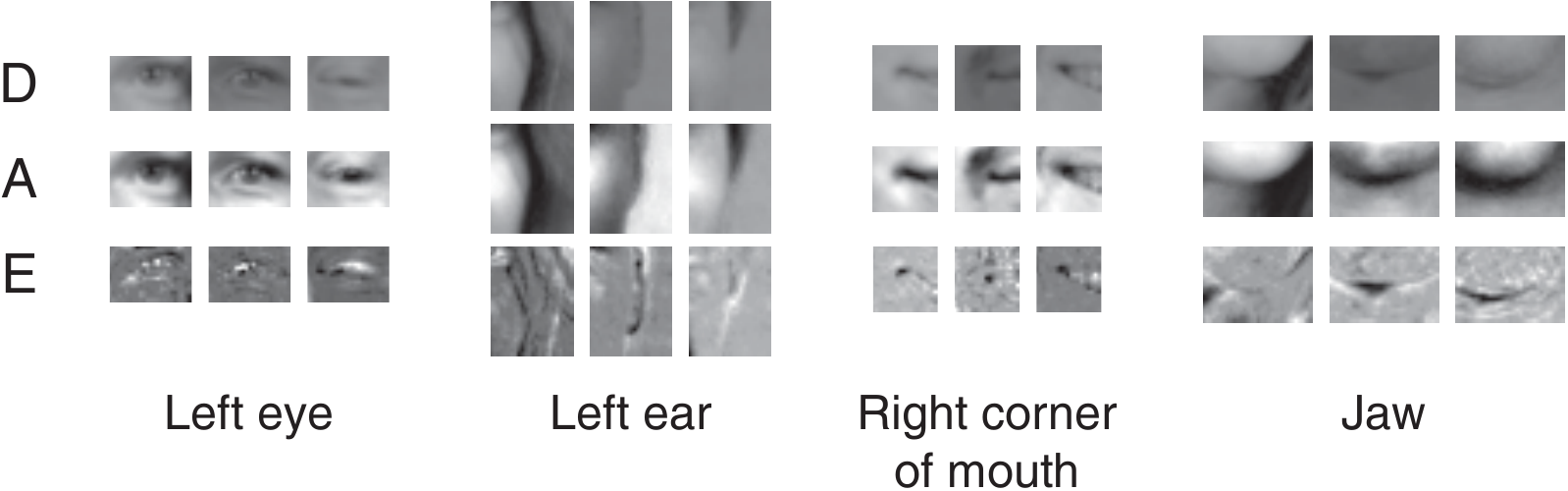}
\par\end{center}%
\end{minipage}
\par\end{centering}

}
\par\end{centering}

\protect\caption{Part dictionary learning. Parts are initialized at dashed red boxes,
and aligned in batch. The solid green boxes denote the obtained domains
of the parts. $D$ is a learned part dictionary (images cropped from
the obtained domains), and $A+E$ is its best low-rank and sparse
decomposition, where $A$ is the low-rank component, and $E$ is the
sparse error. The cropped images are normalized in terms of intensity
for display convenience. }
\end{figure}

We learn the part dictionaries from a training gallery of $n$ face
images, denoted as $D=[d_{1},d_{2},\ldots,d_{n}]$. Every $k^{\textrm{th}}$
column $d_{k}$ in $D$ represents a stacked vector form of face image
in the gallery. Learning part dictionaries is concerned with optimizing
the $d$-dimensional $\{\sigma_{k}\in\mathbb{G}\}_{k=1}^{n}$ for
holistic deformations and $\{\nu_{k}^{\langle i\rangle}\in\mathbb{G}\}_{k=1}^{n}$,
$i=1,\dots,m$, for part deformations, so that after part-based alignment
every $i^{\textrm{th}}$ part dictionary, denoted as $D^{\langle i\rangle}=[d_{1}^{\langle i\rangle},d_{2}^{\langle i\rangle},\ldots,d_{n}^{\langle i\rangle}]$
with $d_{k}^{\langle i\rangle}\doteq d_{k}\circ\sigma_{k}\circ\nu_{k}^{\langle i\rangle}$,
contains facial parts that have been registered into some canonical
form, as shown in Fig.~\ref{fig:batch-CPA}.

We observe that the registered facial parts in each part dictionary
correspond to appearance of different subjects at the same facial
region. These registered facial parts will ideally resemble each other
if nuisances due to inter-subject, illumination%
\footnote{If sufficient number of illumination conditions exist, illumination
variations may not be classified as a nuisance in that they can be
linearly modeled for low-rankness. %
}, expression, and/or pose variations can be decomposed out. In other
words, the part dictionary $D^{\langle i\rangle}$ would be low-rank
after decomposing out the aforementioned nuisances, which can be modeled
as a sparse error matrix. This sparse and low-rank decomposition was
used by \citet{peng2011rasl} to align a batch of linearly correlated
images, such as frames in a video sequence or face images of a same
subject. Motivated by \citep{peng2011rasl}, in this paper, we leverage
the very similar low-rank (\emph{matrices of part dictionaries without
nuisance}s) and sparsity (\emph{error matrices modeling various variations})
properties to align and form the part dictionaries. However, if we
directly apply techniques of \citet{peng2011rasl} to perform independent
part-based alignment, the alignment process often converges to less
meaningful solutions as shown in Fig.~\ref{fig:batch-independent}.
The reason is similar to that causing failure of applying the method
of \citet{andrew2012practical} in the case of aligning parts of a
probe face to the gallery set, as we explained in Section \ref{sec:CPA}.
Instead, we constrain part-based alignment for learning part dictionaries
by the tree-structured shape model. .

To simplify the notations, we write $\boldsymbol{\sigma}=[\sigma_{1},\ldots,\sigma_{m}]\in\mathbb{R}^{d\times n}$,
and combine $\{\nu_{k}^{\langle i\rangle}\}_{k=1,2,\ldots,n}^{i=1,2,\ldots,m}$
into a third-order tensor $\check{\boldsymbol{\nu}}\in\mathbb{R}^{d\times m\times n}$.
We define two operators that apply on $\check{\boldsymbol{\nu}}$
as 
\begin{align*}
T^{\langle i\rangle}(\check{\boldsymbol{\nu}})=\left[\nu_{1}^{\langle i\rangle},\nu_{2}^{\langle i\rangle},\ldots,\nu_{n}^{\langle i\rangle}\right] & \in\mathbb{R}^{d\times n},\\
T_{k}(\check{\boldsymbol{\nu}})=\left[\nu_{k}^{\langle1\rangle},\nu_{k}^{\langle2\rangle},\ldots,\nu_{k}^{\langle m\rangle}\right] & \in\mathbb{R}^{d\times m},
\end{align*}
where $T^{\langle i\rangle}(\cdot)$ extracts deformation parameters
of the $i^{\textrm{th}}$ part of the $n$ face images, and $T_{k}(\cdot)$
extracts those of $m$ facial parts of the $k^{\textrm{th}}$ image.
We then write $D^{\langle i\rangle}\doteq D\circ\boldsymbol{\sigma}\circ T^{\langle i\rangle}(\check{\boldsymbol{\nu}})$,
where ``$\circ$'' applies column-wisely. With these definitions
our objective for learning part dictionaries can be written as 
\begin{align*}
\min_{\underset{i=1,2,\ldots,m}{A^{\langle i\rangle},E^{\langle i\rangle},\check{\boldsymbol{\nu}},\boldsymbol{\sigma}}} & \Biggl\{\sum_{i=1}^{m}\left(\Vert A^{\langle i\rangle}\Vert_{*}+\lambda^{\langle i\rangle}\Vert E^{\langle i\rangle}\Vert_{1}\right)\\
 & +\eta\sum_{k=1}^{n}g\left(T_{k}(\check{\boldsymbol{\nu}}),\mathcal{Z}\right)\Biggl\}
\end{align*}
\begin{gather}
\mathrm{s.t.}\quad D\circ\boldsymbol{\sigma}\circ T^{\langle i\rangle}(\check{\boldsymbol{\nu}})=A^{\langle i\rangle}+E^{\langle i\rangle},\label{eq:mpRASL-obj}
\end{gather}
where $\Vert\cdot\Vert_{*}$ is the nuclear norm, which is a convex
surrogate function of matrix rank. As shown by \citet{peng2011rasl,acm2011rpca},
the penalty parameters $\{\lambda^{\langle i\rangle}\}_{i=1}^{m}$
can be set as the reciprocal of the square root of $E^{\langle i\rangle}$'s
row number, denoted as $\omega^{\langle i\rangle}$. In this paper,
we set $\lambda^{\langle i\rangle}=\hat{\lambda}/\sqrt{\omega^{\langle i\rangle}}$
with the constant $\hat{\lambda}=1$. We accordingly set $\eta=\hat{\eta}\cdot\sum_{i=1}^{m}(\omega^{\langle i\rangle})^{-1/2}$
with $\hat{\eta}$ chosen as $0.02$. To solve (\ref{eq:mpRASL-obj}),
we use a similar alternating strategy as in Section \ref{sub:CPA-Optimization}.
Appendix~\ref{sec:mpBatch-solution} gives details of the algorithmic
procedure. It requires the initialization of $\check{\boldsymbol{\nu}}$,
which can be obtained using the same method as in Session~\ref{sub:CPA-InitializationIssue}.
Fig.~\ref{fig:batch-CPA} shows example images of learned part dictionaries
by solving (\ref{eq:mpRASL-obj}). Compared with independent part-based
alignment (example results shown in Fig.~\ref{fig:batch-independent}),
the proposed alignment method with the constraint from a tree-structured
model gives much more meaningful results.

\subsection{Learning the Tree-structured Shape Model Jointly with Part Dictionaries}

\begin{figure*}
\begin{centering}
\subfloat[CPA learning with no prior for the tree-structured shape model \label{fig:learning-without-prior}]{\begin{centering}
\includegraphics[width=0.47\textwidth]{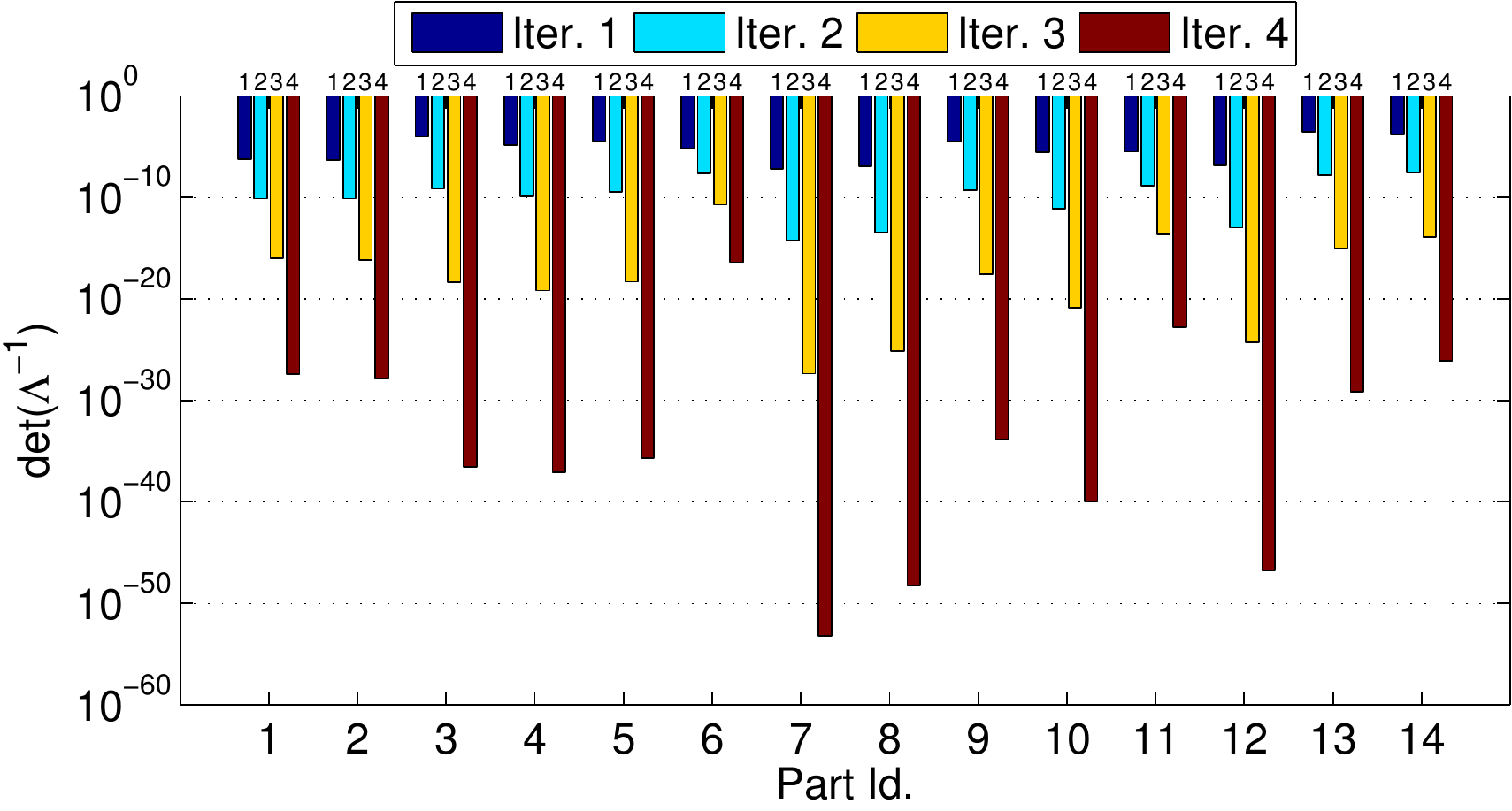}\hspace{0.03\textwidth}\includegraphics[width=0.47\textwidth]{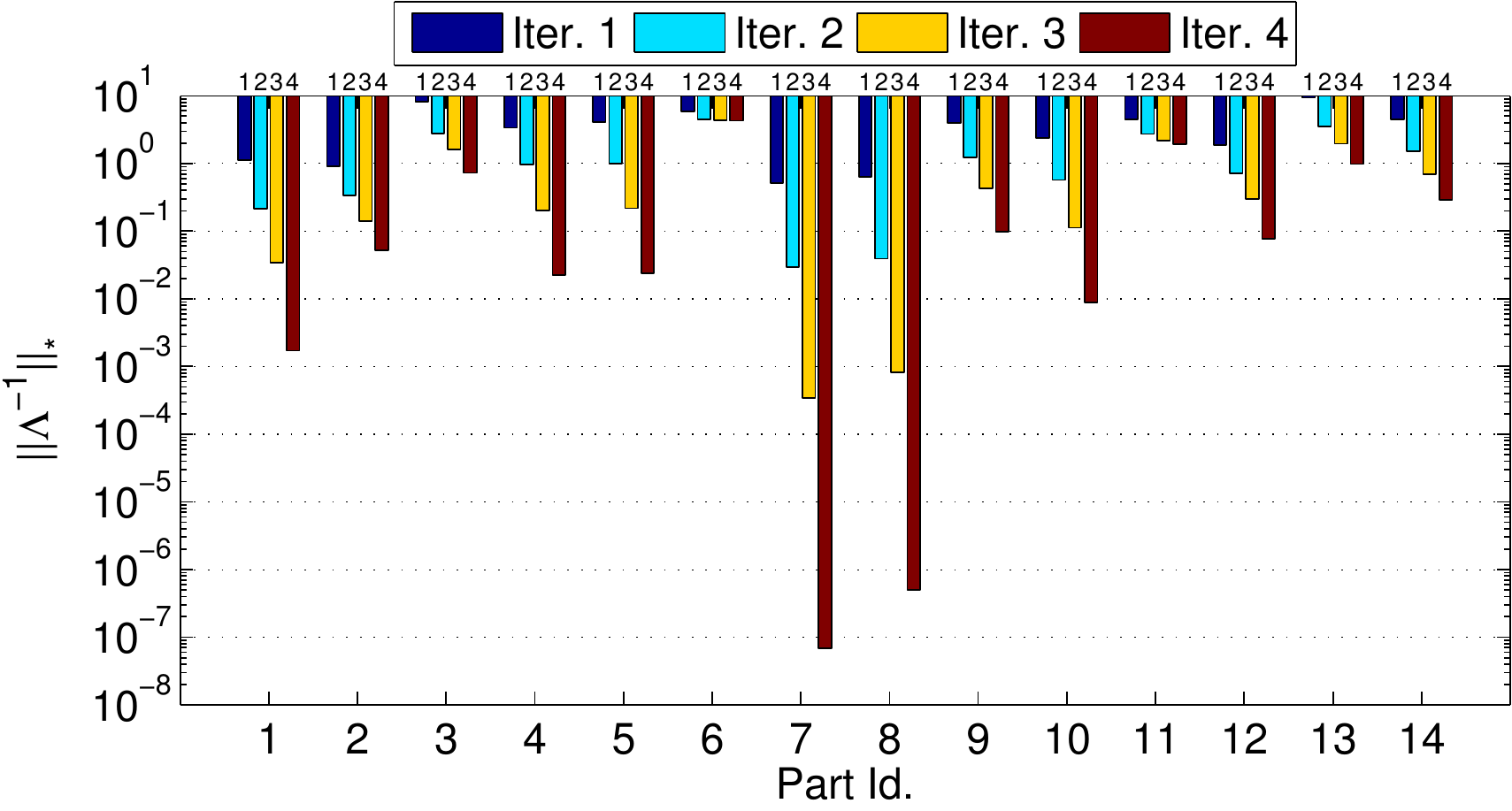}
\par\end{centering}

}
\par\end{centering}

\begin{centering}
\subfloat[CPA learning with Gaussian-Wishart prior for the tree-structured shape
model\label{fig:learning-with-prior}]{\begin{centering}
\includegraphics[width=0.47\textwidth]{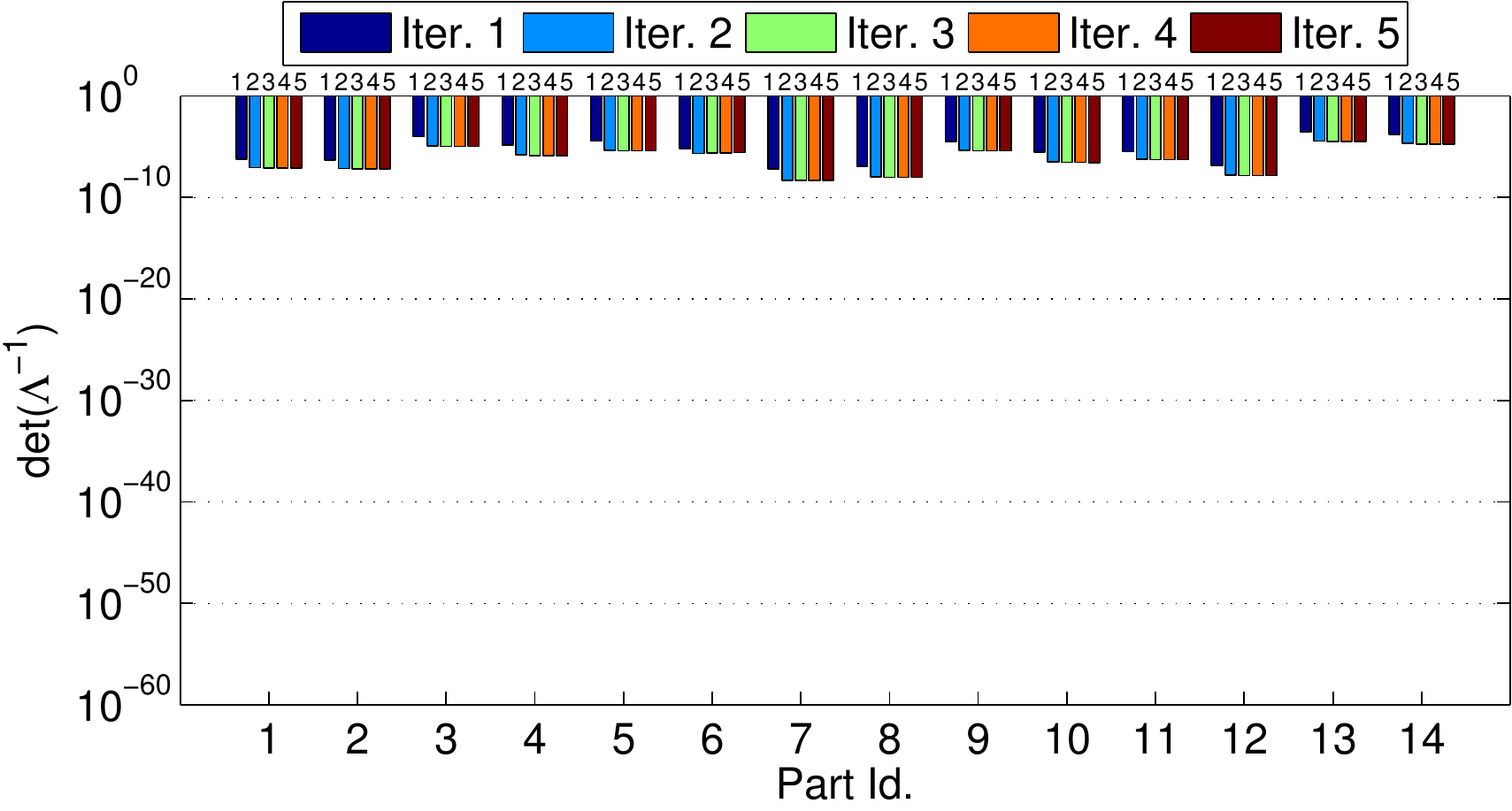}\hspace{0.03\textwidth}\includegraphics[width=0.47\textwidth]{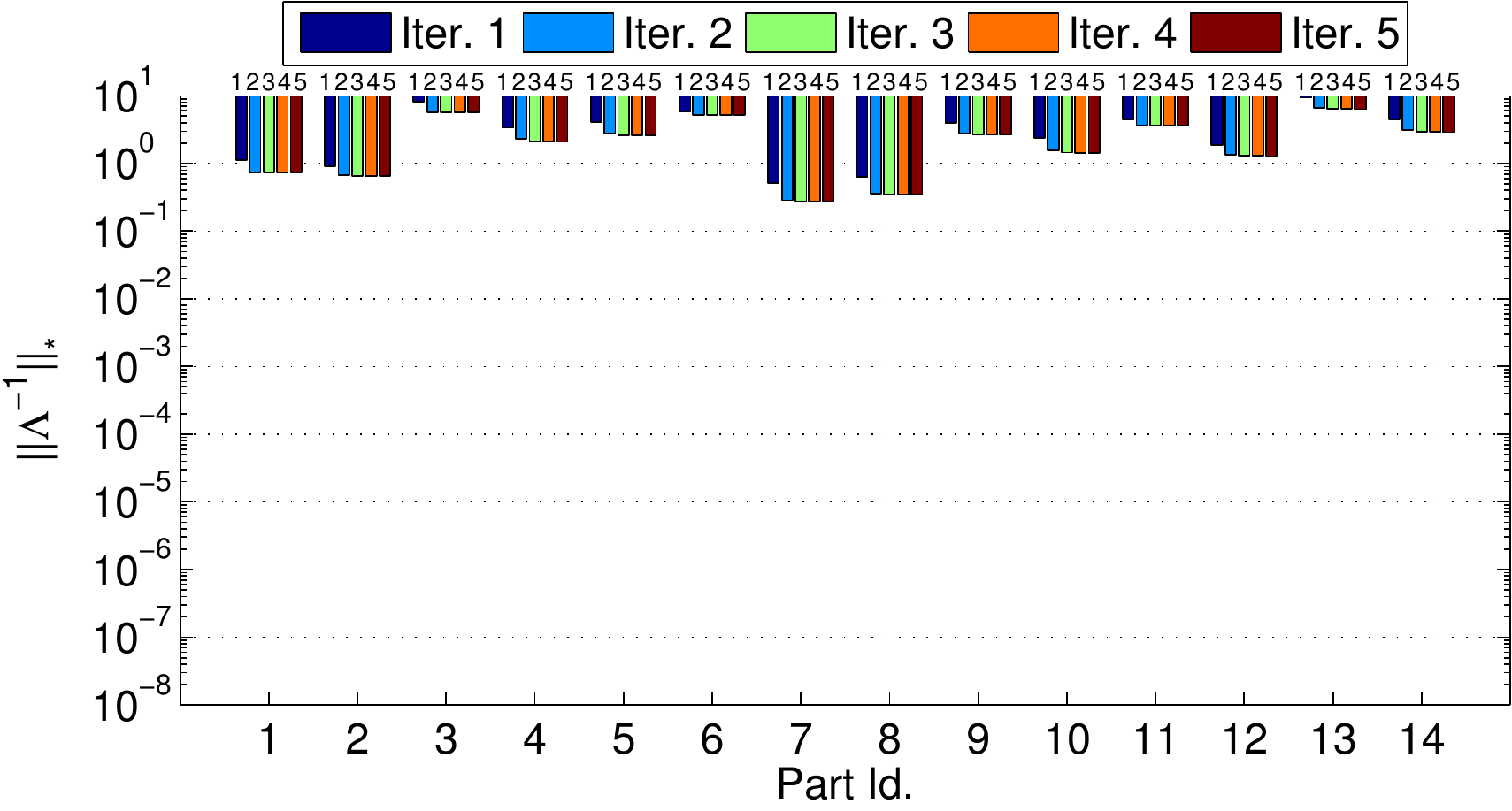}
\par\end{centering}

}
\par\end{centering}

\protect\caption{Strength of the variances of the Gaussian distributions for the differences
of part transformations at different iterations of CPA learning. For
a Gaussian distribution, the determinant (left-side figures) and nuclear
norm (right-side figures) of its covariance matrix indicate the overall
strength of its variances. Larger values indicate better flexibility
of the model, and vice versa. In this illustration, we use 60 images
from the MUCT dataset \citep{MUCT} to learn the CPA model define
in Fig.~\ref{fig:CPA-model}. Both the determinant and nuclear norm
are averaged over all the Gaussian distributions attached to the tree
edges. }
\end{figure*}

To jointly learn the tree-structured shape model and part dictionaries,
it is natural to take $\mathcal{Z}$ in (\ref{eq:mpRASL-obj}) as
the additional variables to optimize. To solve this revised problem,
one can alternately update part dictionaries using the algorithms
in Appendix~\ref{sec:mpBatch-solution}, which involves the variables
$\{A^{\langle i\rangle}\}_{i=1}^{m}$, $\{E^{\langle i\rangle}\}_{i=1}^{m}$,
$\check{\boldsymbol{\nu}}$, and $\boldsymbol{\sigma}$, and estimate
$\mathcal{Z}$ by maximum likelihood (ML) estimation, which fits parameters
$\{z^{\langle i\rangle}=(\mu^{\langle i\rangle},\Lambda^{\langle i\rangle})\}_{i=1}^{m}$
of Gaussian distributions with the updated $\{T_{k}(\check{\boldsymbol{\nu}})\}_{k=1}^{n}$.
Unfortunately, this approach empirically appears to give degenerate
solutions corresponding to a ``non-elastic'' shape model, where
the Gaussian distributions modeling the deformation differences of
connected nodes in the tree have variances of almost zero magnitude.
Fig.~\ref{fig:learning-without-prior} illustrates this phenomenon,
where we measure the determinant and nuclear norm%
\footnote{For a Gaussian distribution, the determinant of its covariance matrix
is the product of the standard deviations in its principle directions;
and, the nuclear norm is the sum of the variances in those directions.
Both of the two indicates the overall strength of its variances.%
} of the covariance matrices $\{{\Lambda^{\langle i\rangle}}^{-1}\}_{i=1}^{m}$,
which are iteratively updated in an alternating optimization process.

To remedy this problem, we consider imposing a prior on $\mathcal{Z}$
to regularize the learning of tree-structured shape model. Denote
$h(\mathcal{Z},\Phi)=-\ln p(\mathcal{Z}|\Phi)$ with $\Phi$ as the
hyper-parameters. Incorporating $h(\mathcal{Z},\Phi)$ into (\ref{eq:mpRASL-obj})
results in the following new objective for joint learning of $\mathcal{Z}$
and part dictionaries 
\begin{align*}
\min_{\underset{i=1,2,\ldots,m}{A^{\langle i\rangle},E^{\langle i\rangle},\check{\boldsymbol{\nu}},\boldsymbol{\sigma},\mathcal{Z}}} & \Biggl\{\sum_{i=1}^{m}\left(\Vert A^{\langle i\rangle}\Vert_{*}+\lambda^{\langle i\rangle}\Vert E^{\langle i\rangle}\Vert_{1}\right)\\
+\eta & \left(\sum_{k=1}^{n}g\left(T_{k}(\check{\boldsymbol{\nu}}),\mathcal{Z}\right)+h(\mathcal{Z},\Phi)\right)\Biggl\}
\end{align*}
\begin{gather}
\mathrm{s.t.}\quad D\circ\boldsymbol{\sigma}\circ T^{\langle i\rangle}(\check{\boldsymbol{\nu}})=A^{\langle i\rangle}+E^{\langle i\rangle}.\label{eq:mpBatch-obj}
\end{gather}
Let $\nu_{\delta,k}^{\langle i\rangle}=\nu_{k}^{\langle i\rangle}-\nu_{k}^{\langle j\rangle}$.
Based on (\ref{eq:joint-prob}) we have 
\begin{equation}
\sum_{k=1}^{n}g\left(T_{k}(\check{\boldsymbol{\nu}}),\mathcal{Z}\right)=-\sum_{i=1}^{m}\ln p(\{\nu_{\delta,k}^{\langle i\rangle}\}_{k=1}^{n}|z^{\langle i\rangle}),\label{eq:likelihood}
\end{equation}
where $z^{\langle i\rangle}=(\mu^{\langle i\rangle},\Lambda^{\langle i\rangle})$,
and $\nu_{\delta,k}^{\langle i\rangle}$ is drawn from the Gaussian
distribution with mean $\mu^{\langle i\rangle}$ and precision $\Lambda^{\langle i\rangle}$.
We impose prior on $\{\mu^{\langle i\rangle},\Lambda^{\langle i\rangle}\}$
using Gaussian-Wishart distribution, which is the conjugate prior
of the Gaussian distribution. Denote $\phi^{\langle i\rangle}$ as
the parameters of the Gaussian-Wishart prior for $z^{\langle i\rangle}$.
We have 
\begin{equation}
h(\mathcal{Z},\Phi)=-\sum_{i=1}^{m}\ln p(z^{\langle i\rangle}|\phi^{\langle i\rangle}).\label{eq:para-prior}
\end{equation}
where $\Phi=(\phi^{\langle1\rangle},\phi^{\langle2\rangle},\ldots,\phi^{\langle m\rangle})$
is the tuple of the hyper-parameters.

Given $\Phi$, we solve (\ref{eq:mpBatch-obj}) by alternately updating
the following two steps: 
\begin{enumerate}
\item Update $\mathcal{Z}$ by solving $\min_{\mathcal{Z}}\sum_{k=1}^{n}g\left(T_{k}(\check{\boldsymbol{\nu}}),\mathcal{Z}\right)+h(\mathcal{Z},\Phi)$; 
\item Update $\{A^{\langle i\rangle}\}_{i=1}^{m},\{E^{\langle i\rangle}\}_{i=1}^{m},\check{\boldsymbol{\nu}},\boldsymbol{\sigma}$
by solving (\ref{eq:mpRASL-obj}). 
\end{enumerate}
To update the first step, note that by Bayes' rule, 
\begin{align}
\sum_{k=1}^{n} & g\left(T_{k}(\check{\boldsymbol{\nu}}),\mathcal{Z}\right)+h(\mathcal{Z},\Phi)\label{eq:CPA-MAP-score}\\
= & -\sum_{i=1}^{m}\left\{ \ln p(z^{\langle i\rangle}|\{\nu_{\delta,k}^{\langle i\rangle}\}_{k=1}^{n},\phi^{\langle i\rangle})+\ln p(\{\nu_{\delta,k}^{\langle i\rangle}\}_{k=1}^{n})\right\} ,\nonumber 
\end{align}
where $p(\{\nu_{\delta,k}^{\langle i\rangle}\}_{k=1}^{n})$ does not
change with $z^{\langle i\rangle}$, and the terms of the summation
have no overlap variables. Thus the above first step can be updated
by independently solving the $m$ sub-problems 
\begin{equation}
\underset{z^{\langle i\rangle}}{\max}\, p(z^{\langle i\rangle}|\{\nu_{\delta,k}^{\langle i\rangle}\}_{k=1}^{n},\phi^{\langle i\rangle}),\label{eq:FinalEqnForTreeModelLearning}
\end{equation}
for $i=1,\dots,m$. It is a standard maximum \emph{a posterior} (MAP)
inference for the Gaussian distribution, where the posterior still
follows a Gaussian-Wishart distribution, whose parameters are different
from those of the prior. The maximum point of a Gaussian-Wishart PDF
can be derived in closed form in terms of its parameters. Please refer
to Appendix~\ref{sec:MAP-GW-Prior} for details of the Gaussian-Wishart
prior and inference problems relevant to it. For the above second
step, it is the problem of learning part dictionaries given a fixed
$\mathcal{Z}$, which is addressed in the previous subsection.

Now, it comes the problem: how to set hyper-parameters $\{\phi^{\langle i\rangle}\}_{i=1}^{m}$
properly? Since the initialization of $\check{\boldsymbol{\nu}}$
gives rough evidence on how the tree-structured shape model should
be, we set $\phi^{\langle i\rangle}$ to be consistent with the ML
estimation on $z^{\langle i\rangle}$ with the initial $\check{\boldsymbol{\nu}}$
for $i=1,2,\ldots,m$. Thus, the trained CPA model will not deviate
much from its rough estimation. Meanwhile, we also set $\phi^{\langle i\rangle}$
properly to control the prior's weight, i.e., to make the prior contribute
to the MAP estimation as much as $\vartheta n$ training samples,
where $\vartheta>0$. Still, please refer to Appendix~\ref{sec:MAP-GW-Prior}
for how to set $\phi$ with explicit physical meaning. The above learning
procedure for a CPA model is summarized in Algorithm~\ref{alg:CPA-model-learning}.

\begin{algorithm}
\begin{algorithmic}[1]

\Require Training images $D$, initial $\check{\boldsymbol{\nu}},\boldsymbol{\sigma}$,
and prior weight $\vartheta$

\Ensure  learned $\check{\boldsymbol{\nu}},\boldsymbol{\sigma},\mathcal{Z}$

\State $\mathcal{Z}\leftarrow\min_{\mathcal{Z}}\sum_{k=1}^{n}g\left(T_{k}(\check{\boldsymbol{\nu}}),\mathcal{Z}\right)$

\State Set $\Phi$ in consistency to $\mathcal{Z}$ and $\vartheta$

\While{ not converged }

\State Update $\{A^{\langle i\rangle}\}_{i=1}^{m},\{E^{\langle i\rangle}\}_{i=1}^{m},\check{\boldsymbol{\nu}},\boldsymbol{\sigma}$
by solving (\ref{eq:mpRASL-obj}). 

\State $\mathcal{Z}\leftarrow\min_{\mathcal{Z}}\sum_{k=1}^{n}g\left(T_{k}(\check{\boldsymbol{\nu}}),\mathcal{Z}\right)+h(\mathcal{Z},\Phi)$

\EndWhile

\end{algorithmic}

\protect\caption{CPA model learning \label{alg:CPA-model-learning}}
\end{algorithm}

The above algorithms for CPA model learning require a careful initialization
on $\boldsymbol{\sigma}$ and $\check{\boldsymbol{\nu}}$, which can
be done either by detectors \citep{viola2001adaboost,zhu2012dpmface}
or by manual annotations. In particular, for every $k^{\textrm{th}}$
training image, we localize the two eye corners to determine the similarity
transform $\sigma_{k}$. We then localize the facial landmarks at
the centers of the facial parts%
\footnote{We always design the parts to center at the facial landmarks that
are easy to annotate.%
}, and initialize $\nu_{k}^{\langle i\rangle}$ to satisfy that the
$i^{\textrm{th}}$ part has the pre-defined orientation and size relative
to the canonical form of the entire face that is determined by $\sigma_{k}$.
Although this initialization scheme is very exercisable, but it makes
the part transformation differences between linked parts identical
for all the $n$ training images. Consequently, we cannot get statistical
evidence to learn $\mathcal{Z}$. To overcome this difficulty, we
propose a heuristic initialization scheme for learning the CPA model.
In particular, we start with the part dictionary learning (\ref{eq:mpRASL-obj})
by setting $\eta=0$, i.e., without any shape constraint, and run
it for only a few (here, $5$) iterations of the generalized Gaussian
method composed of the repeats of the linearized problem (\ref{eq:mpBatch-obj-linearized})
(in the Appendix~\ref{sec:mpBatch-solution}). The updated part transformations
are used for the actual initialization of the CPA model learning.

Fig.~\ref{fig:learning-with-prior} shows the effect of our proposed
objective (\ref{eq:mpBatch-obj}) for joint learning of the CPA model,
where we investigate the covariance matrices $\{{\Lambda^{\langle i\rangle}}^{-1}\}_{i=1}^{m}$
of the Gaussian distributions obtained in each iteration of our algorithm.
Compared with Fig.~\ref{fig:learning-without-prior}, the stability
of their determinant and nuclear norm values suggests that a better
tree-structured shape model is obtained.

Up to now we have assumed that \emph{the configuration of the tree},
i.e., how the nodes (facial parts) are connected to form the tree,
is given. To learn the configuration of a tree, \citet{zhu2012dpmface}
used \citet{chow-liu} algorithm for the application of facial landmark
localization. Chow-Liu algorithm finds the configuration of a Bayesian
network that best approximates the joint distribution of all the variables.
However, for a node with children, our tree-structured shape model
(as well as \citet{zhu2012dpmface}'s shape model) assumes the independence
between its part transformation and its part transformation differences
with its children (refer to Section~\ref{sub:tree}), which is not
an assumption for Chow-Liu algorithm.

Instead, we take the tree configuration that minimizes the objective
function for learning the CPA model, which can be reduced to $\sum_{k=1}^{n}g\left(T_{k}(\check{\boldsymbol{\nu}}),\mathcal{Z}\right)+h(\mathcal{Z},\Phi)$
as (\ref{eq:CPA-MAP-score}). According to (\ref{eq:likelihood})
and (\ref{eq:para-prior}), we can decomposed it into the summation
of part-wise scores, where each term associates only with the edge
linking one node and its parent. In view of this, we first link the
$(m+1)$ nodes (recall the node of ``0'') into a complete directed
graph, compute the score associating with each of the $(m+1)m$ edges,
and take the minimum spanning tree rooted at ``0'' to be the optimal
configuration of the tree.

\subsection{Learning Mixture of CPA Models \label{sub:mCPA}}

The presented CPA model can be extended as a \emph{mixture of CPA}
models \emph{(mCPA)} to cope with a larger range of pose and/or expression
variations. Each component of the mCPA is parameterized the same as
that of a standard CPA model, but it may have a different number of
facial parts, since some facial parts may be occluded under pose changes.

For an mCPA model with $c$ components, $\mathcal{I}_{\iota}\subseteq\{1,2,\ldots,m\}$
denotes the index set of available parts for the $\iota^{\textrm{th}}$
component. Correspondingly, we also use $\{D_{\iota}^{\langle i\rangle}\}_{i=1}^{m}$
to denote the part dictionaries and $\mathcal{Z}_{\iota}=\{z_{\iota}^{\langle1\rangle},z_{\iota}^{\langle2\rangle},\ldots,z_{\iota}^{\langle m\rangle}\}$
to denote the parameters of tree-structured shape models. If the $\iota^{\textrm{th}}$
component does not has the $i^{\textrm{th}}$ part, i.e. $i\notin\mathcal{I}_{\iota}$,
$D_{\iota}^{\langle i\rangle}$ and $z_{\iota}^{\langle i\rangle}$
would just be treated as void notations and would not be used.

Learning an mCPA model is based on the formulation (\ref{eq:mpBatch-obj})
for the standard CPA model learning. The difference is that we consider
learning models of the $c$ components altogether, so that the low-rank
and sparse properties of corresponding part dictionaries belonging
to different components can be overall leveraged, resulting in consistent
part dictionaries across the $c$ components. Given $n$ training
face images, we write $D_{\iota}$ for the $n_{\iota}$ face images
assigned to the $\iota^{\textrm{th}}$ components, $\boldsymbol{\sigma}_{\iota}\in\mathbb{R}^{d\times n_{\iota}}$
for their holistic deformations, and $\check{\boldsymbol{\nu}}{}_{\iota}\in\mathbb{R}^{d\times m\times n_{\iota}}$
for their part deformations. Our objective for learning the mCPA model
can be written as 
\begin{gather}
\min_{\underset{i=1,2,\ldots,m;\,\iota=1,2,\ldots,c}{A^{\langle i\rangle},E^{\langle i\rangle},\check{\boldsymbol{\nu}}{}_{\iota},\boldsymbol{\sigma}{}_{\iota},\mathcal{Z}{}_{\iota}}}\Biggl\{\sum_{i=1}^{m}\left(\Vert A^{\langle i\rangle}\Vert_{*}+\lambda^{\langle i\rangle}\Vert E^{\langle i\rangle}\Vert_{1}\right)\label{eq:mCPA-learning-obj}\\
+\eta\sum_{\iota=1}^{c}\left(\sum_{k=1}^{n_{\iota}}g\left(T_{k}(\check{\boldsymbol{\nu}}{}_{\iota}),\mathcal{Z}{}_{\iota}\right)+h(\mathcal{Z}{}_{\iota},\Phi{}_{\iota})\right)\Biggl\},\nonumber 
\end{gather}
s.t. 
\begin{gather*}
D_{\iota}\circ\boldsymbol{\sigma}_{\iota}\circ T^{\langle i\rangle}(\check{\boldsymbol{\nu}}_{\iota})=A_{\iota}^{\langle i\rangle}+E_{\iota}^{\langle i\rangle}\;\mbox{for }i\in\mathcal{I}_{\iota}.\\
A_{\iota}^{\langle i\rangle},E_{\iota}^{\langle i\rangle}\mbox{ are empty matrices for }i\notin\mathcal{I}_{\iota},\\
A^{\langle i\rangle}=[A_{1}^{\langle i\rangle},A_{2}^{\langle i\rangle},\ldots,A_{c}^{\langle i\rangle}],\\
E^{\langle i\rangle}=[E_{1}^{\langle i\rangle},E_{2}^{\langle i\rangle},\ldots,E_{c}^{\langle i\rangle}].
\end{gather*}
The above objective can be solved similarly as for (\ref{eq:mpBatch-obj}).
Namely, we alternately update the part dictionaries and the tree-structured
shape models. After solving (\ref{eq:mCPA-learning-obj}), we have
the part dictionaries $D_{\iota}^{\langle i\rangle}\doteq D_{\iota}\circ\boldsymbol{\sigma}_{\iota}\circ T^{\langle i\rangle}(\check{\boldsymbol{\nu}}_{\iota})$
and the shape model parameters $\{\mathcal{Z}_{\iota}\}_{\iota=1}^{c}$.

Given a gallery set, ideally an mCPA model should be learned from
face images of this gallery. However, in some cases the gallery does
not contain non-frontal/non-neural face images to learn the corresponding
component models of the mCPA. To remedy this problem, we first learn
an mCPA model using a separate \emph{training set} that contains face
images of all the interested poses and expressions of a few subjects,
and then learn the part dictionaries on the gallery set of interest
while fixing the learned tree-structured shape model from that separate
training set.

With a learned mCPA model, we use the same method as that of the standard
CPA model to recognize a probe face, which requires us to select the
correct component corresponding to its pose and expression before
hand. In a fully automatic face recognition system, we may figure
out the pose and expression by off-the-shelf methods, which, we will
discuss in Section~\ref{sub:comparision}.

\section{Experiments \label{sec:experiments}}

In this section, we present experiments to evaluate the proposed CPA
method in the context of face recognition across illuminations, poses,
and expressions. We used images of frontal face with neutral expression
as the gallery, and face images with illumination, pose, and expression
variations as the probes. We used the CMU Multi-PIE \citep{MultiPIE}
and MUCT \citep{MUCT} datasets to conduct our experiments. The CMU
Multi-PIE dataset contains face images with well controlled illumination,
pose, and expression variations, and is thus intensively used for
controlled experiments.

\begin{table}
\protect\caption{Canonical part locations and sizes for the mCPA component for frontal
view and neutral expression in the experiments. \label{tab:part-info}}

\renewcommand\arraystretch{1.2}

\begin{centering}
\begin{tabular}{rlllllrr}
\hline 
\multirow{2}{*}{No.} & \multirow{2}{*}{Center location} &  & \multicolumn{2}{c}{Relative} &  & \multicolumn{2}{c}{Absolute}\tabularnewline
\cline{4-5} \cline{7-8} 
 &  &  & \multicolumn{1}{c}{w} & \multicolumn{1}{c}{h} &  & \multicolumn{1}{c}{w} & \multicolumn{1}{c}{h}\tabularnewline
\hline 
1 & Center of R-eyebrow &  & 0.4 & 0.2 &  & 24 & 16\tabularnewline
2 & Center of L-eyebrow &  & 0.4 & 0.2 &  & 24 & 16\tabularnewline
3 & Outer corner of R-eye &  & 0.53 & 0.4 &  & 32 & 32\tabularnewline
4 & Center of R-eye &  & 0.4 & 0.2 &  & 24 & 16\tabularnewline
5 & Inner corner of R-eye &  & 0.58 & 0.44 &  & 35 & 35\tabularnewline
6 & Inner corner of L-eye &  & 0.58 & 0.44 &  & 35 & 35\tabularnewline
7 & Center of L-eye &  & 0.4 & 0.2 &  & 24 & 16\tabularnewline
8 & Outer corner of L-eye &  & 0.53 & 0.4 &  & 32 & 32\tabularnewline
9 & R-wing of nose &  & 0.27 & 0.4 &  & 16 & 32\tabularnewline
10 & L-wing of nose &  & 0.27 & 0.4 &  & 16 & 32\tabularnewline
11 & Apex nasi &  & 0.53 & 0.28 &  & 32 & 22\tabularnewline
12 & Philtrum &  & 1.07 & 0.44 &  & 64 & 35\tabularnewline
13 & R-corner of mouth &  & 0.32 & 0.24 &  & 19 & 19\tabularnewline
14 & L-corner of mouth &  & 0.32 & 0.24 &  & 19 & 19\tabularnewline
15 & Mouth center &  & 0.67 & 0.28 &  & 40 & 22\tabularnewline
16 & Center of underlip &  & 0.53 & 0.2 &  & 32 & 16\tabularnewline
17 & Bottom of jaw &  & 0.53 & 0.28 &  & 32 & 22\tabularnewline
18 & R-ear &  & 0.4 & 0.4 &  & 24 & 32\tabularnewline
19 & L-ear &  & 0.4 & 0.4 &  & 24 & 32\tabularnewline
20 & R-cheek &  & 0.53 & 0.4 &  & 32 & 32\tabularnewline
21 & L-cheek &  & 0.53 & 0.4 &  & 32 & 32\tabularnewline
\hline 
\end{tabular}
\par\end{centering}

\medskip{}

{\small{}Remark: The relative sizes are in terms of the conventional
facial region aligned with the two eyes, i.e., 60$\times$80 window
with two outer eye corners at (5,22) and (56,22); and, the absolute
sizes are measured in pixels for the part dictionaries. }{\small \par}

\renewcommand\arraystretch{1}
\end{table}

We designed an mCPA model whose component for frontal view and neutral
expression consists of $21$ facial parts of varying sizes. The basic
constellation of the $21$ parts is listed in Table~\ref{tab:part-info}.
For the other components, the availability of a part is determined
by its visibility. The parameters used for the mCPA learning are set
as $\hat{\lambda}=1$, $\hat{\eta}=0.02$, and $\vartheta=0.25$ for
all the experiments reported in this section.


With learned mCPA models, we first evaluated our method in Section
\ref{sub:pose-expression} in the scenario of face recognition across
pose and expression with illumination variation. In particular, we
show the advantages of our CPA method over other alternatives, such
as the methods of a holistic face alignment followed by a holistic
or part-based face recognition. We then demonstrate in Section \ref{sub:alg-efficacy}
the effectiveness of our CPA method when using part-based face recognition
strategy. In Section \ref{sub:occlussion}, we test on face images
with synthesized occlusions to demonstrate the robustness of our method.
Finally, we compared with the state-of-the-art across-pose face recognition
methods in Section \ref{sub:comparision}.

For controlled experiments reported in Sections \ref{sub:pose-expression},
\ref{sub:alg-efficacy}, and \ref{sub:occlussion}, we initialized
face locations by manually annotating eye corner points, and assumed
that the pose and expression of each probe face are given. For practical
experiments in Section \ref{sub:comparision} that conduct fully automatic
face recognition across pose, we used off-the-shelf face detector
\citep{viola2001adaboost} and pose estimator \citep{zhu2012dpmface}
to initialize our method.


\subsection{Face Recognition Across Pose and Expression with Different Illumination
\label{sub:pose-expression}}

Our choice of off-the-shelf methods for face recognition across illumination
is based on SRC \citep{john2009src}, for which we used multiple gallery
images of varying illuminations for each subject. Under this setting,
we compare CPA with the following three baseline alternatives.
\begin{enumerate}
\item The first one manually aligns probe face images using labeled eye-corner
points. After manual alignment, face recognition is conducted in a
holistic manner. This alternative method is termed as ``manual+holistic''. 
\item The second one automatically aligns probe face images using the holistic
alignment algorithm of \citet{andrew2012practical}. After alignment,
face recognition is again conducted in a holistic manner. This alternative
is termed as ``holistic+holistic''. If using SRC \citep{john2009src}
as a classifier, this alternative is essentially the same as in \citep{andrew2012practical},
which is similar to our CPA method in the way that the gallery is
pruned to a subset before being used for recognition. For a fair comparison,
we also made the subset consist of $P$ subjects for the alignment
algorithm of \citet{andrew2012practical}. 
\item The third one also automatically aligns probe face images using the
holistic alignment algorithm of \citet{andrew2012practical}. The
same pruning scheme was used as for the second alternative. However,
after alignment, a part-based face recognition strategy is used where
positions of local parts are pre-defined and fixed relative to the
global face. This alternative is termed as ``holistic+parted''. 
\end{enumerate}
Our CPA method will be occasionally referred to as ``mCPA+parted''
in accordance with the names of these baselines. The chosen subject
number for pruning the gallery was set to $P=20$.

\subsubsection{Evaluation on the Multi-PIE Dataset \label{sub:pose-expression-MultiPIE}}

\begin{figure*}
\begin{centering}
\subfloat[$-\mbox{30}^{\circ}$]{\begin{centering}
\includegraphics[width=0.175\textwidth]{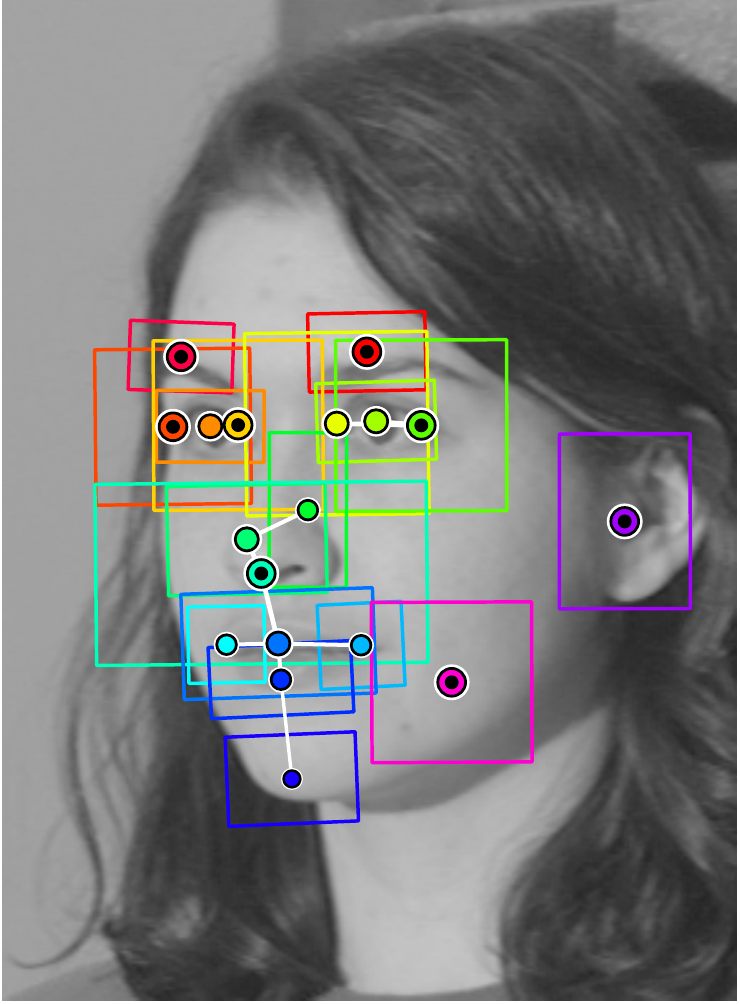}
\par\end{centering}

}\hfill{}\subfloat[$-\mbox{15}^{\circ}$]{\begin{centering}
\includegraphics[width=0.175\textwidth]{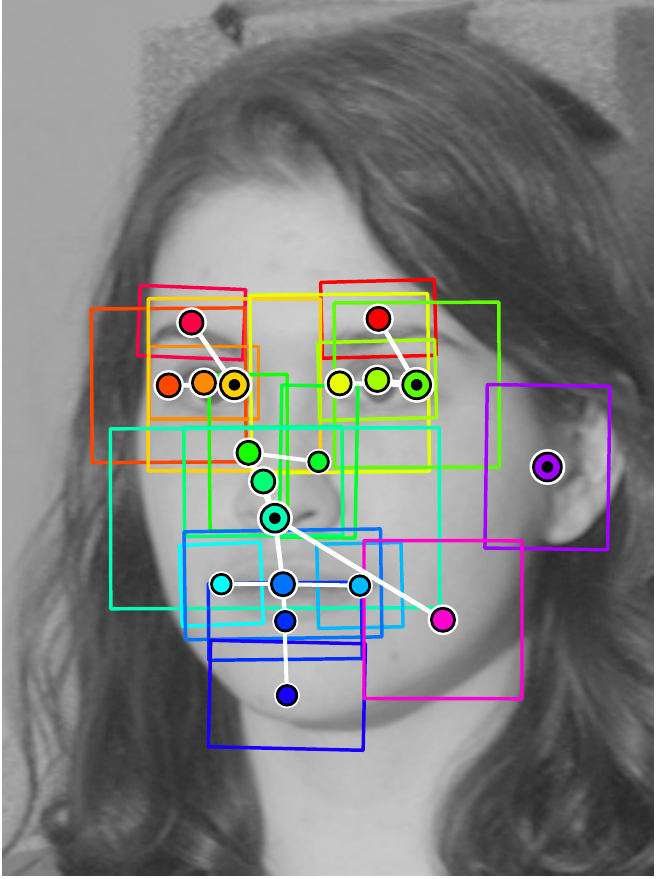}
\par\end{centering}

}\hfill{}\subfloat[$\mbox{0}^{\circ}$]{\begin{centering}
\includegraphics[width=0.175\textwidth]{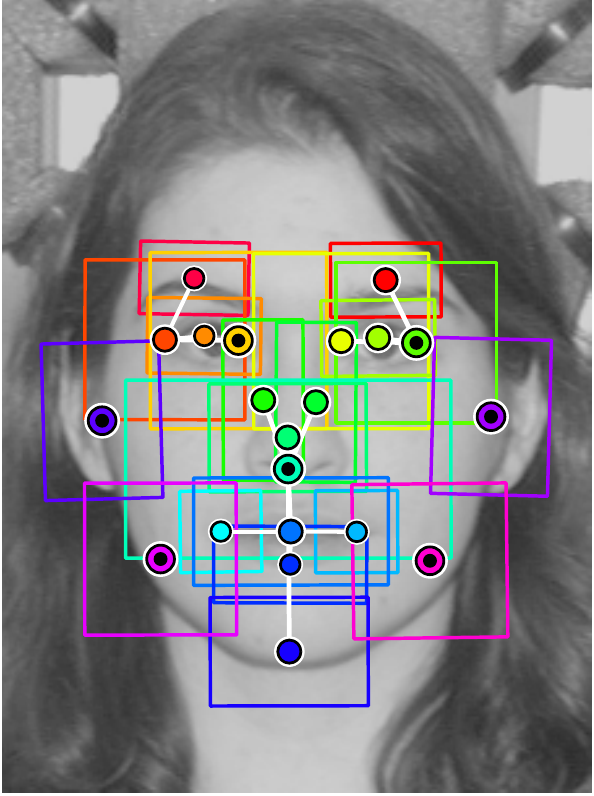}
\par\end{centering}

}\hfill{}\subfloat[$+\mbox{15}^{\circ}$]{\begin{centering}
\includegraphics[width=0.175\textwidth]{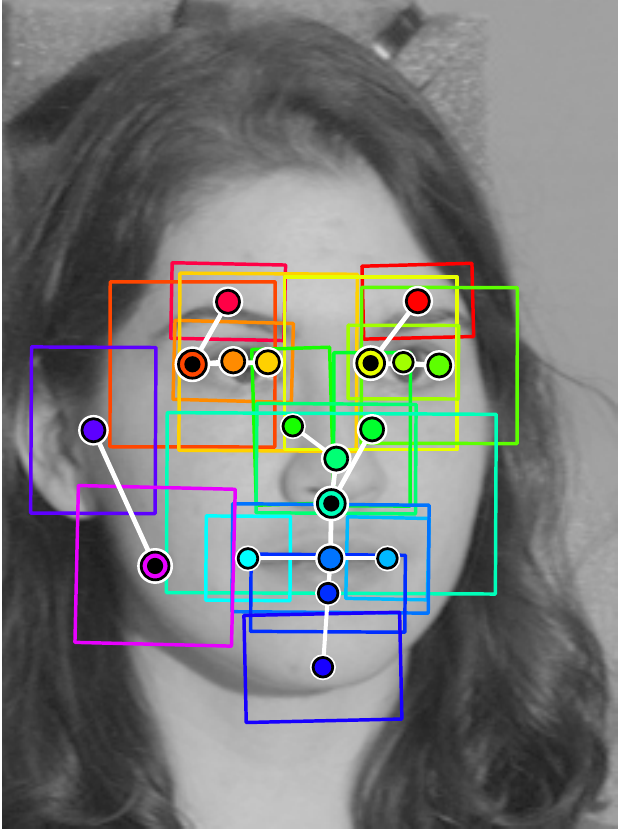}
\par\end{centering}

}\hfill{}\subfloat[$+\mbox{30}^{\circ}$]{\begin{centering}
\includegraphics[width=0.175\textwidth]{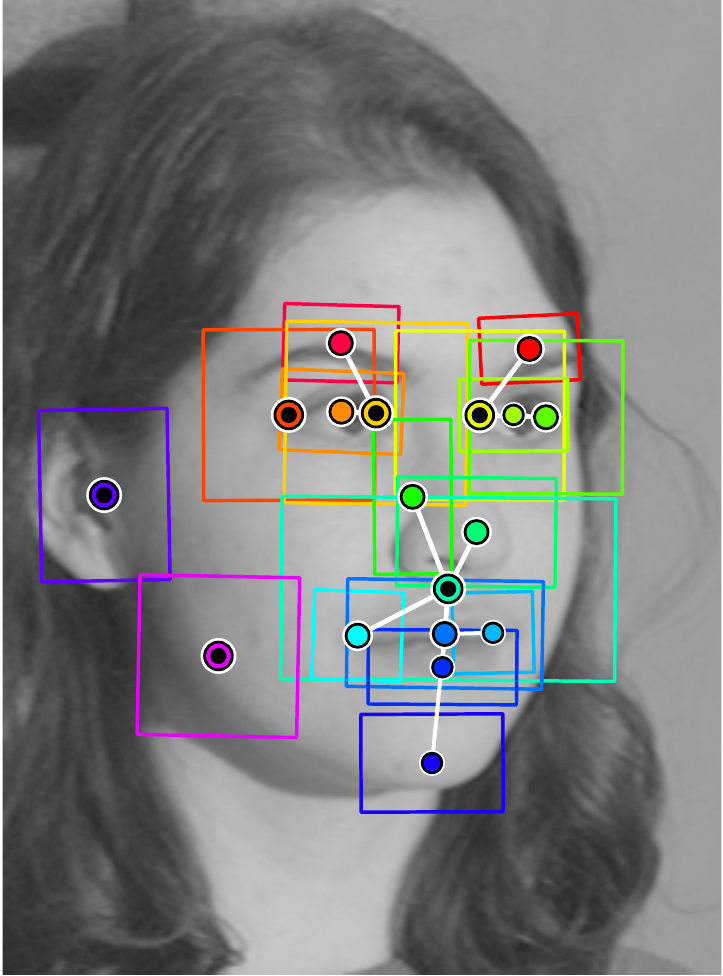}
\par\end{centering}

}
\par\end{centering}

\protect\caption{\label{fig:mCPA-poses-MultiPIE}The mCPA components fitted to faces
with different poses. Refer to the caption of Fig.~\ref{fig:CPA-instance}
for explanation.}
\end{figure*}

\begin{figure*}
\subfloat[Surprise]{\begin{centering}
\includegraphics[width=0.175\textwidth]{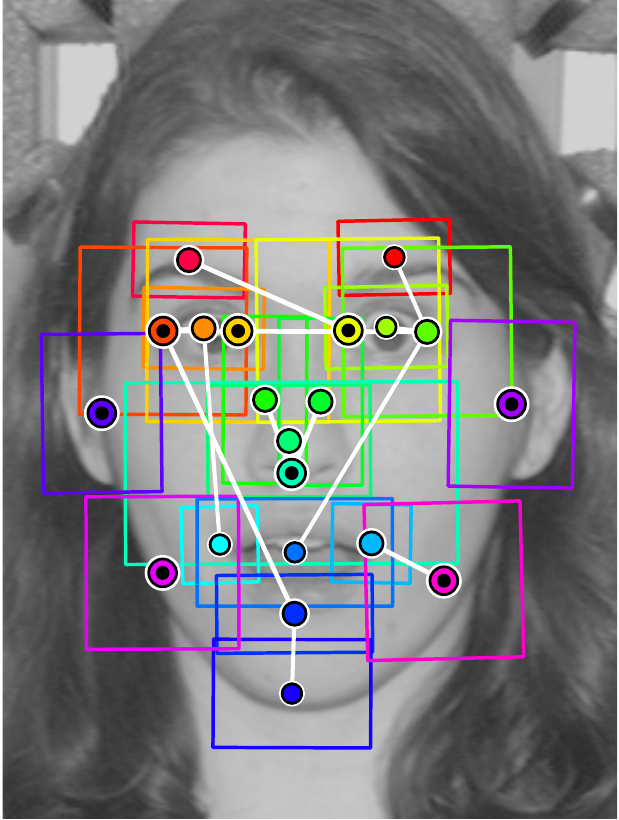}
\par\end{centering}

}\hfill{}\subfloat[Squint]{\begin{centering}
\includegraphics[width=0.175\textwidth]{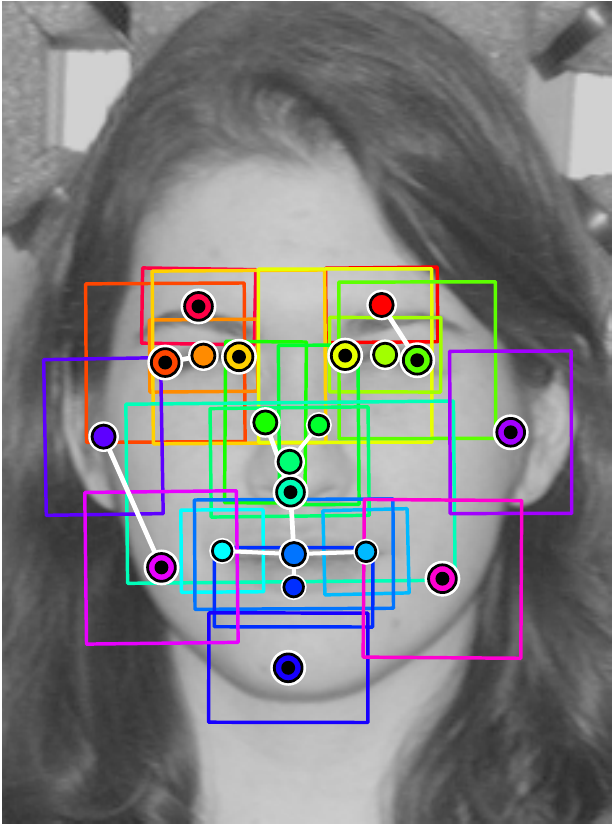}
\par\end{centering}

}\hfill{}\subfloat[Smile]{\begin{centering}
\includegraphics[width=0.175\textwidth]{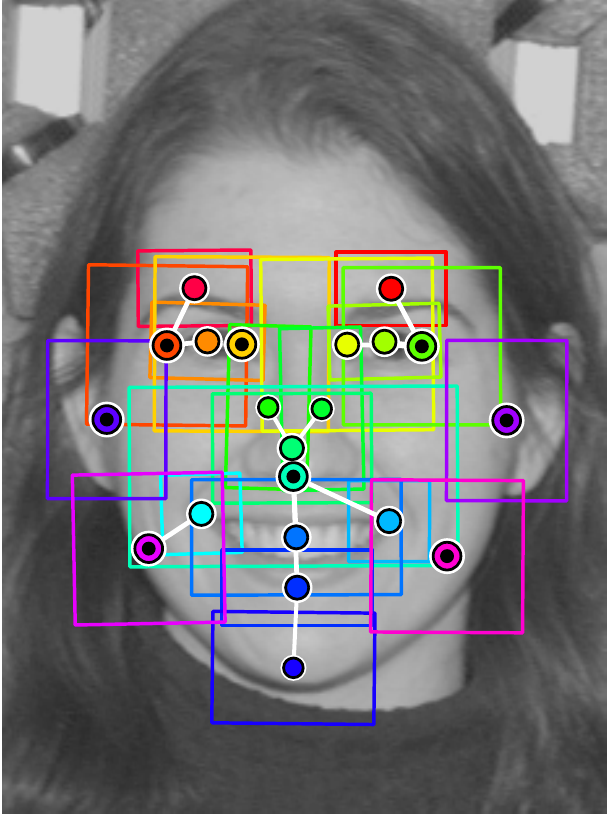}
\par\end{centering}

}\hfill{}\subfloat[Disgust]{\begin{centering}
\includegraphics[width=0.175\textwidth]{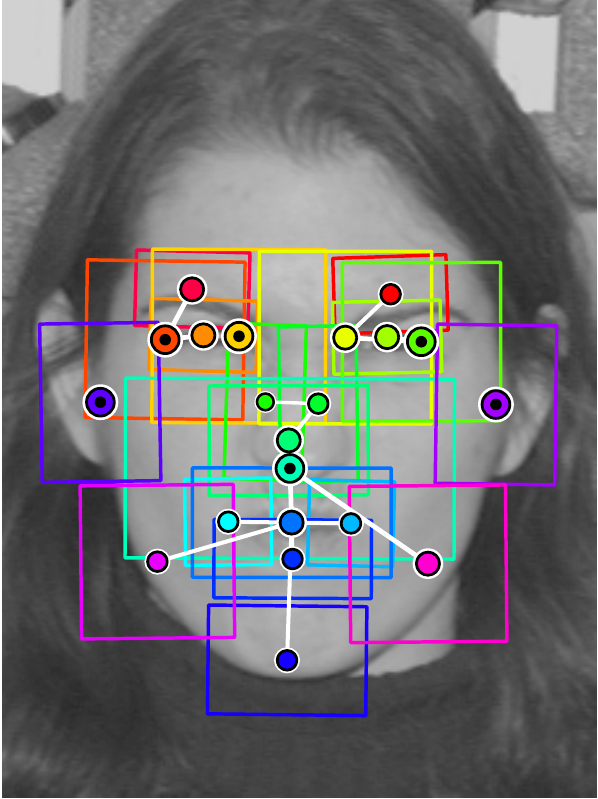}
\par\end{centering}

}\hfill{}\subfloat[Scream]{\begin{centering}
\includegraphics[width=0.175\textwidth]{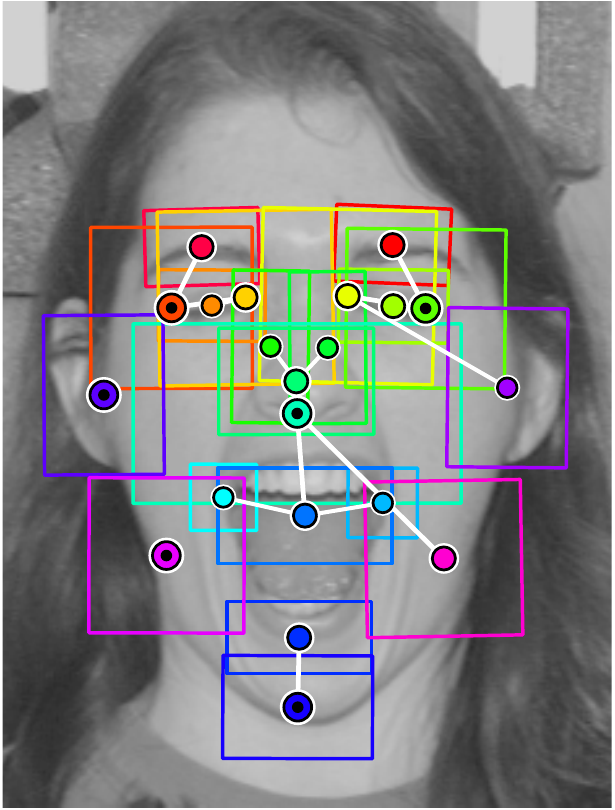}
\par\end{centering}

}

\protect\caption{\label{fig:mCPA-expressions-MultiPIE}The mCPA components fitted to
faces with non-neutral expressions. Refer to the caption of Fig.~\ref{fig:CPA-instance}
for explanation.}
\end{figure*}

The CMU Multi-PIE \citep{MultiPIE} is the largest publicly available
dataset suitable for test of our CPA method. It contains face images
of 337 subjects captured over the time span of about 5 months. These
face images are organized into 4 sessions according to their capture
time. There are 200 \textasciitilde{} 250 subjects present in each
session, where every subject is imaged from 15 different viewpoints
with the flashlight varying in 18 different directions . Face images
in Multi-PIE are with 6 well controlled facial expressions, including
the neutral one and 5 non-neutral ones. 

In our experiments, we used face images of 7 different illuminations
(illumination id: \{0, 1, 7, 13, 14, 16, 18\}) in the training and
gallery sets, and those of a novel illumination (illumination id:
10) in the probe set. We used 9 subjects (subject id: \{267, 272,
273, 274, 276, 277, 278, 286, 289\}), who appear in all the 2\textsuperscript{nd},
3\textsuperscript{rd}, 4\textsuperscript{th} sessions, for training
of the tree-structured shape model. The 249 subjects that appear in
Session~1 (subject id: 1 \textasciitilde{} 249) were used for testing.
Their face images in Session~1 constituted the gallery set, and those
in the other sessions were the probes. Table~\ref{tab:expr-glossary-MultiPIE}
gives more specifications on the probes.

For experiments of face recognition across pose on the Multi-PIE dataset,
we used the neutral-expression face images of 5 viewpoints of yaw
angles at $\mbox{0}^{\circ}$, $\pm\mbox{15}^{\circ}$, and $\pm\mbox{30}^{\circ}$.
For those across expression, we used face images of all the 5 available
non-neutral expressions under the frontal viewpoint. We built a 10-component
mCPA model as shown in Fig.~\ref{fig:mCPA-poses-MultiPIE} and Fig.~\ref{fig:mCPA-expressions-MultiPIE}.
Unless otherwise mentioned, these experiment settings were also used
in Section~\ref{sub:alg-efficacy}, \ref{sub:occlussion}, and \ref{sub:comparision}. 

\begin{table}
\protect\caption{\label{tab:pose-expr-MultiPIE}Face recognition across pose and expression
with different illumination on the first 249 subjects of Multi-PIE
dataset. Gallery illuminations: \{0, 1, 7, 13, 14, 16, 18\}. Probe
illumination: 10. }

\renewcommand\arraystretch{1.2}

\begin{centering}
\subfloat[Glossary for expressions of the probes. \label{tab:expr-glossary-MultiPIE}]{

\centering{}%
\begin{tabular}{cccccc}
\hline 
E.0 & E.1 & E.2 & E.3 & E.4 & E.5\tabularnewline
\hline 
Neutral & Surprise & Squint & Smile & Disgust & Scream\tabularnewline
S{[}234{]}R1 & S2R2 & S2R3 & S3R2 & S3R3 & S4R3\tabularnewline
\hline 
\multicolumn{6}{l}{Remark: S$a$R$b$ -- Recording number $b$ in Session $a$.}\tabularnewline
\end{tabular}}
\par\end{centering}

\begin{centering}
\subfloat[Recognition rates (\%) for across-pose and neutral-expression (E.0)
settings. \label{tab:poses-MultiPIE}]{

\centering{}%
\begin{tabular}{cc|r@{\extracolsep{0pt}.}lr@{\extracolsep{0pt}.}lr@{\extracolsep{0pt}.}lr@{\extracolsep{0pt}.}lr@{\extracolsep{0pt}.}l}
\hline 
\multirow{2}{*}{Align.} & \multirow{2}{*}{Recog.} & \multicolumn{2}{c}{$-\mbox{30}^{\circ}$} & \multicolumn{2}{c}{$-\mbox{15}^{\circ}$} & \multicolumn{2}{c}{$\mbox{0}^{\circ}$} & \multicolumn{2}{c}{$+\mbox{15}^{\circ}$} & \multicolumn{2}{c}{$+\mbox{30}^{\circ}$}\tabularnewline
\cline{3-12} 
 &  & \multicolumn{2}{c}{13\_0} & \multicolumn{2}{c}{14\_0} & \multicolumn{2}{c}{05\_1} & \multicolumn{2}{c}{05\_0} & \multicolumn{2}{c}{04\_1}\tabularnewline
\hline 
manual & holistic & 8&03 & 49&40 & 91&37 & 58&03 & 8&84\tabularnewline
holistic & holistic & 13&45 & 60&64 & 92&97 & 80&52 & 32&73\tabularnewline
holistic\textsuperscript{} & parted & 14&86 & 61&85 & 94&78 & 62&85 & 13&25\tabularnewline
mCPA & parted & \textbf{54}&\textbf{62} & \textbf{93}&\textbf{78} & \textbf{99}&\textbf{60} & \textbf{95}&\textbf{18} & \textbf{67}&\textbf{07}\tabularnewline
\hline 
\end{tabular}}
\par\end{centering}

\begin{centering}
\subfloat[Recognition rates (\%) for across-expression and frontal-pose (05\_1)
settings \label{tab:expr-0-MultiPIE}]{

\centering{}%
\begin{tabular}{cc|r@{\extracolsep{0pt}.}lr@{\extracolsep{0pt}.}lr@{\extracolsep{0pt}.}lr@{\extracolsep{0pt}.}lr@{\extracolsep{0pt}.}l}
\hline 
Align. & Recog. & E&1 & E&2 & E&3 & E&4 & E&5\tabularnewline
\hline 
manual & holistic & 44&85 & 75&76 & 64&78 & 49&69 & 28&74\tabularnewline
holistic & holistic & 67&88 & 81&21 & 68&55 & 65&41 & 36&78\tabularnewline
holistic\textsuperscript{} & parted & 67&88 & 81&21 & 79&25 & 61&01 & 43&10\tabularnewline
mCPA & parted & \textbf{84}&\textbf{85} & \textbf{93}&\textbf{33} & \textbf{89}&\textbf{94} & \textbf{85}&\textbf{53} & \textbf{58}&\textbf{62}\tabularnewline
\hline 
\end{tabular}}
\par\end{centering}

\begin{centering}

\par\end{centering}

\renewcommand\arraystretch{1}
\end{table}

Table~\ref{tab:poses-MultiPIE} reports recognition results of different
alternative methods on face images with varying degrees of pose changes.
Compared with ``manual+holistic'', ``holistic+holistic'' gives
improved performance, which suggests that automatic alignment can
improve face registration accuracy and consequently help for face
recognition across pose, even in a holistic alignment manner. The
performance of ``holistic+parted'' is unstable, which tells that
simple part-based recognition without part alignment is not a feasible
approach. Our CPA method significantly outperforms all these baselines.
For the 4 non-frontal viewpoints, CPA improves over the best alternative
method ``holistic+holistic'' approximately by 15\% \textasciitilde{}
40\% in terms of recognition rate.

An interesting observation from Table~\ref{tab:poses-MultiPIE} is
that our CPA method can \textit{almost perfectly} perform face recognition
under the across-session setting of Multi-PIE, where probe face images
of frontal viewpoint and neutral expression are from Sessions 2, 3
,4 of Multi-PIE and gallery face images are from Session 1. Under
this setting, our method gives 99.60\% recognition rate, about 5\%
higher than that of ``holistic+holistic'', i.e., the method in \citep{andrew2012practical}.
To the best of our knowledge, this is the best publicly known result
under this setting. Note that the challenges of face recognition across
sessions are mostly due to appearance change of human faces after
a certain period of time, since illumination variation should have
ideally been compensated by the multiple images of varying illuminations
in the gallery. Nevertheless, our method suggests that by assuming
human face as a piece-wise planar surface and using part-wise alignment,
the problem of face recognition across sessions can to a large extent
be overcome.

Table~\ref{tab:expr-0-MultiPIE} reports recognition results on frontal-view
face images of varying expressions. Comparative performance of different
alternative methods is very similar to that reported in Table~\ref{tab:poses-MultiPIE}
for face recognition across pose. Our CPA method outperforms the second
best ``holistic+holistic'' method by 12\% \textasciitilde{} 21\%
in terms of recognition rate.

\subsubsection{Evaluation on the MUCT Dataset}

In practical face recognition scenarios, illumination usually show
various levels of strength, and people often present near-neutral
expressions. The MUCT dataset is a good simulation of such scenarios,
where face images of natural expressions like frowns and minor smiles
are captured under frontal lighting of 3 strength levels. We used
the first 10 subjects (subject id: 0 \textasciitilde{} 9) of MUCT
for training the tree-structured shape model, and the rest 265 for
testing. Frontal-view face images (labeled as ``a'') were used as
gallery, and those of the 4 available non-frontal viewpoints (labeled
as ``b,c,d,e'') were used as probes.

\begin{table}
\protect\caption{Recognition rates (\%) for across-pose settings on the MUCT dataset.
\label{tab:poses-MUCT}}

\renewcommand\arraystretch{1.2}

\begin{centering}
\begin{tabular}{cc|r@{\extracolsep{0pt}.}lr@{\extracolsep{0pt}.}lr@{\extracolsep{0pt}.}lr@{\extracolsep{0pt}.}l}
\hline 
\multirow{3}{*}{Align.} & \multirow{3}{*}{Recog.} & \multicolumn{4}{c|}{Yaw Only} & \multicolumn{4}{c}{Pitch Only}\tabularnewline
\cline{3-10} 
 &  & \multicolumn{2}{c}{$+\mbox{20}^{\circ}$} & \multicolumn{2}{c|}{$+\mbox{38}^{\circ}$} & \multicolumn{2}{c}{$+\mbox{21}^{\circ}$} & \multicolumn{2}{c}{$-\mbox{22}^{\circ}$}\tabularnewline
\cline{3-10} 
 &  & \multicolumn{2}{c}{b} & \multicolumn{2}{c}{c} & \multicolumn{2}{c}{d} & \multicolumn{2}{c}{e}\tabularnewline
\hline 
manual & holistic & 99&17 & 55&89 & 95&15 & 99&58\tabularnewline
holistic & holistic & 99&86 & 93&20 & 98&61 & 99&58\tabularnewline
holistic\textsuperscript{} & parted & 99&17 & 69&21 & 81&14 & 98&75\tabularnewline
mCPA & parted & \multicolumn{2}{c}{\textbf{100}} & \textbf{99}&\textbf{86} & \textbf{99}&\textbf{86} & \multicolumn{2}{c}{\textbf{100}}\tabularnewline
\hline 
\end{tabular}
\par\end{centering}

\renewcommand\arraystretch{1}
\end{table}

Table~\ref{tab:poses-MUCT} reports recognition results of different
alternative methods. Consistent to the results reported in Section
\ref{sub:pose-expression-MultiPIE}, our CPA method outperforms all
the other 3 alternatives. It in fact performs \textit{almost perfectly}
for all the 4 degrees of pose change.


\subsection{Effectiveness of Part-based Recognition in CPA method \label{sub:alg-efficacy}}

The CPA method integrates part alignment with part-based recognition
in a cohesive way. Part-based face recognition fuses weaker predictions
from individual parts to make a stronger final decision, where a variety
of methods can be used for recognition of individual parts. In this
section, we investigate the varying discriminative power of individual
parts, and also how existing representative face recognition methods
perform for part recognition in CPA. We also present experiments to
show the efficacy of the proposed pruning scheme in Algorithm~\ref{alg:recognition}.
These investigations were conducted on the Multi-PIE dataset under
the setting of face recognition across pose with different illumination
for the gallery and probe (the setting in Section~\ref{sub:pose-expression-MultiPIE}
for producing Table~\ref{tab:poses-MultiPIE}).


\subsubsection{Part Discriminativeness}

\begin{figure*}
\begin{centering}
\includegraphics[width=1\textwidth]{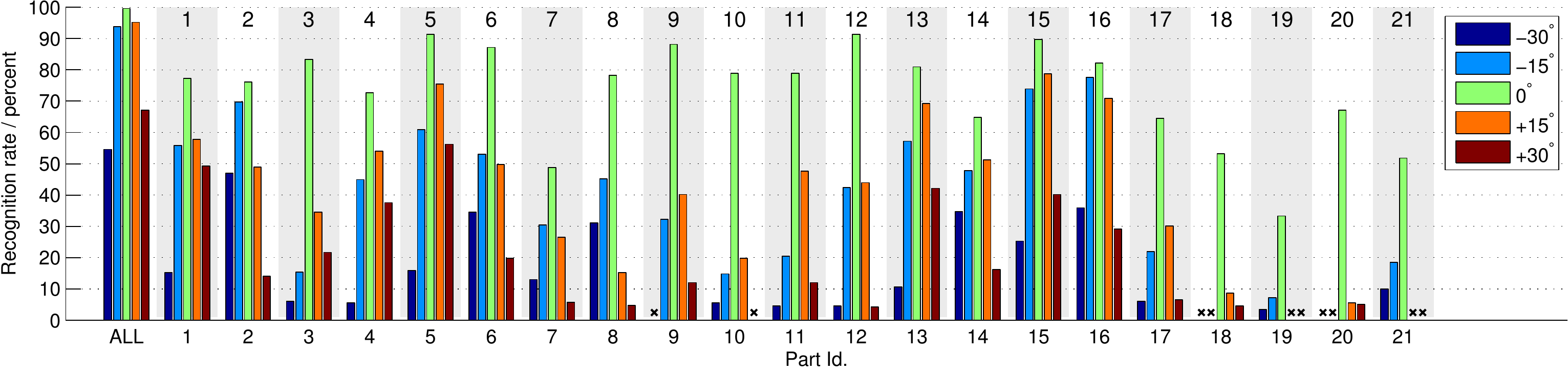}
\par\end{centering}

\protect\caption{Accuracy of recognizing faces by single parts on Multi-PIE (first
249 subjects). \label{fig:single-parts} Parts are aligned by CPA,
but used individually for recognition. Unavailable parts are referred
to as ``\textsf{x}''. ``ALL'' means voting the predicted labels
of all the parts. }
\end{figure*}

Fig.~\ref{fig:single-parts} reports results obtained by recognizing
each of the 21 parts that are jointly aligned by our CPA method. The
recognition rates on most of the 21 parts are around 60\% \textasciitilde{}
90\% for the frontal pose and 40\% \textasciitilde{} 80\% for the
$\pm\mbox{15}^{\circ}$ poses. These results suggest that the discriminativeness
of individual parts is good, but not strong enough for a high-accuracy
recognition performance. By fusing the predictions from individual
parts, a high-accuracy final recognition (``ALL'' in Fig.~\ref{fig:single-parts})
can be achieved for pose change of within $-\mbox{15}^{\circ}$ \textasciitilde{}
$+\mbox{15}^{\circ}$ yaw angles. Even for a larger pose change of
$\pm\mbox{30}^{\circ}$ yaw angles, our CPA method by fusing predictions
from aligned individual parts performs fairly well, while recognition
rates for most of individual parts are below 20\%. This again demonstrates
the efficacy of our proposed CPA method that integrates part-based
recognition with part alignment.

\subsubsection{Integrated with Different Recognition Methods}

Recognition of aligned individual parts in CPA is realized by off-the-shelf
face recognition methods. Representatives of these methods include
Nearest Subspace (NS) \citep{NS}, Linear Discriminate Analysis (LDA)
\citep{pami1997fisherface}, and Local Binary Pattern (LBP) \citep{pami2006lbp}.
In this section, we investigate how these different choices of recognition
methods perform when integrated into our CPA method. Here we compared
the CPA method (``mCPA+parted'') with only ``holistic+holistic''.

For LDA, we learned its projection matrix after pruning candidate
subjects and aligning gallery images, both of which are necessary
steps for ``holistic+holistic'' and our CPA method. For LBP, we
also extracted LBP features in the same stage, i.e., after the preparation
of aligned gallery subsets. All the experimental settings were the
same as those used in preceding sections, expect replacing the previously
used SRC with NS, LDA, or LBP. Considering the promise of LBP for
one-shot face recognition where only a single image per subject is
available in gallery, we also conducted experiments under this one-shot
setting, where probe and gallery face images are of the same illumination.
Table~\ref{tab:different-methods-remarks} summarizes the illumination
settings of experiments reported in this section.


\begin{table}
\protect\caption{Integrated with different recognition methods on Multi-PIE (first
249 subjects) \label{tab:different-methods}}

\renewcommand\arraystretch{1.2}

\begin{centering}

\par\end{centering}

\begin{centering}
\subfloat[Illumination settings \label{tab:different-methods-remarks}]{

\centering{}%
\begin{tabular}{l|cc}
\hline 
Strategy & Gallery illumi. & Probe illumi.\tabularnewline
\hline 
DI: Different illum. & 0, 1, 7, 13, 14, 16, 18 & 10\tabularnewline
SI: Same illum. & 7 & 7\tabularnewline
\hline 
\end{tabular}}
\par\end{centering}

\begin{centering}
\subfloat[Recognition rates for across-pose and neutral-expression settings
\label{tab:different-methods-results}]{

\centering{}%
\begin{tabular}{c|c|r@{\extracolsep{0pt}.}lr@{\extracolsep{0pt}.}lr@{\extracolsep{0pt}.}lr@{\extracolsep{0pt}.}lr@{\extracolsep{0pt}.}l}
\hline 
\multirow{2}{*}{Recog.} & \multirow{2}{*}{Align.} & \multicolumn{2}{c}{$-\mbox{30}^{\circ}$} & \multicolumn{2}{c}{$-\mbox{15}^{\circ}$} & \multicolumn{2}{c}{$\mbox{0}^{\circ}$} & \multicolumn{2}{c}{$+\mbox{15}^{\circ}$} & \multicolumn{2}{c}{$+\mbox{30}^{\circ}$}\tabularnewline
\cline{3-12} 
 &  & \multicolumn{2}{c}{13\_0} & \multicolumn{2}{c}{14\_0} & \multicolumn{2}{c}{05\_1} & \multicolumn{2}{c}{05\_0} & \multicolumn{2}{c}{04\_1}\tabularnewline
\hline 
NS & holistic & 8&84 & 44&38 & 76&31 & 60&44 & 20&68\tabularnewline
(DI) & mCPA & \textbf{43}&\textbf{98} & \textbf{85}&\textbf{94} & \textbf{98}&\textbf{19} & \textbf{88}&\textbf{76} & \textbf{65}&\textbf{26}\tabularnewline
\hline 
LDA & holistic & 9&64 & 42&57 & 84&94 & 63&45 & 16&67\tabularnewline
(DI) & mCPA & \textbf{26}&\textbf{51} & \textbf{73}&\textbf{90} & \textbf{98}&\textbf{80} & \textbf{77}&\textbf{31} & \textbf{28}&\textbf{57}\tabularnewline
\hline 
LBP & holistic & 34&34 & 79&52 & 95&18 & 88&35 & 39&76\tabularnewline
(DI) & mCPA & \textbf{59}&\textbf{44} & \textbf{86}&\textbf{95} & \textbf{97}&\textbf{39} & \textbf{87}&\textbf{55} & \textbf{65}&\textbf{06}\tabularnewline
\hline 
LBP & holistic & 57&03 & 92&77 & 97&39 & 92&97 & 60&44\tabularnewline
(SI) & mCPA & \textbf{83}&\textbf{53} & \textbf{96}&\textbf{99} & \textbf{98}&\textbf{80} & \textbf{95}&\textbf{38} & \textbf{76}&\textbf{91}\tabularnewline
\hline 
\end{tabular}}
\par\end{centering}

\renewcommand\arraystretch{1}
\end{table}

Recognition rates of CPA with integration of different methods are
reported in Table~\ref{tab:different-methods-results}. Table~\ref{tab:different-methods-results}
tells that with any choice of NS, LDA, or LBP, our proposed CPA for
part-based alignment is superior to holistic face alignment. This
confirms that our proposed CPA helps face recognition by effectively
aligning individual facial parts.

\subsubsection{Pruning Efficacy}

Pruning scheme in Algorithm \ref{alg:recognition} certainly leads
to more efficient algorithm. In this section, we are interested in
investigating how it impacts the recognition performance. To this
end, we modified Algorithm~\ref{alg:recognition} by simply removing
the pruning scheme, i.e., setting $P$ to the subject number in the
gallery. Bottom half of Table~\ref{tab:pruning-effects} lists recognition
rates of the non-pruning CPA method together with their difference
with those of the original CPA. For the non-frontal poses, noticeable
performance drops occur when the pruning scheme is removed from CPA.
Top half of Table~\ref{tab:pruning-effects} gives more evidence
on the impact of the pruning scheme. Taking the non-pruning version
as the baseline, we find that, for the non-frontal poses, the original
CPA method corrects one time more recognition errors than it introduces,
where ``correcting'' a recognition error means correctly recognizing
an image that is falsely recognized by the baseline, and ``introducing''
means the opposite. Experimental results reported in Table~\ref{tab:pruning-effects}
were obtained by using SRC for part recognition. All other settings
were the same as those used in preceding sections.

\begin{table}
\protect\caption{Impact of the pruning scheme on CPA (first 249 subjects of Multi-PIE):
\label{tab:pruning-effects} the standard CPA is compared with the
CPA without pruning scheme (the baseline). \emph{Top half:} ``Corrected''
errors are those occurred for the baseline but not for the standard
CPA, and the ``introduced'' errors are the opposite. \emph{Bottom
half:} $\Delta\mbox{RR}=\mbox{(Standard CPA's RR)}-\mbox{(Baseline's RR)}$. }

\renewcommand\arraystretch{1.2}

\begin{centering}
\begin{tabular}{l|c|rrrrr}
\hline 
Errors ...  & Cnt. & $-\mbox{30}^{\circ}$ & $-\mbox{15}^{\circ}$ & $\mbox{0}^{\circ}$ & $+\mbox{15}^{\circ}$ & $+\mbox{30}^{\circ}$\tabularnewline
\cline{3-7} 
by pruning & type & 13\_0 & 14\_0 & 05\_1 & 05\_0 & 04\_1\tabularnewline
\hline 
\multirow{2}{*}{Corrected} & Num. & 77 & 26 & 0 & 26 & 78\tabularnewline
 & \% & 15.46 & 5.22 & 0.00 & 4.22 & 15.66\tabularnewline
\hline 
\multirow{2}{*}{Introduced} & Num. & 33 & 14 & 1 & 14 & 39\tabularnewline
 & \% & 6.63 & 2.81 & 0.20 & 2.01 & 7.83\tabularnewline
\hline 
\hline 
\multicolumn{2}{l|}{RR -- no pruning \%} & 45.78 & 91.37 & 99.80 & 92.97 & 59.24\tabularnewline
\multicolumn{2}{l|}{$\Delta$RR with orig.\ \%} & 8.84 & 2.41 & $-$0.20 & 2.21 & 7.83\tabularnewline
\hline 
\multicolumn{7}{l}{Remarks:\hfill{}RR -- recognition rate; \hfill{}orig. -- pruning
with $P=20$. }\tabularnewline
\end{tabular}
\par\end{centering}

\renewcommand\arraystretch{1}
\end{table}

\subsection{Recognition with Synthetic Random Block Occlusion \label{sub:occlussion}}

\begin{figure}
\begin{centering}
\includegraphics[width=0.15\columnwidth]{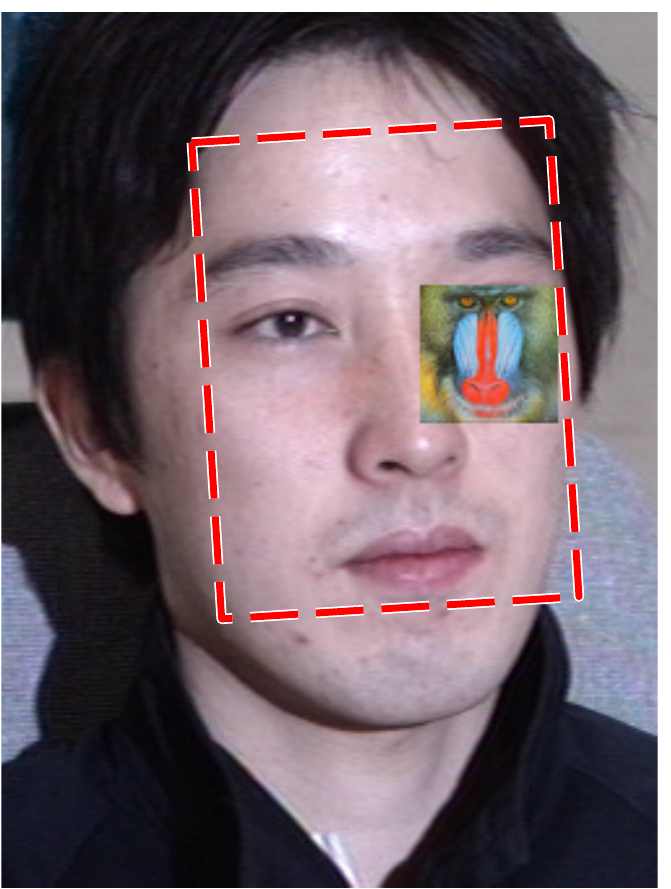}\hfill{}\includegraphics[width=0.15\columnwidth]{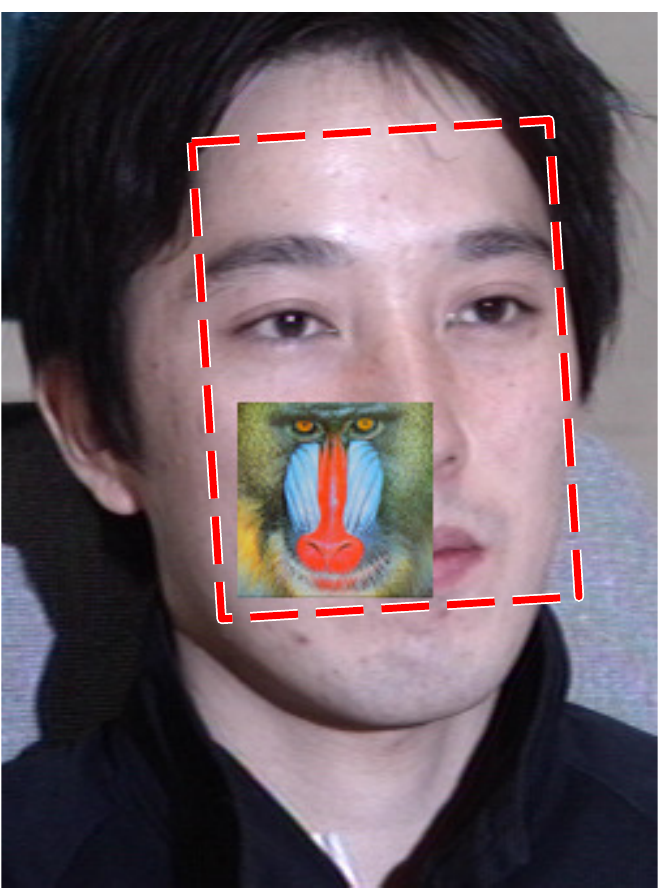}\hfill{}\includegraphics[width=0.15\columnwidth]{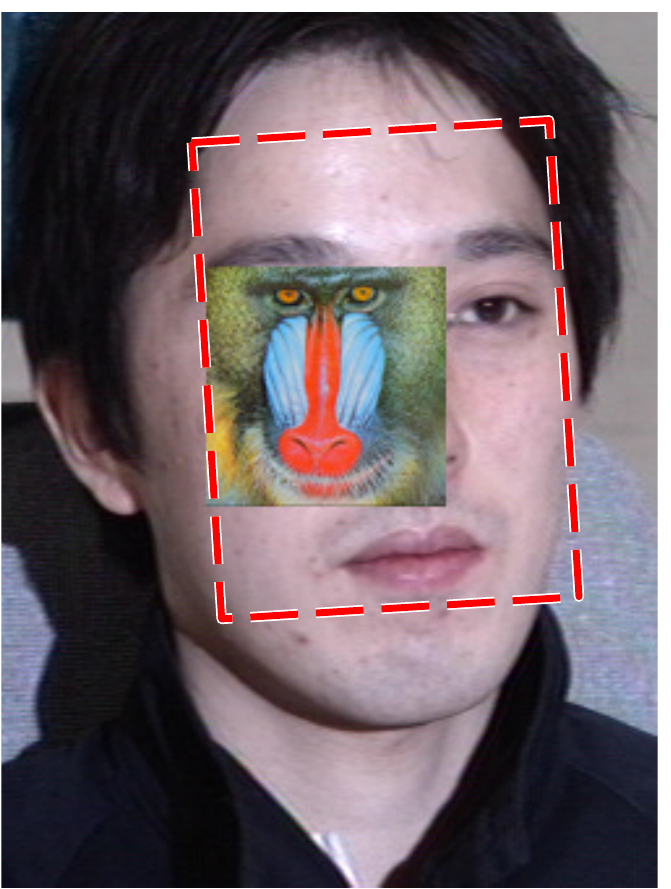}\hfill{}\includegraphics[width=0.15\columnwidth]{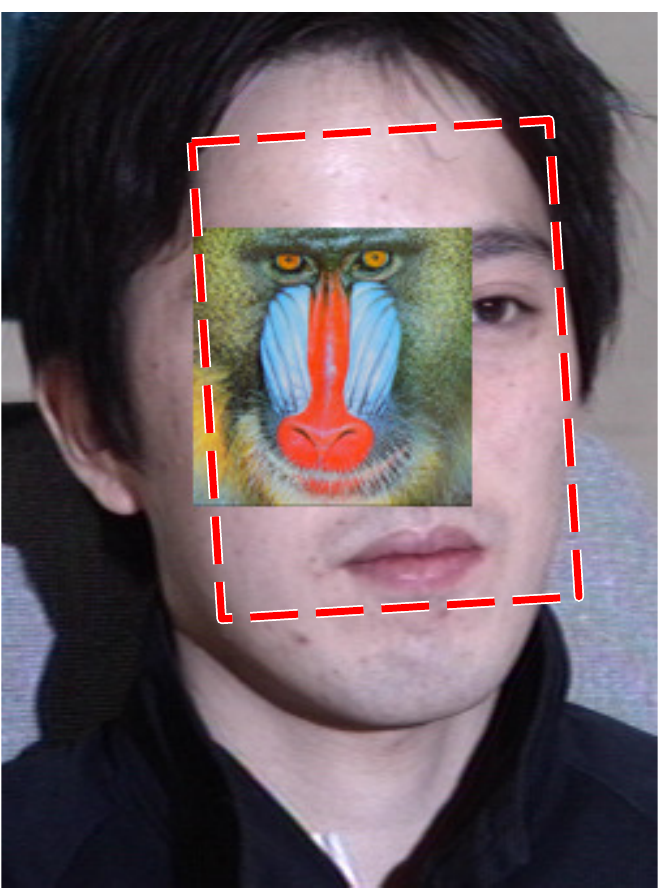}\hfill{}\includegraphics[width=0.15\columnwidth]{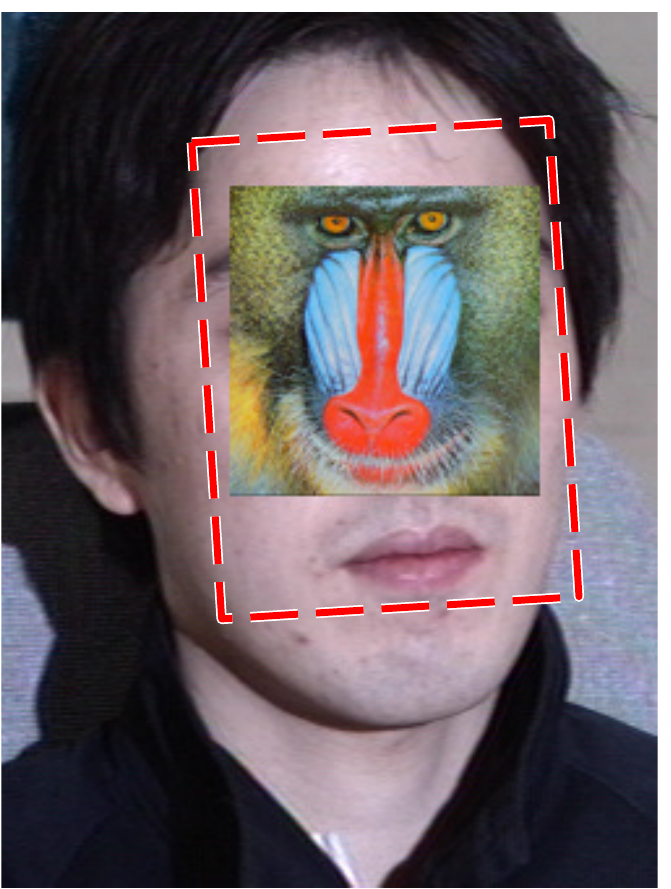}\hfill{}\includegraphics[width=0.15\columnwidth]{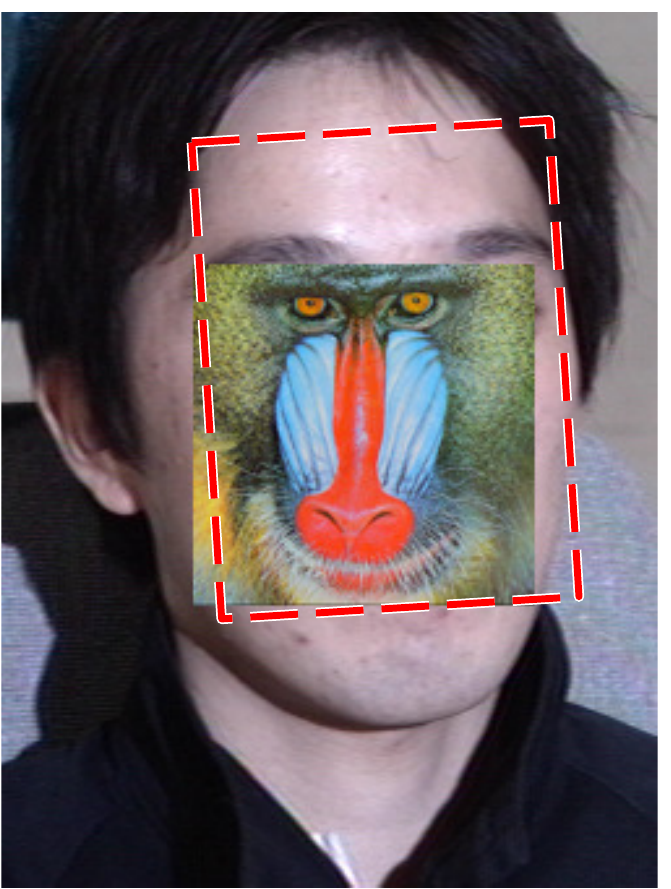}
\par\end{centering}

{\small{}}%
\makebox[0.15\columnwidth]{%
{\small{}\hfill{}10\%\hfill{}}%
}{\small{}\hfill{}}%
\makebox[0.15\columnwidth]{%
{\small{}\hfill{}20\%\hfill{}}%
}{\small{}\hfill{}}%
\makebox[0.15\columnwidth]{%
{\small{}\hfill{}30\%\hfill{}}%
}{\small{}\hfill{}}%
\makebox[0.15\columnwidth]{%
{\small{}\hfill{}40\%\hfill{}}%
}{\small{}\hfill{}}%
\makebox[0.15\columnwidth]{%
{\small{}\hfill{}50\%\hfill{}}%
}{\small{}\hfill{}}%
\makebox[0.15\columnwidth]{%
{\small{}\hfill{}60\%\hfill{}}%
}{\small \par}

\protect\caption{Images with synthetic block occlusions of different size. The percents
below the images indicates the occlusion ratio with respect to the
regions enclosed with dashed lines. \label{fig:occluded-images}}
\end{figure}

We report experiments in this section to demonstrate the robustness
of CPA against partial occlusion. As shown in Fig.~\ref{fig:occluded-images},
we synthesized partially occluded probe face images by adding block
occlusion at random positions of a probe face. The size of occluded
blocks varied from 10\% to 60\% of the holistic face. Experiment settings
were set the same as in Section~\ref{sub:pose-expression-MultiPIE}.
The method \citep{andrew2012practical}, i.e.\ ``holistic+holistic'',
was taken as the baseline.

\begin{figure}
\begin{centering}
\subfloat[Recognition rates at different occlusion levels for the frontal-pose
and neutral-expression setting \label{fig:occlusion-levels}]{\begin{centering}
\includegraphics[width=0.9\columnwidth]{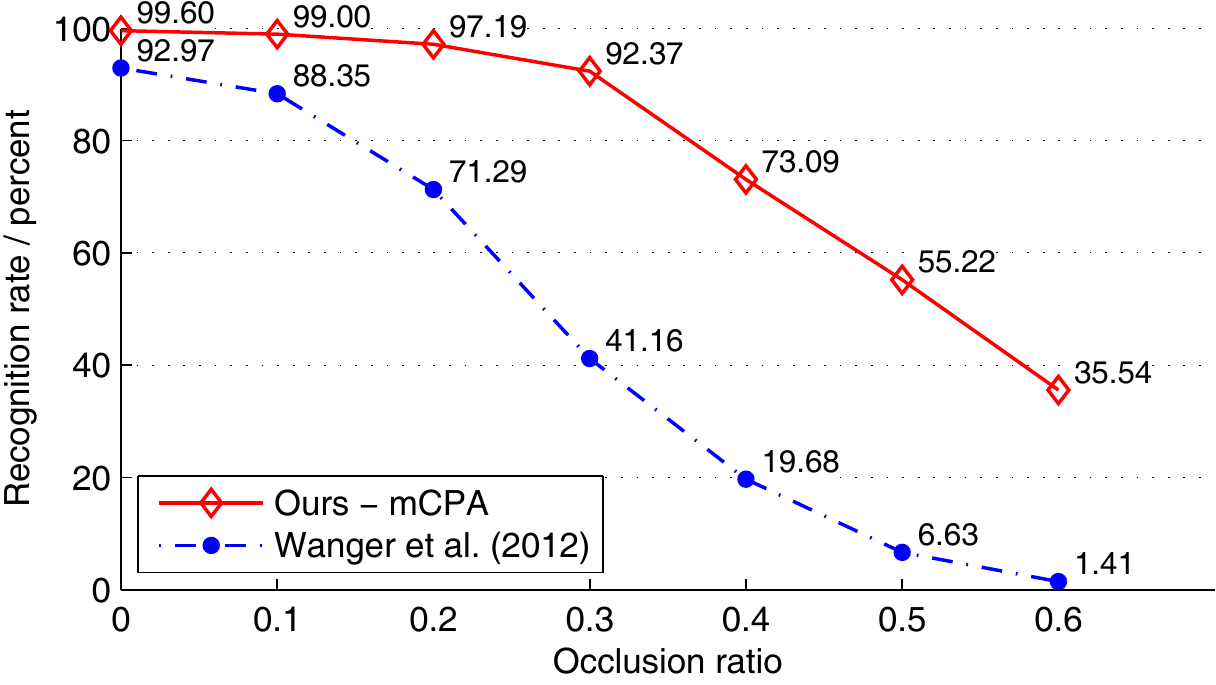}
\par\end{centering}

}
\par\end{centering}

\begin{centering}
\subfloat[Recognition rates for across-pose and neural-expression settings with
the occlusion ratio of 30\% \label{fig:occlusion-poses}]{\begin{centering}
\includegraphics[width=0.9\columnwidth]{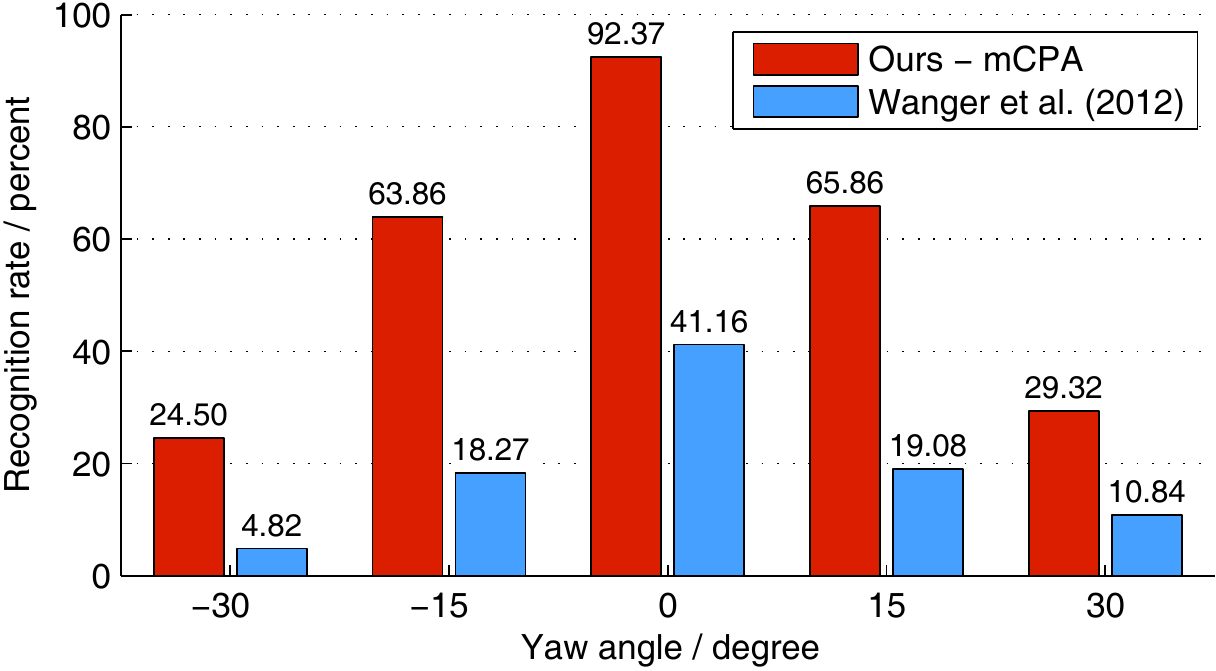}
\par\end{centering}

}
\par\end{centering}

\protect\caption{Recognition rates on Multi-PIE (first 249 subjects) when synthetic
random block occlusions are present }
\end{figure}

Fig.~\ref{fig:occlusion-levels} reports recognition rates on frontal-view
face images at different occlusion ratios, and Fig.~\ref{fig:occlusion-poses}
reports those of varying viewpoints at the occlusion ratio of 30\%.
Compared with \citep{andrew2012practical}, Fig.~\ref{fig:occlusion-levels}
and Fig.~\ref{fig:occlusion-poses} show that our CPA method is more
robust against partial occlusion. When a small portion (10\%) of face
images is occluded, recognition rate of \citep{andrew2012practical}
drops below 90\%. In contrast, recognition rates of CPA remain above
90\% when the occlusion ratios increase from 0\% to 30\%. For the
30\% occlusion ratio, recognition rate of CPA is more than 45\% higher
than that of \citep{andrew2012practical} for pose change of within
$\pm15^{\circ}$.

\subsection{Comparison with the State-of-the-art \label{sub:comparision}}

In this section, we used the Multi-PIE dataset to compare CPA with
the state-of-the-art Morphable Displacement Field (MDF) method \citep{eccv2012facepose}
for face recognition across pose. MDF achieves good performance by
learning an MDF model from a large number of 3D face shapes, while
training of CPA only requires 2D face images of a few subjects.

Like our CPA method, MDF can also be equipped with different features
and classifiers for recognition. In order fairly compare the two methods,
in this paper, we used the same LDA classifier for both the two methods~%
\footnote{Experimental results of MDF were produced by \citet{eccv2012facepose}.
As SRC was not implemented in their experiment pipeline, we thank
them for their kind help to run MDF with the alternative LDA classifier. %
}.  While LDA models were learned from pruned gallery in the CPA method,
a single LDA model was learned from the entire gallery set in advance
for the MDF method, as it does not prune gallery during recognition. 

We compare the CPA and MDF methods using probe face images that are
under different pose and illumination conditions from those of gallery
images. The experiment settings were largely the same as those of
face recognition across pose with different illumination used in Session~\ref{sub:pose-expression-MultiPIE},
except that: 
\begin{enumerate}
\item In order to be consistent with the experimental protocol in the work
of \citet{eccv2012facepose}, we used the last 137 subjects instead
of the first 229 ones in Multi-PIE for testing. More specifically,\textbf{
}for each subject, face images in the session where he/she first appears
were included in the gallery set, and those in the other sessions
were in the probe set. 
\item We learned the tree-structured shape model in CPA using face images
of 9 other subjects (subject id: \{038, 040, 041, 042, 043, 044, 046,
047, 048\}), who appear in all the four sessions of Multi-PIE. 
\item We set $P=10$ for the pruning scheme used in Algorithm~\ref{alg:recognition}. 
\end{enumerate}

Table~\ref{tab:soa-lda} reports recognition rates of CPA and MDF
on neutral-expression face images of varying degrees of pose change,
where both fully automatic and semi-automatic (manual initialization
with known pose) experiments are reported. For the case of semi-automatic
experiments using LDA as the classifier, our CPA method outperforms
MDF when the degrees of pose change are within $\pm\mbox{15}^{\circ}$.
When the degrees of pose change increase to $\pm\mbox{30}^{\circ}$,
CPA performs worse than MDF. The performance inconsistency between
smaller and larger degrees of pose change is in fact determined by
the algorithm nature of the two methods. CPA is based on 2D similarity
transform of individual facial parts, whose training only requires
face images of a few subjects, while a large number of 3D face shapes
are necessary for learning an MDF model. By learning more generic
knowledge of 3D face shapes, MDF is able to better cope with face
recognition across a larger degree of pose change. However, for the
range between $-15$ and $+15$ degrees, which are more often encountered
in practical scenarios such as access control, and face recognition
in which is also more reliable, CPA gives more accurate recognition
results than MDF does. In addition, our CPA method with SRC gives
similarly better results in this range of pose change.

We realize a fully automatic CPA method by initializing our algorithm
using \citet{viola2001adaboost}'s face detector, followed by a coarse
and holistic alignment using the method of \citet{andrew2012practical}.
For knowledge of face pose used in the mCPA model, we consider two
situations where poses of probe faces are either known in advance
or estimated by \citet{zhu2012dpmface}'s method. The accuracy rate
of Zhu et al.'s pose estimator is 96.56\% under our experiment settings~%
\footnote{It seems that Zhu et al.'s method \citep{zhu2012dpmface} can work
very well only when the training and test images are with the same
illumination. To handle the illumination variation in our experiments,
we normalized image illumination by the non-local means (NLM) based
method \citep{nlm-illumination} before pose estimation.%
}. 

Table~\ref{tab:comparison-automatic} reports recognition results
of fully automatic CPA. Fully automatic CPA (the last two rows) performs
comparably with the semi-automatic one (the first row) when pose changes
of probe face images are within $\pm\mbox{15}^{\circ}$. Note that
the fully automatic alignment works from coarse to fine granularity,
say, sequentially uses \citet{viola2001adaboost}'s face detector,
the method of \citet{andrew2012practical} for a holistic alignment,
and part-based alignment by CPA. The experimental results show that
this coarse-to-fine strategy works well for reasonable degrees of
pose change (e.g., within $\pm\mbox{15}^{\circ}$). For larger degrees
of pose change, performance of holistic alignment by \citep{andrew2012practical}
drops. Consequently, our method loses the chance of correcting its
alignment failure. In addition, the last two rows of Table~\ref{tab:comparison-automatic}
also tells that the mCPA model performs equally well when poses of
probe faces are either given or automatically estimated. This confirms
the effectiveness of our proposed method for automatic part-based
face alignment.

\begin{table}
\protect\caption{Face recognition across pose and expression with different illumination
on the last 137 subjects of Multi-PIE. \label{tab:comparision-poses-soa}
Gallery illuminations: \{0, 1, 7, 13, 14, 16, 18\}. Probe illumination:
10. }

\renewcommand\arraystretch{1.2}

\begin{centering}
\subfloat[Recognition rates of MDF and CPA when integrated with LDA. Initialization
is manually done by fitting eye-corner annotations, and pose is known
in advance. \label{tab:soa-lda}]{

\centering{}%
\begin{tabular}{c|ccccc}
\hline 
\multirow{2}{*}{Alignment} & $-\mbox{30}^{\circ}$ & $-\mbox{15}^{\circ}$ & $\mbox{0}^{\circ}$ & $+\mbox{15}^{\circ}$ & $+\mbox{30}^{\circ}$\tabularnewline
\cline{2-6} 
 & 13\_0 & 14\_0 & 05\_1 & 05\_0 & 04\_1\tabularnewline
\hline 
MDF & \textbf{68.71} & 82.21 & / & 80.37 & \textbf{74.85}\tabularnewline
mCPA & 38.65 & \textbf{85.28} & \textbf{98.77} & \textbf{86.50} & 47.24\tabularnewline
\hline 
\end{tabular}}
\par\end{centering}

\begin{centering}
\subfloat[Recognition rates of the CPA method integrated with SRC for different
initialization and pose estimation schemes. \label{tab:comparison-automatic}]{

\centering{}%
\begin{tabular}{cc|ccccc}
\hline 
Initial- & \multirow{2}{*}{Pose} & $-\mbox{30}^{\circ}$ & $-\mbox{15}^{\circ}$ & $\mbox{0}^{\circ}$ & $+\mbox{15}^{\circ}$ & $+\mbox{30}^{\circ}$\tabularnewline
\cline{3-7} 
ization &  & 13\_0 & 14\_0 & 05\_1 & 05\_0 & 04\_1\tabularnewline
\hline 
manual & known & 53.37 & 87.12 & 98.16 & 87.73 & 76.07\tabularnewline
auto\textsuperscript{{*}} & known & 32.52 & 82.82 & 99.39 & 94.48 & 57.06\tabularnewline
auto\textsuperscript{{*}} & auto\textsuperscript{{*}{*}} & 33.13 & 80.98 & 99.39 & 94.48 & 57.06\tabularnewline
\hline 
\multicolumn{7}{l}{\textsuperscript{{*}}~face detector followed by holistic alignment;
\textsuperscript{{*}{*}}~pose estimator.}\tabularnewline
\end{tabular}}
\par\end{centering}

\renewcommand\arraystretch{1}
\end{table}

\section{Conclusion}

In this paper we propose a method termed CPA for across-pose and -expression
face recognition, which can be benefited by pixel-wisely accurate
alignment. The CPA model consists of appearance evidence of each part
and a tree-structured shape model for constraining part deformation,
both of which can be automatically learned from training images. To
align a probe image, we fit its parts to the appearance evidence with
consideration of constraint from the learned tree-structured shape
model. This objective is formulated as a norm minimization problem
regularized by the graph likelihoods, which can be efficiently solved
by an alternating optimization method. CPA can easily incorporate
many existing face recognition method for part-based recognition.
Intensive experiments show the efficacy of CPA in handling illumination,
pose, and/or expression changes when integrated with an recognition
method robust to illumination changes. In further research, we are
interested in applying applying/adapting CPA to other computer vision
applications.

\section*{Acknowledgement}

Thanks to \citep{eccv2012facepose} for kindly producing the experimental
results on the MDF model that are reported in Table~\ref{tab:comparision-poses-soa}.

\appendices{}

\section{\label{sec:DT-Linear-System}Linear System in (\ref{eq:dt-linear-system})}

Let $Q=[q_{1},q_{2},\ldots,q_{m}]\in\mathbb{R}^{d\times m}$, (\ref{eq:dt-linear-system})
is the same as the linear equation array: 
\begin{equation}
G^{\langle i\rangle}\Delta\nu^{\langle i\rangle}+\eta\left(\dfrac{\partial g(\boldsymbol{\nu}+\Delta\boldsymbol{\nu},\mathcal{Z})}{\partial\Delta\nu^{\langle i\rangle}}\right)^{T}=q_{i},\label{eq:dt-linear-sub}
\end{equation}
for $i=1,2,\ldots,m$, which can be written in a more standard form
\begin{equation}
\begin{pmatrix}W_{11} & W_{12} & \cdots & W_{1m}\\
W_{21} & W_{22} & \cdots & W_{2m}\\
\vdots & \vdots & \ddots & \vdots\\
W_{m1} & W_{m2} & \cdots & W_{mm}
\end{pmatrix}\begin{pmatrix}\Delta\nu^{\langle1\rangle}\\
\Delta\nu^{\langle2\rangle}\\
\vdots\\
\Delta\nu^{\langle m\rangle}
\end{pmatrix}=\begin{pmatrix}c_{1}\\
c_{2}\\
\vdots\\
c_{m}
\end{pmatrix},
\end{equation}
where $c_{i}\in\mathbb{R}^{d}$ is a column vector, and $W_{ij}\in\mathbb{R}^{d\times d}$.
Recall that $\mathcal{E}$ is the edge set of the tree-structured
model. In particular, if $i$ is the parent of $j$, it holds $(i,j)\in\mathcal{E}$.
Now, we can find that 
\[
c_{i}=q_{i}+\eta\left(\Lambda^{\langle i\rangle}(\mu^{\langle i\rangle}-\nu_{\delta}^{\langle i\rangle})-\sum_{j\in\{(i,j)\in\mathcal{E}\}}\Lambda^{\langle j\rangle}(\mu^{\langle j\rangle}-\nu_{\delta}^{\langle j\rangle})\right),
\]
for $i=1,2,\ldots,m$, and 
\[
W_{ij}=\begin{cases}
G^{\langle i\rangle}+\eta\sum_{l\in\{i\}\cup\{l:(i,l)\in\mathcal{E}\}}\Lambda^{\langle l\rangle}, & j=i,\\
-\eta\Lambda^{\langle j\rangle}, & (i,j)\in\mathcal{E},\\
-\eta\Lambda^{\langle i\rangle}, & (j,i)\in\mathcal{E},\\
\mathbf{0}, & \mbox{otherwise},
\end{cases}
\]
for for $i,j=1,2,\ldots,m$.

\section{\label{sec:solving-transforms}Solving Transformations in (\ref{eq:CPA-obj-holistic-2})
with Fixed Parts}

$\mathbb{G}$ is the 2-D similarity group parametrized in $\mathbb{R}^{d}$
($d=4$) as it is defined in (\ref{eq:similarity-para}). Given initial
values on $\sigma\in\mathbb{G}$ and $\{\nu^{\langle i\rangle}\}_{i=1}^{m}\in\mathbb{G}$,
(\ref{eq:CPA-obj-holistic-2}) seeks the minimum of $g\left(\boldsymbol{\nu},\mathcal{Z}\right)$
while keeping the value of $\sigma\circ\nu^{\langle i\rangle}$ unchanged.
Let $\boldsymbol{\nu}=[\nu^{\langle1\rangle},\nu^{\langle2\rangle},\ldots,\nu^{\langle m\rangle}]\in\mathbb{R}^{d\times m}$,
and $\Delta\boldsymbol{\nu}=[\Delta\nu^{\langle1\rangle},\Delta\nu^{\langle2\rangle},\ldots,\Delta\nu^{\langle m\rangle}]\in\mathbb{R}^{d\times m}$.
We reformulate (\ref{eq:CPA-obj-holistic-2}) as 
\begin{gather}
\min_{\Delta\boldsymbol{\nu},\Delta\sigma}g\left(\boldsymbol{\nu}+\Delta\mbox{\ensuremath{\boldsymbol{\nu}}},\mathcal{Z}\right),\label{eq:CPA-obj-holistic-3}\\
\mathrm{s.t.}\quad(\sigma+\Delta\mbox{\ensuremath{\sigma}})\circ(\nu^{\langle i\rangle}+\Delta\mbox{\ensuremath{\nu}}^{\langle i\rangle})=\sigma\circ\nu^{\langle i\rangle}.\label{eq:CPA-constraint-holistic-3}
\end{gather}
After optimizing it, $\sigma$ is updated to $\sigma+\Delta\sigma$,
and $\boldsymbol{\nu}$ is updated to $\boldsymbol{\nu}+\Delta\boldsymbol{\nu}$.

Let 
\begin{alignat*}{2}
 & \mbox{\ensuremath{\nu}}^{\langle i\rangle} & \,=\; & (s^{\langle i\rangle},\theta^{\langle i\rangle},t_{u}^{\langle i\rangle},t_{v}^{\langle i\rangle}),\\
 & \Delta\sigma & \,=\; & (\delta s^{*},\delta\theta^{*},\delta t_{u}^{*},\delta t_{v}^{*}),
\end{alignat*}
where $s^{\cdot},\theta^{\cdot},t_{u}^{\cdot},t_{v}^{\cdot}$ (``$\cdot$''
for ``$\langle i\rangle$'' and ``$*$'') respectively denote
the scale, rotation, horizontal translation, and vertical translation
of a 2-D similarity transformation. (\ref{eq:CPA-constraint-holistic-3})
is equivalent to 
\begin{equation}
\Delta\mbox{\ensuremath{\nu}}^{\langle i\rangle}=\left(\begin{array}{c}
-\delta s^{*}\\
-\delta\theta^{*}\\
-t_{u}^{\langle i\rangle}+f_{u}^{*}+f_{u}^{\langle i\rangle}\\
-t_{v}^{\langle i\rangle}+f_{v}^{*}+f_{v}^{\langle i\rangle}
\end{array}\right),\label{eq:hoslistic-partial-explicit}
\end{equation}
where, 
\begin{alignat*}{2}
f_{u}^{*}= &  & \frac{(t_{u}^{*},t_{v}^{*})}{\exp(\delta s^{*}+s^{*})}\left(\begin{array}{c}
\cos(\delta\theta^{*}+\theta^{*})\\
\sin(\delta\theta^{*}+\theta^{*})
\end{array}\right),\\
f_{v}^{*}= &  & \frac{(t_{v}^{*},-t_{u}^{*})}{\exp(\delta s^{*}+s^{*})}\left(\begin{array}{c}
\cos(\delta\theta^{*}+\theta^{*})\\
\sin(\delta\theta^{*}+\theta^{*})
\end{array}\right),
\end{alignat*}
and, 
\begin{alignat*}{2}
f_{u}^{\langle i\rangle}= &  & \frac{(t_{u}^{\langle i\rangle},t_{v}^{\langle i\rangle})}{\exp(\delta s^{*})}\left(\begin{array}{c}
\cos\delta\theta^{*}\\
\sin\delta\theta^{*}
\end{array}\right),\\
f_{v}^{\langle i\rangle}= &  & \frac{(t_{v}^{\langle i\rangle},-t_{u}^{\langle i\rangle})}{\exp(\delta s^{*})}\left(\begin{array}{c}
\cos\delta\theta^{*}\\
\sin\delta\theta^{*}
\end{array}\right).
\end{alignat*}
Substituting (\ref{eq:hoslistic-partial-explicit}) into the objective
function in (\ref{eq:CPA-obj-holistic-3}), we obtain the unconstrained
equivalence of the original problem, say, $\min_{\Delta\sigma}g\left(\boldsymbol{\nu}+\Delta\mbox{\ensuremath{\boldsymbol{\nu}}},\mathcal{Z}\right)$,
where $\Delta\boldsymbol{\nu}$ is expended as (\ref{eq:hoslistic-partial-explicit}).
We solve this problem by gradient descent. Now, we need to find $g\left(\boldsymbol{\nu}+\Delta\mbox{\ensuremath{\boldsymbol{\nu}}},\mathcal{Z}\right)$'s
gradient in $\Delta\sigma$.

Recall that $\mathcal{E}$ is the tree edge set. For $i=1,2,\ldots,m$,
let $j$ satisfies $(j,i)\in\mathcal{E}$ ($j$ is the parent of $i$),
and 
\[
\varphi^{\langle i\rangle}=\left((\mbox{\ensuremath{\nu}}^{\langle i\rangle}+\Delta\mbox{\ensuremath{\nu}}^{\langle i\rangle})-(\mbox{\ensuremath{\nu}}^{\langle i\rangle}+\Delta\mbox{\ensuremath{\nu}}^{\langle j\rangle})-\mu^{\langle i\rangle}\right),
\]
We then have 
\begin{equation}
g\left(\boldsymbol{\nu}+\Delta\mbox{\ensuremath{\boldsymbol{\nu}}},\mathcal{Z}\right)=\frac{1}{2}\sum_{i=1}^{m}{\varphi^{\langle i\rangle}}^{T}\Lambda^{\langle i\rangle}\varphi^{\langle i\rangle}+b,
\end{equation}
and, 
\begin{equation}
\nabla_{(\Delta\sigma)}g\left(\boldsymbol{\nu}+\Delta\mbox{\ensuremath{\boldsymbol{\nu}}},\mathcal{Z}\right)=\sum_{i=1}^{m}{\varphi^{\langle i\rangle}}^{T}\Lambda^{\langle i\rangle}\frac{\partial\varphi^{\langle i\rangle}}{\partial(\Delta\sigma)}.
\end{equation}
For $i$ satisfying $(0,i)\in\mathcal{E}$ ($i$ is directly linked
with the root), it holds

\[
\varphi^{\langle i\rangle}=\left(\begin{array}{c}
s^{\langle i\rangle}-\delta s^{*}\\
\theta^{\langle i\rangle}-\delta\theta^{*}\\
f_{u}^{\langle i\rangle}-f_{u}^{*}\\
f_{v}^{\langle i\rangle}-f_{v}^{*}
\end{array}\right)-\mu^{\langle i\rangle},
\]

\begin{align*}
 & \frac{\partial\varphi^{\langle i\rangle}}{\partial(\Delta\sigma)}=\\
 & \;\left(\begin{array}{cccc}
-1 & 0 & 0 & 0\\
0 & -1 & 0 & 0\\
f_{u}^{*}-f_{u}^{\langle i\rangle} & f_{v}^{\langle i\rangle}-f_{v}^{*} & -\frac{\cos(\delta\theta^{*}+\theta^{*})}{\exp(\delta s^{*}+s^{*})} & -\frac{\sin(\delta\theta^{*}+\theta^{*})}{\exp(\delta s^{*}+s^{*})}\\
f_{v}^{*}-f_{v}^{\langle i\rangle} & f_{u}^{*}-f_{u}^{\langle i\rangle} & \frac{\sin(\delta\theta^{*}+\theta^{*})}{\exp(\delta s^{*}+s^{*})} & -\frac{\cos(\delta\theta^{*}+\theta^{*})}{\exp(\delta s^{*}+s^{*})}
\end{array}\right).
\end{align*}
For $i$ satisfying $(j,i)\in\mathcal{E},j\neq0$ ($i$ is not directly
linked with the root), it holds 
\[
\varphi^{\langle i\rangle}=\left(\begin{array}{c}
s^{\langle i\rangle}-s^{\langle j\rangle}\\
\theta^{\langle i\rangle}-\theta^{\langle j\rangle}\\
f_{u}^{\langle i\rangle}-f_{u}^{\langle j\rangle}\\
f_{v}^{\langle i\rangle}-f_{v}^{\langle j\rangle}
\end{array}\right)-\mu^{\langle i\rangle},
\]

\[
\frac{\partial\varphi^{\langle i\rangle}}{\partial(\Delta\sigma)}=\left(\begin{array}{cccc}
0 & 0 & 0 & 0\\
0 & 0 & 0 & 0\\
f_{u}^{\langle j\rangle}-f_{u}^{\langle i\rangle} & f_{v}^{\langle i\rangle}-f_{v}^{\langle j\rangle} & 0 & 0\\
f_{v}^{\langle j\rangle}-f_{v}^{\langle i\rangle} & f_{u}^{\langle j\rangle}-f_{u}^{\langle i\rangle} & 0 & 0
\end{array}\right).
\]

\section{\label{sec:mpBatch-solution}Solution to (\ref{eq:mpBatch-obj})}

We solve (\ref{eq:mpBatch-obj}) by the alternately conducting the
following two steps: 
\begin{enumerate}
\item Fix $\boldsymbol{\sigma}$ and solve $\check{\boldsymbol{\nu}},\{A^{\langle i\rangle}\}_{i=1}^{m},\{E^{\langle i\rangle}\}_{i=1}^{m}$. 
\item Fix $\{A^{\langle i\rangle}\}_{i=1}^{m},\{E^{\langle i\rangle}\}_{i=1}^{m}$
and solve $\boldsymbol{\sigma},\check{\boldsymbol{\nu}}$. 
\end{enumerate}

\textbf{Step 1}: Given fixed $\boldsymbol{\sigma}$, we update $\check{\boldsymbol{\nu}}$
by a generalization of the Gauss-Newton method. To be specific, for
a linear update from $\check{\boldsymbol{\nu}}$ to $\check{\boldsymbol{\nu}}+\Delta\check{\boldsymbol{\nu}}$
($\Delta\check{\boldsymbol{\nu}}\in\mathbb{R}^{d\times m\times n}$),
we approximate the equality constraint in (\ref{eq:mpBatch-obj})
by its first-order Taylor expansion at $T^{\langle i\rangle}(\check{\boldsymbol{\nu}})$,
i.e., $D\circ\boldsymbol{\sigma}\circ T^{\langle i\rangle}(\check{\boldsymbol{\nu}})\approx D\circ\boldsymbol{\sigma}\circ T^{\langle i\rangle}(\check{\boldsymbol{\nu}})+\sum_{k=1}^{n}J_{k}^{\langle i\rangle}T^{\langle i\rangle}(\Delta\check{\boldsymbol{\nu}})\epsilon_{k}\epsilon_{k}^{T}$,
where $J_{k}^{\langle i\rangle}\doteq\partial(d_{k}\circ\sigma_{k}\circ\nu_{k}^{\langle i\rangle})/\partial\nu_{k}^{\langle i\rangle}$
is the Jacobian w.r.t.\ $\nu_{k}^{\langle i\rangle}$, and $\{\epsilon_{k}\}_{k=1}^{n}$
denotes the standard basis of $\mathbb{R}^{n}$. The above linearization
leads to the following problem to optimize $\{A^{\langle i\rangle}\}_{i=1}^{m},\{E^{\langle i\rangle}\}_{i=1}^{m},\Delta\check{\boldsymbol{\nu}}$
\begin{align}
\min_{\underset{i=1,2,\ldots,m}{A^{\langle i\rangle},E^{\langle i\rangle},\Delta\check{\boldsymbol{\nu}}}} & \Biggl\{\sum_{i=1}^{m}\left(\Vert A^{\langle i\rangle}\Vert_{*}+\lambda^{\langle i\rangle}\Vert E^{\langle i\rangle}\Vert_{1}\right)\nonumber \\
 & +\eta\sum_{k=1}^{n}g\left(T_{k}(\check{\boldsymbol{\nu}}+\Delta\check{\boldsymbol{\nu}}),\mathcal{Z}\right)\Biggl\}\label{eq:mpBatch-obj-linearized}
\end{align}
\begin{gather*}
\mathrm{s.t.}\quad D\circ\boldsymbol{\sigma}\circ T^{\langle i\rangle}(\check{\boldsymbol{\nu}})+\sum_{k=1}^{n}J_{k}^{\langle i\rangle}T^{\langle i\rangle}(\Delta\check{\boldsymbol{\nu}})\epsilon_{k}\epsilon_{k}^{T}=A^{\langle i\rangle}+E^{\langle i\rangle}.
\end{gather*}
We repeatedly solve (\ref{eq:mpBatch-obj-linearized}) to update $\check{\boldsymbol{\nu}}$,
until it converges to a local minimum, which gives the solution to
the original problem(\ref{eq:mpBatch-obj}) .

We solve the convex problem (\ref{eq:mpBatch-obj-linearized}) by
adapting the ALM. Let 
\begin{align*}
h( & A^{\langle i\rangle},E^{\langle i\rangle},T^{\langle i\rangle}(\Delta\check{\boldsymbol{\nu}}))=\\
 & D\circ\boldsymbol{\sigma}\circ T^{\langle i\rangle}(\check{\boldsymbol{\nu}})+\sum_{k=1}^{n}J_{k}^{\langle i\rangle}T^{\langle i\rangle}(\Delta\check{\boldsymbol{\nu}})\epsilon_{k}\epsilon_{k}^{T}-A^{\langle i\rangle}-E^{\langle i\rangle}.
\end{align*}
The augmented Lagrange function is written as 
\begin{align}
 & L_{\beta}(\{A^{\langle i\rangle}\}_{i=1}^{m},\{E^{\langle i\rangle}\}_{i=1}^{m},\Delta\check{\boldsymbol{\nu}},\{\Gamma^{\langle i\rangle}\}_{i=1}^{m})=\\
 & \sum_{i}^{m}\biggl\{\Vert A^{\langle i\rangle}\Vert_{*}+\lambda^{\langle i\rangle}\Vert E^{\langle i\rangle}\Vert_{1}+\left\langle \Gamma^{\langle i\rangle},h(A^{\langle i\rangle},E^{\langle i\rangle},T^{\langle i\rangle}(\Delta\check{\boldsymbol{\nu}}))\right\rangle \nonumber \\
 & +\frac{\mu}{2}\left\Vert h(A^{\langle i\rangle},E^{\langle i\rangle},T^{\langle i\rangle}(\Delta\check{\boldsymbol{\nu}}))\right\Vert _{F}^{2}\biggl\}+\sum_{k=1}^{n}g\left(T_{k}(\check{\boldsymbol{\nu}}+\Delta\check{\boldsymbol{\nu}}),\mathcal{Z}\right),\nonumber 
\end{align}
where $\{\Gamma^{\langle i\rangle}\}_{i=1}^{m}$ are the Lagrange
multiplier matrices, and the matrix inner product $\langle\cdot,\cdot\rangle$
is defined as $\langle A,B\rangle=\operatorname{trace}(A^{T}B)$.
Given initial $\{\Gamma^{\langle i\rangle}\}_{i=1}^{m}$, ALM iteratively
and alternately updates $\{A^{\langle i\rangle}\}_{i=1}^{m},\{E^{\langle i\rangle}\}_{i=1}^{m},\Delta\check{\boldsymbol{\nu}}$,
and $\{\Gamma^{\langle i\rangle}\}_{i=1}^{m}$ by 
\begin{enumerate}
\item $\{A^{\langle i\rangle}\}_{i=1}^{m},\{E^{\langle i\rangle}\}_{i=1}^{m},\Delta\check{\boldsymbol{\nu}}\leftarrow$\vspace{-0.4em}

\begin{equation}
\mbox{\hspace{-2em}}\underset{\underset{i=1,2,\ldots,m}{A^{\langle i\rangle},E^{\langle i\rangle},\Delta\check{\boldsymbol{\nu}}}}{\arg\min}\, L_{\beta}(\{A^{\langle i\rangle}\}_{i=1}^{m},\{E^{\langle i\rangle}\}_{i=1}^{m},\Delta\check{\boldsymbol{\nu}},\{\Gamma^{\langle i\rangle}\}_{i=1}^{m});\label{eq:mpBatch-alm-iter}
\end{equation}

\item $\Gamma^{\langle i\rangle}\leftarrow\Gamma^{\langle i\rangle}+\beta_{l}\cdot h(A^{\langle i\rangle},E^{\langle i\rangle},T^{\langle i\rangle}(\Delta\check{\boldsymbol{\nu}}))$,
\\
 for $i=1,2,\ldots,m;$ 
\end{enumerate}
where $l$ is the iteration number, and $\{\beta_{l}\}_{l=1,2,\ldots}$
is an sequence increasing to sufficient large.

Directly solving (\ref{eq:mpBatch-alm-iter}) w.r.t.\ all the unknown
variables is still a difficult problem. Instead, we alternately update
them in three groups, i.e., $\{A^{\langle i\rangle}\}_{i=1}^{m}$,
$\{E^{\langle i\rangle}\}_{i=1}^{m}$, and $\Delta\check{\boldsymbol{\nu}}$,
so that each subproblem associated with any single group of variables
has a closed form solution. More specifically, for $i=1,2,\ldots,m$,
we update $A^{\langle i\rangle}$ and $E^{\langle i\rangle}$ sequentially
by 
\begin{align*}
R^{\langle i\rangle} & \leftarrow D\circ\boldsymbol{\sigma}\circ T^{\langle i\rangle}(\check{\boldsymbol{\nu}})+(1/\beta_{l})\Gamma^{\langle i\rangle},\\
(\hat{U}, & \hat{\Sigma},\hat{V})\leftarrow\operatorname{svd}\left(R^{\langle i\rangle}+\sum_{k}^{n}J_{k}^{\langle i\rangle}T^{\langle i\rangle}(\Delta\check{\boldsymbol{\nu}})\epsilon_{k}\epsilon_{k}^{T}-E^{\langle i\rangle}\right),\\
A^{\langle i\rangle} & \leftarrow\hat{U}\mathcal{S}_{1/\beta_{l}}(\hat{\Sigma}){\hat{V}}^{T},\\
E^{\langle i\rangle} & \leftarrow\mathcal{S}_{\lambda^{\langle i\rangle}/\beta_{l}}\left(R^{\langle i\rangle}+\sum_{k}^{n}J_{k}^{\langle i\rangle}T^{\langle i\rangle}(\Delta\check{\boldsymbol{\nu}})\epsilon_{k}\epsilon_{k}^{T}-A^{\langle i\rangle}\right),
\end{align*}
where $R^{\langle i\rangle}$ is an auxiliary variable used for notation
convenience, $\mathcal{S}_{\alpha}(\cdot)$ is the soft-thresholding
function defined in (\ref{eq:soft-thresholding}). Then, for $k=1,2,\ldots,n$,
we find the optimum $T_{k}(\Delta\check{\boldsymbol{\nu}})$ by solving
\begin{align}
\mathbf{0}= & \frac{\partial L_{\mu}(\{A^{\langle i\rangle}\}_{i=1}^{m},\{E^{\langle i\rangle}\}_{i=1}^{m},\Delta\check{\boldsymbol{\nu}})}{\partial\Delta\nu_{k}^{\langle i\rangle}}\label{eq: delta-vu-check}\\
= & \beta_{l}\left((R^{\langle i\rangle}-A^{\langle i\rangle}-E^{\langle i\rangle})\epsilon_{k}+J_{k}^{\langle i\rangle}\Delta\nu_{k}^{\langle i\rangle}\right){J_{k}^{\langle i\rangle}}\nonumber \\
 & +\eta\frac{\partial g(T_{k}(\check{\boldsymbol{\nu}}),\mathcal{Z})}{\partial\Delta\nu_{k}^{\langle i\rangle}},\nonumber 
\end{align}
for $i=1,2,\ldots,m$. The $m$ equations forms a sparse linear system
of the form in (\ref{eq:dt-linear-system}) (expended form in Appendix~\ref{sec:DT-Linear-System}),
whose parameters are $\{G^{\langle i\rangle}\}_{i=1}^{m}$ and $Q$.
Here, $G^{\langle i\rangle}=\beta_{l}{J_{k}^{\langle i\rangle}}^{T}J_{k}^{\langle i\rangle}$,
the $i^{\textrm{th}}$ column of $Q$ is $\beta_{l}(A^{\langle i\rangle}+E^{\langle i\rangle}-R^{\langle i\rangle})\epsilon_{k}$.
Still, we use the inexact scheme for ALM, in which $\{A^{\langle i\rangle}\}_{i=1}^{m}$,
$\{E^{\langle i\rangle}\}_{i=1}^{m}$, and $\Delta\check{\boldsymbol{\nu}}$
are alternately updated for only once in each iteration of ALM.

\textbf{Step 2}: Given fixed $\{A^{\langle i\rangle}\}_{i=1}^{m},\{E^{\langle i\rangle}\}_{i=1}^{m}$,
$\sigma_{k}\circ\nu_{k}^{\langle i\rangle}$ can be taken as constant
for any possible $i$ and $k$. For every $k=1,2,\ldots,n$, we use
the same technique as in Session~\ref{sub:CPA-holistic} to solve
$\sigma_{k}$ and $T^{\langle i\rangle}(\check{\boldsymbol{\nu}})$.
More specifically, denote $\zeta_{k}^{\langle i\rangle}=\sigma_{k}\circ\nu_{k}^{\langle i\rangle}$,
which should not be changed, and update $\sigma_{k}$ and $T_{k}(\check{\boldsymbol{\nu}})$
by 
\begin{equation}
\min_{T_{k}(\check{\boldsymbol{\nu}}),\sigma_{k}}g\left(T_{k}(\check{\boldsymbol{\nu}}),\mathcal{Z}\right)\quad\mathrm{s.t.}\;\sigma_{k}\circ\nu_{k}^{\langle i\rangle}=\zeta_{k}^{\langle i\rangle}.
\end{equation}
The solution to this type of problem is present in Appendix~\ref{sec:solving-transforms}

By the way, $\boldsymbol{\sigma}$ might be initialized by perfect
manual annotations in practice for part dictionary learning. In this
case, we may fairly take it as a known variable the original problem
(\ref{eq:mpRASL-obj}) so that the algorithm efficiency can be improved
by reducing an alternating loop.

\section{Maximum a Posteriori with Gaussian-Wishart Prior \label{sec:MAP-GW-Prior}}

The PDF of the Wishart distribution is 
\begin{gather*}
\mathcal{W}(\Lambda|V,r)=\frac{1}{\rho}|\Lambda|^{(r-d-1)/2}\exp\left(-\frac{1}{2}\operatorname{Tr}(\Lambda V^{-1})\right),\\
\mathrm{s.t.}\quad\rho=2^{rd/2}|V|^{r/2}\Gamma_{d}(r/2),
\end{gather*}
where $\Gamma_{d}(\cdot)$ is the $d$-D Gamma function. The PDF of
the Gaussian-Wishart distribution is 
\[
\Theta(\mu,\Lambda|u,\kappa,V,r)=\mathcal{N}(\mu|u,(\kappa\Lambda)^{-1})\cdot\mathcal{W}(\Lambda|V,r).
\]
where $\mathcal{N}(\cdot|\cdot,\cdot)$ denotes the Gaussian PDF defined
in (\ref{eq:norm-PDF}). The Gaussian-Wishart distribution is the
conjugate prior of the Gaussian distribution, where ``conjugate''
means that the prior and the corresponding posterior follow the same
type of distribution only with different parameters.

Given $n$ observations $\{\nu_{k}\}_{k=1}^{n}$ drawn from $\mathcal{N}(\mu,\Lambda^{-1})$,
let $\mu_{\mathrm{ML}},\Lambda_{\mathrm{ML}}$ denote the ML estimations
for $\mu,\Lambda$. Now, taking the Gaussian-Wishart distribution
$\Theta(u_{0},\kappa_{0},V_{0},r_{0})$ as the prior for $\mu,\Lambda$,
their posterior is 
\[
p(\mu,\Lambda^{-1}|\{\nu_{k}\}_{k=1}^{n},u_{0},\kappa_{0},V_{0},r_{0})=\Theta(\mu,\Lambda|u_{n},\kappa_{n},V_{n},r_{n}),
\]
where, 
\begin{alignat*}{2}
 & r_{n} & \,=\; & r_{0}+n,\\
 & \kappa_{n} & \,=\; & \kappa_{0}+n,\\
 & u_{n} & \,=\; & \frac{\kappa_{0}u_{0}+n\mu_{\mathrm{ML}}}{\kappa_{0}^{\langle i\rangle}+n},\\
 & V_{n} & \,=\; & \left({V_{0}^{\langle i\rangle}}^{-1}+n\Lambda_{\mathrm{ML}}^{-1}+\frac{\kappa_{0}n}{\kappa_{0}+n}H\right)^{-1},\\
 & H & \,=\; & (\mu_{\mathrm{ML}}-u_{0})(\mu_{\mathrm{ML}}-u_{0})^{T}.
\end{alignat*}
By finding the maximum of $\Theta(\mu,\Lambda^{-1}|u_{n},\kappa_{n},V_{n},r_{n})$,
we obtained the MAP estimations of $\mu,\Lambda$: 
\begin{align*}
\Lambda_{\mathrm{MAP}}=\; & (r_{0}+n-d)V_{n},\\
\mu_{\mathrm{MAP}}=\; & u_{n}.
\end{align*}
Inversely, taking $n=0$, we can set $\Theta(u_{0},\kappa_{0},V_{0},r_{0})$
to be consistent with specific Gaussian distribution $\mathcal{N}(\mu_{0},\Lambda_{0}^{-1})$,
say, 
\begin{align*}
u_{0} & =\mu_{0},\\
V_{0} & =\Lambda_{0}^{-1}/(r_{0}-d).
\end{align*}
In addition,  note  that using the prior $\Theta(u_{0},\kappa_{0},V_{0},r_{0})$
equals to incorporating $r_{0}$ (or $\kappa_{0}$) additional samples
drawn from the Gaussian distribution $\mathcal{N}\left(u_{0},\left((r_{0}-d)V_{0}\right)^{-1}\right)$
into the existing observations, where $r_{0}$ and $\kappa_{0}$ should
be normally set to the same value. In view of this, we set $r_{0}=\kappa_{0}=\vartheta n$,
where $\vartheta>0$ determines the weight of the prior w.r.t.\ the
observations. 

\bibliographystyle{IEEETranN}
\bibliography{submission}

\begin{thebibliography}{46}
\providecommand{\natexlab}[1]{#1}
\providecommand{\url}[1]{#1}
\csname url@samestyle\endcsname
\providecommand{\newblock}{\relax}
\providecommand{\bibinfo}[2]{#2}
\providecommand{\BIBentrySTDinterwordspacing}{\spaceskip=0pt\relax}
\providecommand{\BIBentryALTinterwordstretchfactor}{4}
\providecommand{\BIBentryALTinterwordspacing}{\spaceskip=\fontdimen2\font plus
\BIBentryALTinterwordstretchfactor\fontdimen3\font minus
  \fontdimen4\font\relax}
\providecommand{\BIBforeignlanguage}[2]{{%
\expandafter\ifx\csname l@#1\endcsname\relax
\typeout{** WARNING: IEEEtranN.bst: No hyphenation pattern has been}%
\typeout{** loaded for the language `#1'. Using the pattern for}%
\typeout{** the default language instead.}%
\else
\language=\csname l@#1\endcsname
\fi
#2}}
\providecommand{\BIBdecl}{\relax}
\BIBdecl

\bibitem[Zhao et~al.(2003)Zhao, Chellappa, Phillips, and
  Rosenfeld]{ZhaoFaceRecogSurvey}
W.~Zhao, R.~Chellappa, P.~J. Phillips, and A.~Rosenfeld, ``Face recognition: A
  literature survey,'' \emph{ACM Comput. Surv.}, vol.~35, no.~4, pp. 399--458,
  Dec. 2003.

\bibitem[Lee et~al.(2005)Lee, Ho, and Kriegman]{NS}
K.-C. Lee, J.~Ho, and D.~Kriegman, ``Acquiring linear subspaces for face
  recognition under variable lighting,'' \emph{IEEE Transactions on Pattern
  Analysis and Machine Intelligence}, vol.~27, no.~5, pp. 684 --698, may 2005.

\bibitem[Wagner et~al.(2012)Wagner, Wright, Ganesh, Zhou, Mobahi, and
  Ma]{andrew2012practical}
A.~Wagner, J.~Wright, A.~Ganesh, Z.~Zhou, H.~Mobahi, and Y.~Ma, ``Toward a
  practical face recognition system: Robust alignment and illumination by
  sparse representation,'' \emph{IEEE Transactions on Pattern Analysis and
  Machine Intelligence}, vol.~34, no.~2, pp. 372 --386, feb. 2012.

\bibitem[Belhumeur and Kriegman(1998)]{IllumCone}
P.~Belhumeur and D.~Kriegman, ``\BIBforeignlanguage{English}{What is the set of
  images of an object under all possible illumination conditions?}''
  \emph{\BIBforeignlanguage{English}{International Journal of Computer
  Vision}}, vol.~28, pp. 245--260, 1998.

\bibitem[Georghiades et~al.(2001)Georghiades, Belhumeur, and
  Kriegman]{FromFewToMany}
A.~Georghiades, P.~Belhumeur, and D.~Kriegman, ``From few to many: illumination
  cone models for face recognition under variable lighting and pose,''
  \emph{IEEE Transactions on Pattern Analysis and Machine Intelligence},
  vol.~23, no.~6, pp. 643 --660, jun 2001.

\bibitem[Chen et~al.(2006)Chen, Yin, Zhou, Comaniciu, and Huang]{ChenIllumInv}
T.~Chen, W.~Yin, X.~S. Zhou, D.~Comaniciu, and T.~Huang, ``Total variation
  models for variable lighting face recognition,'' \emph{IEEE Transactions on
  Pattern Analysis and Machine Intelligence}, vol.~28, no.~9, pp. 1519 --1524,
  sept. 2006.

\bibitem[Zhou et~al.(2007)Zhou, Aggarwal, Chellappa, and Jacobs]{ZhouIllumInv}
S.~Zhou, G.~Aggarwal, R.~Chellappa, and D.~Jacobs, ``Appearance
  characterization of linear lambertian objects, generalized photometric
  stereo, and illumination-invariant face recognition,'' \emph{IEEE
  Transactions on Pattern Analysis and Machine Intelligence}, vol.~29, no.~2,
  pp. 230 --245, feb. 2007.

\bibitem[Pentland et~al.(1994)Pentland, Moghaddam, and
  Starner]{PentlandMultiView}
A.~Pentland, B.~Moghaddam, and T.~Starner, ``View-based and modular eigenspaces
  for face recognition,'' in \emph{IEEE Computer Society Conference on Computer
  Vision and Pattern Recognition, 1994. Proceedings CVPR '94.}, jun 1994, pp.
  84 --91.

\bibitem[Beymer(1994)]{BeymerVaryingPose}
D.~Beymer, ``Face recognition under varying pose,'' in \emph{IEEE Computer
  Society Conference on Computer Vision and Pattern Recognition, 1994.
  Proceedings CVPR '94.}, jun 1994, pp. 756 --761.

\bibitem[Prince et~al.(2008)Prince, Warrell, Elder, and
  Felisberti]{pami2008tied-factor}
S.~Prince, J.~Warrell, J.~Elder, and F.~Felisberti, ``Tied factor analysis for
  face recognition across large pose differences,'' \emph{IEEE Transactions on
  Pattern Analysis and Machine Intelligence}, vol.~30, no.~6, pp. 970 --984,
  june 2008.

\bibitem[Nagesh and Li(2009)]{cvpr2009expresion-invariant}
P.~Nagesh and B.~Li, ``A compressive sensing approach for expression-invariant
  face recognition,'' in \emph{IEEE Conference on Computer Vision and Pattern
  Recognition, 2009. CVPR 2009.}, june 2009, pp. 1518 --1525.

\bibitem[Gross et~al.(2002)Gross, Matthews, and Baker]{fg2002light-field}
R.~Gross, I.~Matthews, and S.~Baker, ``Eigen light-fields and face recognition
  across pose,'' in \emph{Fifth IEEE International Conference on Automatic Face
  and Gesture Recognition, 2002. Proceedings.}, may 2002, pp. 1 --7.

\bibitem[Chai et~al.(2007)Chai, Shan, Chen, and Gao]{tip2007local-regression}
X.~Chai, S.~Shan, X.~Chen, and W.~Gao, ``Locally linear regression for
  pose-invariant face recognition,'' \emph{IEEE Transactions on Image
  Processing}, vol.~16, no.~7, pp. 1716 --1725, july 2007.

\bibitem[Ashraf et~al.(2008)Ashraf, Lucey, and Chen]{cvpr2008patch}
A.~Ashraf, S.~Lucey, and T.~Chen, ``Learning patch correspondences for improved
  viewpoint invariant face recognition,'' in \emph{IEEE Conference on Computer
  Vision and Pattern Recognition, 2008. CVPR 2008.}, june 2008, pp. 1 --8.

\bibitem[Jorstad et~al.(2011)Jorstad, Jacobs, and Trouve]{cvpr2011flowface}
A.~Jorstad, D.~Jacobs, and A.~Trouve, ``A deformation and lighting insensitive
  metric for face recognition based on dense correspondences,'' in \emph{IEEE
  Conference on Computer Vision and Pattern Recognition (CVPR), 2011}, june
  2011, pp. 2353 --2360.

\bibitem[Arashloo and Kittler(2011)]{pami2011-mrf-face}
S.~Arashloo and J.~Kittler, ``Energy normalization for pose-invariant face
  recognition based on {MRF} model image matching,'' \emph{IEEE Transactions on
  Pattern Analysis and Machine Intelligence}, vol.~33, no.~6, pp. 1274 --1280,
  june 2011.

\bibitem[Castillo and Jacobs(2011)]{cvpr2011widebaseline}
C.~Castillo and D.~Jacobs, ``Wide-baseline stereo for face recognition with
  large pose variation,'' in \emph{IEEE Conference on Computer Vision and
  Pattern Recognition (CVPR), 2011}, june 2011, pp. 537 --544.

\bibitem[Cootes et~al.(2001)Cootes, Edwards, and Taylor]{cootes2001aam}
T.~Cootes, G.~Edwards, and C.~Taylor, ``Active appearance models,'' \emph{IEEE
  Transactions on Pattern Analysis and Machine Intelligence}, vol.~23, no.~6,
  pp. 681 --685, jun 2001.

\bibitem[Matthews and Baker(2004)]{AAMRevisited}
I.~Matthews and S.~Baker, ``Active appearance models revisited,''
  \emph{International Journal of Computer Vision}, vol.~60, no.~2, pp.
  135--164, 2004.

\bibitem[Wiskott et~al.(1997)Wiskott, Fellous, Kruger, and von~der
  Malsburg]{ElasticBunch}
L.~Wiskott, J.-M. Fellous, N.~Kruger, and C.~von~der Malsburg, ``Face
  recognition by elastic bunch graph matching,'' in \emph{Proceedings.,
  International Conference on Image Processing, 1997.}, vol.~1, oct 1997, pp.
  129 --132 vol.1.

\bibitem[Cristinacce and Cootes(2006)]{CLM}
D.~Cristinacce and T.~Cootes, ``Feature detection and tracking with constrained
  local models,'' in \emph{Proc. British Machine Vision Conference}, vol.~3,
  2006, pp. 929--938.

\bibitem[Belhumeur et~al.(2011)Belhumeur, Jacobs, Kriegman, and
  Kumar]{BelhumeurCLM}
P.~Belhumeur, D.~Jacobs, D.~Kriegman, and N.~Kumar, ``Localizing parts of faces
  using a consensus of exemplars,'' in \emph{IEEE Conference on Computer Vision
  and Pattern Recognition (CVPR), 2011}, june 2011, pp. 545 --552.

\bibitem[Saragih et~al.(2011)Saragih, Lucey, and Cohn]{SaragihCLM}
J.~Saragih, S.~Lucey, and J.~Cohn, ``\BIBforeignlanguage{English}{Deformable
  model fitting by regularized landmark mean-shift},''
  \emph{\BIBforeignlanguage{English}{International Journal of Computer
  Vision}}, vol.~91, pp. 200--215, 2011.

\bibitem[Felzenszwalb et~al.(2010)Felzenszwalb, Girshick, McAllester, and
  Ramanan]{pff2010dpm}
P.~Felzenszwalb, R.~Girshick, D.~McAllester, and D.~Ramanan, ``Object detection
  with discriminatively trained part-based models,'' \emph{IEEE Transactions on
  Pattern Analysis and Machine Intelligence}, vol.~32, no.~9, pp. 1627 --1645,
  sept. 2010.

\bibitem[Zhu and Ramanan(2012)]{zhu2012dpmface}
X.~Zhu and D.~Ramanan, ``Face detection, pose estimation, and landmark
  localization in the wild,'' in \emph{IEEE Conference on Computer Vision and
  Pattern Recognition (CVPR), 2012}, june 2012, pp. 2879 --2886.

\bibitem[Wright et~al.(2009)Wright, Yang, Ganesh, Sastry, and Ma]{john2009src}
J.~Wright, A.~Yang, A.~Ganesh, S.~Sastry, and Y.~Ma, ``Robust face recognition
  via sparse representation,'' \emph{IEEE Transactions on Pattern Analysis and
  Machine Intelligence}, vol.~31, no.~2, pp. 210 --227, feb. 2009.

\bibitem[Su et~al.(2006)Su, Shan, Chen, and Gao]{EGFC}
Y.~Su, S.~Shan, X.~Chen, and W.~Gao, ``Hierarchical ensemble of gabor fisher
  classifier for face recognition,'' in \emph{Automatic Face and Gesture
  Recognition, 2006. FGR 2006. 7th International Conference on}, 2006, pp. 6
  pp.--96.

\bibitem[Pan et~al.(2007)Pan, Liao, Zhang, Li, and Zhang]{part-face-nir}
K.~Pan, S.~Liao, Z.~Zhang, S.~Li, and P.~Zhang, ``Part-based face recognition
  using near infrared images,'' in \emph{Computer Vision and Pattern
  Recognition, 2007. CVPR '07. IEEE Conference on}, 2007, pp. 1--6.

\bibitem[Kumar et~al.(2011)Kumar, Banerjee, Vemuri, and
  Pfister]{kumar2011kernel-plurality}
R.~Kumar, A.~Banerjee, B.~Vemuri, and H.~Pfister, ``Maximizing all margins:
  Pushing face recognition with kernel plurality,'' in \emph{2011 IEEE
  International Conference on Computer Vision (ICCV)}, 2011, pp. 2375--2382.

\bibitem[Zhang et~al.(2011)Zhang, Nasrabadi, Zhang, and
  Huang]{zhang2011joint-dynamic}
H.~Zhang, N.~Nasrabadi, Y.~Zhang, and T.~Huang, ``Multi-observation visual
  recognition via joint dynamic sparse representation,'' in \emph{2011 IEEE
  International Conference on Computer Vision (ICCV)}, 2011, pp. 595--602.

\bibitem[Ocegueda et~al.(2011)Ocegueda, Shah, and
  Kakadiaris]{Ocegueda2011which-part}
O.~Ocegueda, S.~Shah, and I.~Kakadiaris, ``Which parts of the face give out
  your identity?'' in \emph{2011 IEEE Conference on Computer Vision and Pattern
  Recognition (CVPR)}, 2011, pp. 641--648.

\bibitem[Gross et~al.(2010)Gross, Matthews, Cohn, Kanade, and Baker]{MultiPIE}
R.~Gross, I.~Matthews, J.~Cohn, T.~Kanade, and S.~Baker, ``{Multi-PIE},''
  \emph{Image and Vision Computing}, vol.~28, no.~5, pp. 807 -- 813, 2010, best
  of Automatic Face and Gesture Recognition 2008.

\bibitem[Milborrow et~al.(2010)Milborrow, Morkel, and Nicolls]{MUCT}
S.~Milborrow, J.~Morkel, and F.~Nicolls, ``{The MUCT Landmarked Face
  Database},'' \emph{Pattern Recognition Association of South Africa}, 2010,
  \url{http://www.milbo.org/muct/}.

\bibitem[Li et~al.(2012)Li, Liu, Chai, Zhang, Lao, and Shan]{eccv2012facepose}
S.~Li, X.~Liu, X.~Chai, H.~Zhang, S.~Lao, and S.~Shan, ``Morphable displacement
  field based image matching for face recognition across pose,'' in
  \emph{European Conference on Computer Vision (ECCV) 2012}, ser. Lecture Notes
  in Computer Science, vol. 7572.\hskip 1em plus 0.5em minus 0.4em\relax
  Springer Berlin Heidelberg, 2012, pp. 102--115.

\bibitem[Asthana et~al.(2011)Asthana, Marks, Jones, Tieu, and
  Rohith]{cvpr2011pose3d}
A.~Asthana, T.~Marks, M.~Jones, K.~Tieu, and M.~Rohith, ``Fully automatic
  pose-invariant face recognition via {3D} pose normalization,'' in \emph{IEEE
  International Conference on Computer Vision (ICCV), 2011}, nov. 2011, pp. 937
  --944.

\bibitem[Blanz and Vetter(2003)]{pami2003face3d}
V.~Blanz and T.~Vetter, ``Face recognition based on fitting a {3D} morphable
  model,'' \emph{IEEE Transactions on Pattern Analysis and Machine
  Intelligence}, vol.~25, no.~9, pp. 1063 -- 1074, sept. 2003.

\bibitem[Yang et~al.(2012)Yang, Hallman, Ramanan, and
  Fowlkes]{yang2012dpmhuman}
Y.~Yang, S.~Hallman, D.~Ramanan, and C.~Fowlkes, ``Layered object models for
  image segmentation,'' \emph{IEEE Transactions on Pattern Analysis and Machine
  Intelligence}, vol.~34, no.~9, pp. 1731 --1743, sept. 2012.

\bibitem[Peng et~al.(2012)Peng, Ganesh, Wright, Xu, and Ma]{peng2011rasl}
Y.~Peng, A.~Ganesh, J.~Wright, W.~Xu, and Y.~Ma, ``{RASL}: Robust alignment by
  sparse and low-rank decomposition for linearly correlated images,''
  \emph{IEEE Transactions on Pattern Analysis and Machine Intelligence},
  vol.~PP, no.~99, p.~1, 2012.

\bibitem[Lin et~al.(2009)Lin, Chen, and Ma]{lin2009alm}
Z.~Lin, M.~Chen, and Y.~Ma, ``The augmented lagrange multiplier method for
  exact recovery of corrupted low-rank matrices,'' University of Illinois at
  Urbana-Champaign, Tech. Rep. UILU-ENG-09-2215, 2009.

\bibitem[Bertsekas(1999)]{bertsekas1999nonlinear}
D.~Bertsekas, \emph{Nonlinear programming}.\hskip 1em plus 0.5em minus
  0.4em\relax Athena Scientific, 1999.

\bibitem[Viola and Jones(2001)]{viola2001adaboost}
P.~Viola and M.~Jones, ``Rapid object detection using a boosted cascade of
  simple features,'' in \emph{Proceedings of the 2001 IEEE Computer Society
  Conference on Computer Vision and Pattern Recognition, 2001. CVPR 2001.},
  vol.~1, 2001, pp. I--511 -- I--518 vol.1.

\bibitem[Ahonen et~al.(2006)Ahonen, Hadid, and Pietikainen]{pami2006lbp}
T.~Ahonen, A.~Hadid, and M.~Pietikainen, ``Face description with local binary
  patterns: Application to face recognition,'' \emph{IEEE Transactions on
  Pattern Analysis and Machine Intelligence}, vol.~28, no.~12, pp. 2037 --2041,
  dec. 2006.

\bibitem[Belhumeur et~al.(1997)Belhumeur, Hespanha, and
  Kriegman]{pami1997fisherface}
P.~Belhumeur, J.~Hespanha, and D.~Kriegman, ``{E}igenfaces vs. {F}isherfaces:
  recognition using class specific linear projection,'' \emph{IEEE Transactions
  on Pattern Analysis and Machine Intelligence}, vol.~19, no.~7, pp. 711 --720,
  jul 1997.

\bibitem[Cand\`{e}s et~al.(2011)Cand\`{e}s, Li, Ma, and Wright]{acm2011rpca}
E.~J. Cand\`{e}s, X.~Li, Y.~Ma, and J.~Wright, ``Robust principal component
  analysis?'' \emph{J. ACM}, vol.~58, no.~3, pp. 11:1--11:37, Jun. 2011.

\bibitem[Chow and Liu(1968)]{chow-liu}
C.~Chow and C.~Liu, ``Approximating discrete probability distributions with
  dependence trees,'' \emph{IEEE Transactions on Information Theory}, vol.~14,
  no.~3, pp. 462 -- 467, may 1968.

\bibitem[\v{S}truc and Pave\v{s}ic(2009)]{nlm-illumination}
V.~\v{S}truc and N.~Pave\v{s}ic, ``Illumination invariant face recognition by
  non-local smoothing,'' in \emph{Biometric ID Management and Multimodal
  Communication}, ser. Lecture Notes in Computer Science, J.~Fierrez,
  J.~Ortega-Garcia, A.~Esposito, A.~Drygajlo, and M.~Faundez-Zanuy, Eds.\hskip
  1em plus 0.5em minus 0.4em\relax Springer Berlin Heidelberg, 2009, vol. 5707,
  pp. 1--8.

\end{thebibliography}

\end{document}